\documentclass{article}

\usepackage{microtype}
\usepackage{graphicx}
\usepackage{subcaption}
\usepackage{booktabs} % for professional tables

\usepackage{hyperref}

\usepackage[accepted]{icml2026}

\usepackage{amsmath,amssymb,amsfonts}
\usepackage{mathtools}
\usepackage{amsthm}
\usepackage{enumitem}
\usepackage{multirow}
\usepackage[svgnames,x11names]{xcolor}

\usepackage{stfloats}
\input{styles/mysymbol.sty}

\usepackage{tikz}
\usetikzlibrary{shapes,arrows}
\usepackage{moresize}
\usepackage{pgfplots}
\usepackage{wrapfig}
\pgfplotsset{compat=1.17}
\usepgfplotslibrary{fillbetween}
\usepgfplotslibrary{groupplots}
\pgfplotstableset{col sep=comma}
\newcommand{\indep}{\perp \!\!\! \perp}

\def \diag {{\rm diag}}
\def \vect {{\rm vec}}

\definecolor{madblack}{HTML}{47407d}
\definecolor{madpink}{HTML}{dc267f}
\definecolor{madblue}{HTML}{648fff}
\definecolor{madyellow}{HTML}{ffb000}

\definecolor{darkpink}{HTML}{C4186C}
\definecolor{darkblue}{HTML}{426CD6}

\pgfplotsset{compat=1.18}
\pgfplotstableset{col sep=comma}
\pgfplotsset{
    width=1.05\textwidth,
    legend cell align=left,
	every axis/.append style={line width=0.5pt},
	every axis plot/.append style={line width=.9pt},
}
\usepackage[capitalize,noabbrev]{cleveref}

\usepackage[textsize=tiny]{todonotes}

\icmltitlerunning{Adaptive Node Feature Selection For Graph Neural Networks}

\begin{document}

\twocolumn[
  \icmltitle{Adaptive Node Feature Selection for Graph Neural Networks}



  \icmlsetsymbol{equal}{*}

  \begin{icmlauthorlist}
    \icmlauthor{Madeline Navarro}{ece}
    \icmlauthor{Ali Azizpour}{ece}
    \icmlauthor{Santiago Segarra}{ece}
  \end{icmlauthorlist}

  \icmlaffiliation{ece}{Department of Electrical and Computer Engineering, Rice University, Houston, TX, USA}

  \icmlcorrespondingauthor{Madeline Navarro}{nav@rice.edu}
  \icmlcorrespondingauthor{Ali Azizpour}{ali.azizpour@rice.edu}
  

  \icmlkeywords{Feature Importance, Feature Selection, Graph Neural Networks}

  \vskip 0.3in
]



\printAffiliationsAndNotice{}  % no special notice (required even if empty)

\begin{abstract}
We propose an adaptive node feature selection approach for graph neural networks (GNNs) that identifies and removes unnecessary features during training.
% The ability to measure how features contribute to model output is key for interpreting decisions, reducing dimensionality, and even improving performance by eliminating unhelpful variables.
The ability to measure how features contribute to model output is key for interpreting decisions and reducing dimensionality by eliminating unhelpful variables.
However, graph-structured data introduces complex dependencies that may be unsuited to classical feature importance metrics.
Inspired by this, we present a data-, model-, and task-agnostic method that determines relevant features during training based on changes in validation performance upon permuting feature values.
We theoretically motivate our approach by characterizing how the relationships between node data and graph structure influences GNN performance.
Empirically, we show that
(i) our highly general approach rivals the performance of tailored feature selection approaches that exploit prior assumptions;
(ii) we return meaningful feature importance scores well before the GNN is fully trained; and
(iii) our scores demonstrably extract relevant properties that inform feature importance for various graph learning settings.
% Not only do we return feature importance scores once training concludes, we also track how relevance evolves as features are successively dropped.
% We can therefore monitor if features are eliminated effectively and also evaluate other metrics with this technique. 
% Our empirical results verify the flexibility of our approach to different graph architectures as well as its adaptability to more challenging graph learning settings.
\end{abstract}

\section{Introduction}\label{S:intro}

Graphs provide powerful yet well-understood representations of complex data~\citep{bronstein2017GeometricDeepLearning}. 
Their rich modeling capabilities motivated graph neural networks (GNNs) for exploiting connectivity for predictive tasks~\citep{wu2021ComprehensiveSurveyGraph}. 
However, insufficient understanding of model decisions renders them untrustworthy for critical applications and potentially inefficient or suboptimal~\citep{dong2022StructuralExplanationBias,yuan2023ExplainabilityGraphNeural,wang2025InnovativeBiomarkerExploration,chien2024Opportunitieschallengesgraph}. 
Deciphering how deep learning models extract information from data is challenging, particularly when data is equipped with complex interdependencies~\citep{zhu2024ImpactFeatureHeterophily}. 
While some tools such as decision trees inherently provide model explanations, the most expressive tools are not directly interpretable and require explanation via heuristic-based metrics~\citep{mandler2024ReviewBenchmarkFeature}. 
As a prominent example, measuring feature importance is a fundamental technique for understanding how a model forms decisions~\citep{wang2024Featureselectionstrategies}. 
In particular, we seek to explain how node features contribute to GNN outputs by answering the following question: \emph{How do graph connections affect how important a given node feature is to GNN performance?}

Beyond interpretability, identifying relevant attributes allows us to build models that are both economical and potent by eliminating unnecessary features~\citep{li2018FeatureSelectionData}. 
Moreover, simplifying models can improve our understanding of complex real-world systems by reducing them to their most parsimonious representations~\citep{georg2023Lowranktensormethods,shao2024reviewfeatureselection}. 
However, interpretable feature selection for GNNs suffers four obstacles.
First, although relationships among graph structure, node labels, and features can dictate node classification accuracy~\citep{luan2024WhenGraphNeural,zheng2024WhatMissingGraph}, many works still rely on classical feature importance metrics that do not account for an underlying graph~\citep{chereda2024Stablefeatureselection,mahmoud2023Nodeclassificationgraph}.
Second, past graph-based feature selection methods often involve assumptions on how graph structure contributes to learning, rendering these techniques problem-specific~\citep{maurya2022Simplifyingapproachnode,maurya2023FeatureselectionKey,zheng2025letyourfeatures}. 
Third, while empirically successful, these models perform black-box feature selection without returning meaningful importance scores, precluding novel domain insights and warranting caution in their use.
Finally, to remove dependence on prior information, we can measure changes in model performance upon perturbing features to assess their contributions~\citep{datta2016AlgorithmicTransparencyQuantitative,fisher2019AllModelsWrong}. 
    % "The holdout randomization test for feature selection in black box models" 
    % "Proxy non-discrimination in data-driven systems" 
    % "Feature selection strategies: a comparative analysis of SHAP-value and importance-based methods"
However, most approaches require implementing and possibly even training a model multiple times, which can be costly for large-scale data or complicated architectures~\citep{alkhoury2025Improvinggraphneural}.
% However, feature selection using perturbation-based scores may require training multiple models, which can be costly for large-scale data or complicated architectures~\citep{alkhoury2025Improvinggraphneural}. 
% As these measurements require a trained model, 
    % "Feature selection for huge data via minipatch learning"? 
% While some works train submodules to learn masks for identifying important features, these approaches can require learning additional parameters, undermining the goal of reducing dimensionality~\citep{maurya2022Simplifyingapproachnode,acharya2020FeatureSelectionExtraction,lin2020GraphNeuralNetworks,zheng2020GSSAPayattention}.
A more in-depth overview of related works is shared in Appendix~\ref{app:related}.

\begin{figure*}[t]
    \centering
    \vspace{-.2cm}
    \hspace{-1.8cm}
    \begin{subfigure}[t]{0.4\textwidth}
        \centering
        \scalebox{.53}{\includegraphics[width=\linewidth]{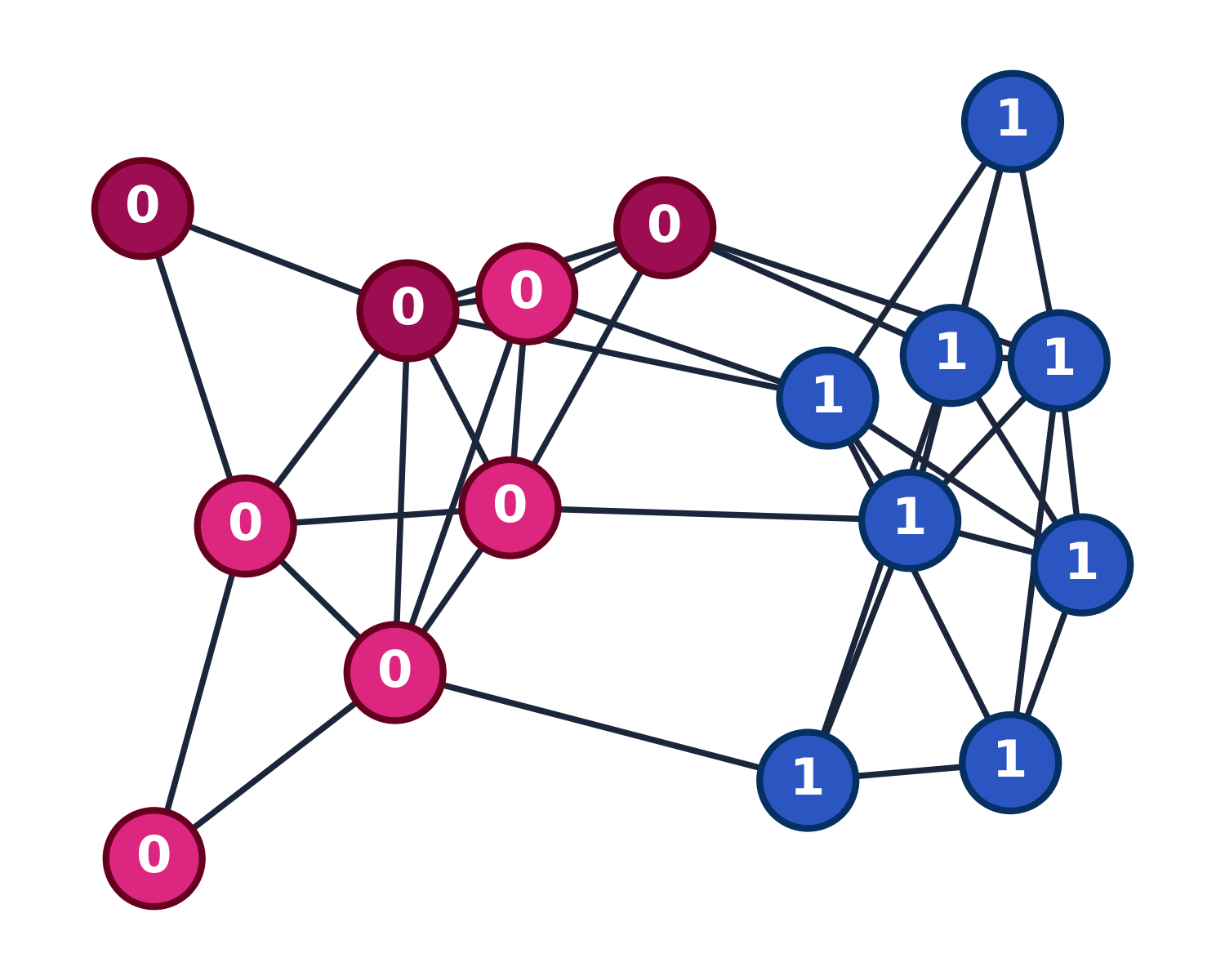}}
        \vspace{-.30cm}
        \caption{Homophilic node features}
        \label{fig:motivation_a}
    \end{subfigure}
    % \hfill
    % \hspace{.2cm}
    \hspace{-2.1cm}
    \begin{subfigure}[t]{0.4\textwidth}
        \centering
        \scalebox{.53}{\includegraphics[width=\linewidth]{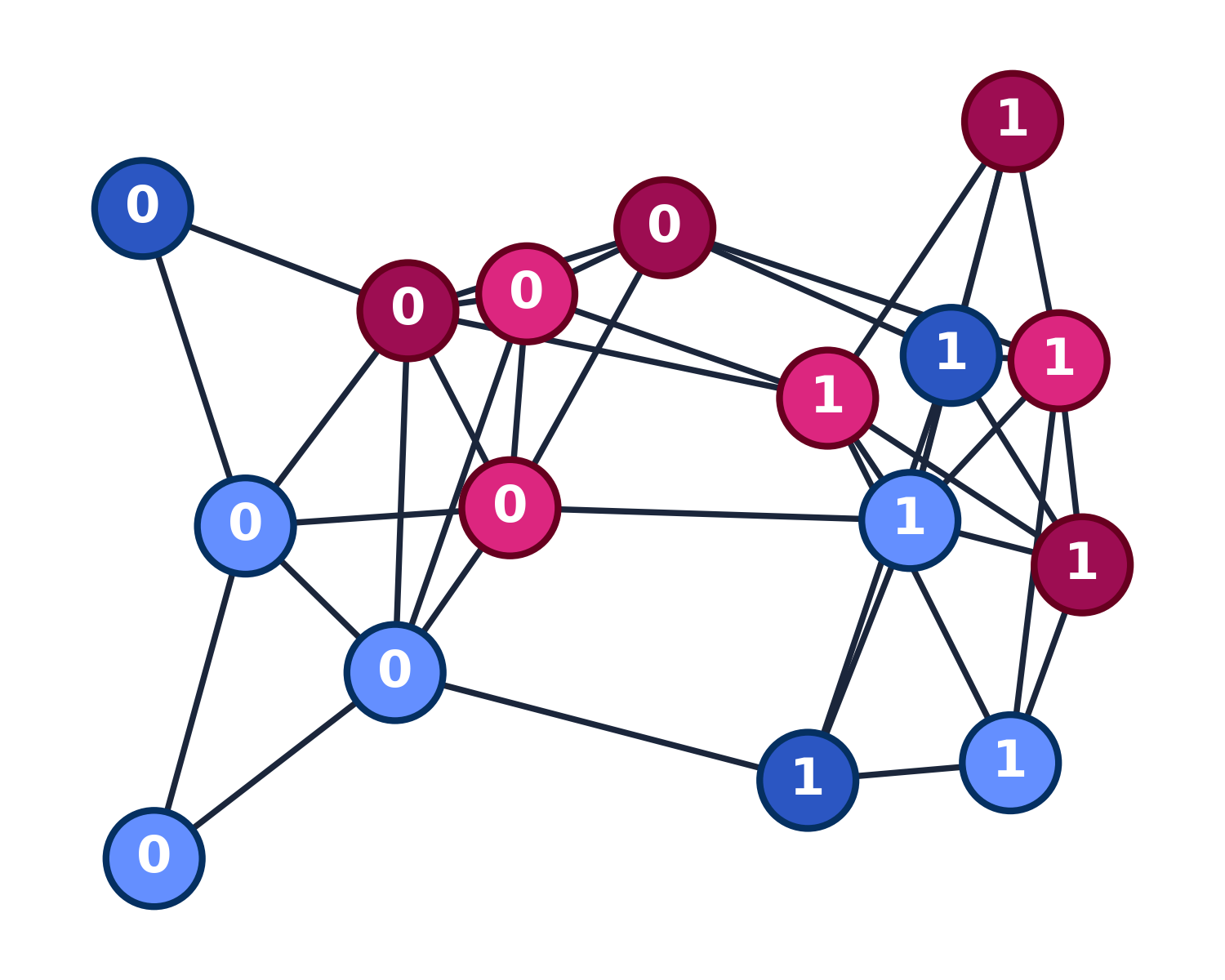}}
        \vspace{-.30cm}
        \caption{Heterophilic node features}
        \label{fig:motivation_b}
    \end{subfigure}
    % \hfill
    % \hspace{.2cm}
    \hspace{-1.1cm}
    \begin{subfigure}[t]{0.41\textwidth}
        \centering
        \scalebox{.53}{\includegraphics[width=\linewidth]{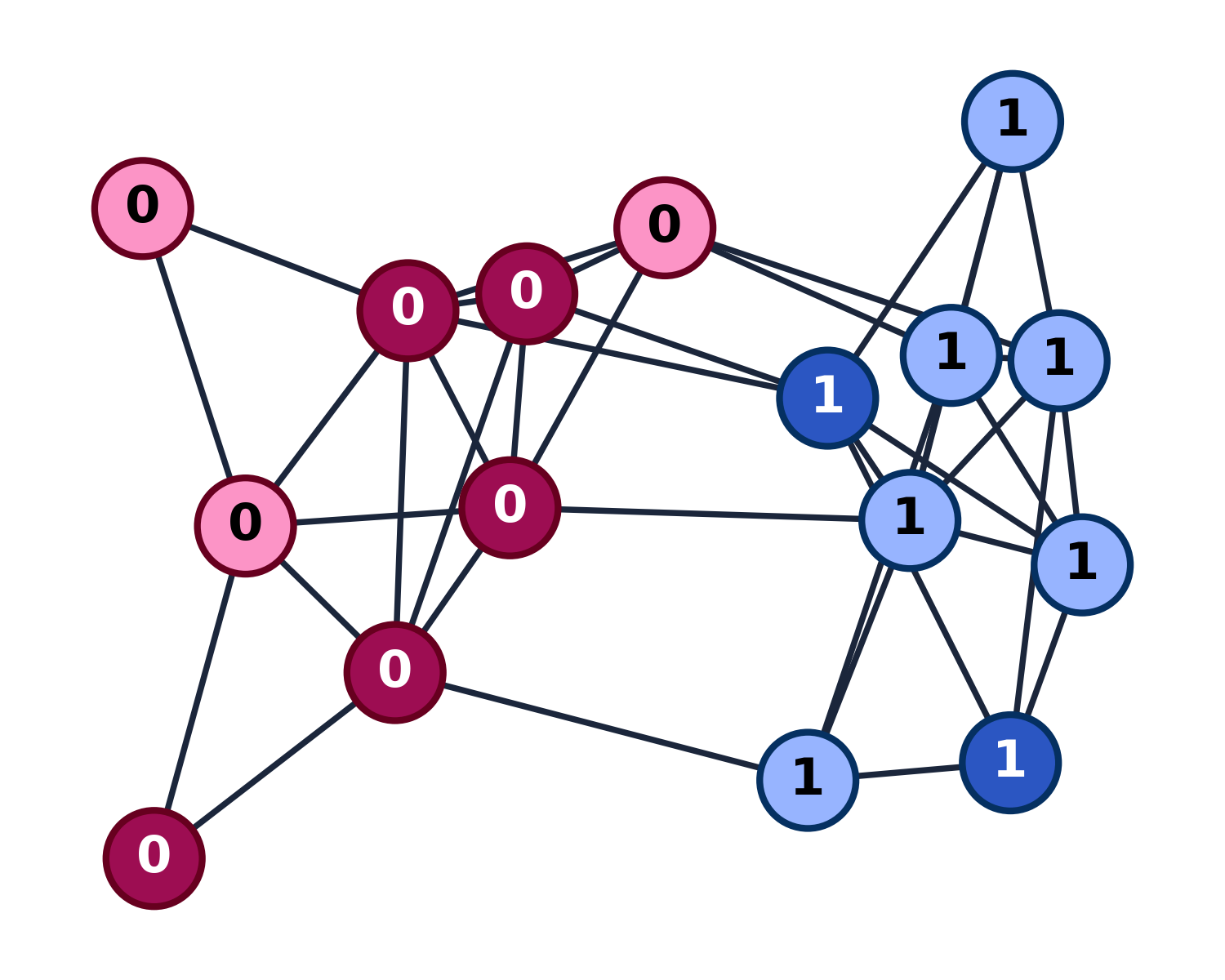}}
        \vspace{-.30cm}
        \caption{Homophilic node features, high within-class variance}
        \label{fig:motivation_c}
    \end{subfigure}
    \hspace{-.7cm}
    \vspace{-.15cm}
    \caption{
        Illustration of how graph structure modulates the effect of features on node classification.
        % Example graphs illustrating how graph structure modulates the effect of node features on node classification.
        Classes are labeled ``0'' or ``1'', and colors denote node features, where {\color{darkpink}red} and {\color{darkblue}blue} indicate features in different distributions and brightness indicates magnitude.
        (a)~Edges directly imply similarity in node labels and features.
        (b)~Edges mostly connect nodes of the same class but also reflect distribution shifts in node features.
        (c)~Both node labels and features are homophilic, but high variance in node features makes classification more challenging.
    }
    \label{f:motivation}
    \vspace{-.35cm}
\end{figure*}

Instead, we propose an \textit{adaptive node feature selection algorithm} that measures \textit{permutation-based feature importance during training} using GNN predictions.
More specifically, we periodically permute the values of each node feature and measure changes in GNN performance on a validation dataset.
Our scores are thus inherently tied to the predictive task and therefore flexible to GNN architecture with no assumptions on graph data.
Indeed, despite being data- and model-agnostic, our approach empirically adapts to different scenarios and rivals other feature selection methods that exploit prior information.
Moreover, unlike node feature selection works that learn importance values via black-box models, we employ well-established permutation tests to quantify feature influence~\citep{altmann2010Permutationimportancecorrected,yang2009FeatureSelectionMLP,brieman2001RandomForests,datta2016AlgorithmicTransparencyQuantitative}, allowing us to theoretically show how permutations reflect node feature influence. 
Our contributions are as follows.

\vspace{-.4cm}

\begin{itemize}[left= 2pt .. 10pt, noitemsep]
    \item 
        We characterize how graph structure, node labels, and features jointly impact GNN performance, both theoretically and empirically.
        For the former, we prove how edges can change how relevant a node feature is to graph convolutional network (GCN) outputs.
        For the latter, we demonstrate how perturbations can reflect which structure- and feature-based properties affect GNN accuracy.
        % how relationship between graph structure and node features influences GNN performance
        % for former, show how connectivity changes the influence of node features for GCN outputs
        % for latter, compare ... which verifies how perturbations can identify different kinds of relationships between feature and graph affecting node classification accuracy
        % 
        % We first characterize the effects of graph structure and node features on GNN performance, both theoretically and empirically.
        % For the former, we show how connections influence the effect of node features on graph convolutional network (GCN) outputs.
        % For the latter, we compare GNN accuracy under various perturbations to distinguish model dependence on graphs versus features.
    \item 
        We propose an adaptive node feature selection approach that dynamically identifies which features are relevant to GNN performance via permutation-based importance scores.
        We measure scores during training to track model quality and verify that we indeed eliminate unhelpful attributes.
        Crucially, we show that our scores reflect feature importance well before the GNN is finished training.
        % Because we measure these scores as training progresses, we can monitor how feature contributions change as the model evolves and variables are eliminated. 
        % We thus visualize importance scores during training to track model quality and verify that we indeed eliminate unhelpful attributes.
    \item 
        We show that our scores identify features relevant to the task by (i) rivaling the performance of other feature importance metrics for benchmark datasets and (ii) showing that our data-agnostic scores always behave most similarly to the top baseline, which differs across datasets.
        % showing that our data-agnostic scores highlight features with similar properties the top baselines, which leverage prior information.
        Thus, our results also verify that our approach is flexible to model architecture and across various settings, such as homophilic or heterophilic node labels.
        % we demonstrate that our algo adaptively learns important features according to task at hand
        % by showing that we rival performance of all features and other FS approaches, as well as showing that our feature importance scores identify features of the same properties as the top performing methods, even though we do not have any of the prior information required by all of these
        % thus, our results include verifying that flexible to model architecture in various settings, ...
        % We demonstrate that our algorithm rivals the performance of a GNN using all available node features in comparison with other node feature selection methods for multiple benchmark datasets.
        % Furthermore, we show that our approach is flexible to model architecture and for various settings, such as homophilic or heterophilic node labels.
\end{itemize}

\subsection{Notation}
For any positive integer $N \in \naturals$, we let $[N] := \{1,2,\dots,N\}$.
For the vector $\bbx \in \reals^N$, we index entries via $x_i$ for any $i\in[N]$, whereas for a matrix $\bbX \in \reals^{N\times M}$, we index entries by $X_{ij}$, rows by $\bbX_{i,:}$, and columns by $\bbX_{:,j}$.
We let boldfaced numbers $\bbzero$ and $\bbone$ represent vectors or matrices of all zeros and ones, respectively, while $\bbI$ is the identity matrix.
% Furthermore, $\bbI$ is the identity matrix and $\bbe_i = \bbI_{:,i}$ is the $i$-th standard basis vector.
% For $\bbzero$, $\bbone$, and $\bbI$, we specify dimensions when it is unclear from context.
% For $\bbzero$, $\bbone$, $\bbI$, and $\bbe_i$, we specify dimensions when it is unclear from context.
The operator $\diag(\bbx) \in \reals^{N\times N}$ evaluated on a vector $\bbx \in \reals^N$ returns a diagonal matrix with entries of $\bbx$ along the diagonal, while $\diag(\bbX) \in \reals^N$ for a square $\bbX \in \reals^{N\times N}$ returns a vector of the diagonal entries of $\bbX$.
We also let $\vect(\bbX)\in\reals^{NM}$ return the concatenation of columns in the matrix $\bbX \in \reals^{N\times M}$.
Moreover, let $\mbI(\cdot)$ denote the indicator function, where $\mbI(\ccalA) = 1$ when its argument $\ccalA$ is true and $\mbI(\ccalA) = 0$ otherwise.

\vspace{-.2cm}

\paragraph{Conflict of Interest Disclosure}
The authors declare no competing interests.

\vspace{-.2cm}
% %%%%%%%%%%%%%%%%%%%%%%%%%%%%%%%%%%%%%%%%%%%%%

% %%%%%%%%%%%%%%%%%%%%%%%%%%%%%%%%%%%%%%%%%%%%%
\section{Feature importance for node classification}\label{S:nfi}

We are interested in semi-supervised node classification, where we have a graph $\ccalG = (\ccalV, \ccalE)$ of $N$ nodes $\ccalV$ and a set of edges $\ccalE \subseteq \ccalV \times \ccalV$ connecting pairs of nodes in $\ccalV$.
To use a graph in model training, we consider the adjacency matrix $\bbA \in \reals_+^{N\times N}$, where $A_{ij} \neq 0$ if and only if the edge $(i,j) \in \ccalE$ connects nodes $i$ and $j$, and $A_{ij} > 0$ denotes weight of the edge $(i,j)$.
We can account for nodes with differing degrees $\bbd := \bbA\bbone$ by employing \textit{normalized} adjacency matrices such as $\tbA := \tilde{\bbD}^{-1/2}(\bbA+\bbI)\tilde{\bbD}^{-1/2}$ for $\tilde{\bbD} := \diag(\bbd + \bbone)$ or a random-walk adjacency matrix $\tbA_{\rm rw} := \tilde{\bbD}^{-1}(\bbA + \bbI)$~\citep{kipf2017SemisupervisedClassificationGraph}.
In addition to graph connections, each node is equipped with $M$ real-valued features, which we collect in the data matrix $\bbX \in \reals^{N\times M}$.
Furthermore, nodes are assigned labels $\bby = [\bby_{\rm train}^\top,\bby_{\rm val}^\top, \bby_{\rm test}^\top]^\top \in [C]^N$, of which we only observe a subset $[\bby_{\rm train}^\top,\bby_{\rm val}^\top] \in [C]^{N_{\rm train} + N_{\rm val}}$ for $N_{\rm train}, N_{\rm val} < N$.
We also let $\bbY \in \{ 0, 1 \}^{N\times C}$ denote the one-hot matrix indicating the class of each node, along with $\bbP := \diag(\bbp)$ for $\bbp := \bbY^\top \bbone \in \naturals^C$, which contains the number of nodes in each class.
We aim to predict the unknown labels $\bby_{\rm test}$ by learning the parameters of a GNN $f(\cdot;\cdot,\bbTheta): \reals^{N\times M} \rightarrow \reals^{N\times H}$ that yields embeddings $\bbZ := f(\bbX;\bbA,\bbTheta)$ such that we may predict labels $\hby = g(\bbZ)$ with some classifier $g : \reals^{N\times H} \rightarrow [C]^N$.

% Importantly, we also aim to identify
Of particular relevance to us is how to identify which node features in $\bbX$ are important for predicting labels $\bby$ while accounting for the graph structure $\bbA$~\citep{maurya2023FeatureselectionKey,chen2020ArePowerfulGraph}.
Some works apply traditional, graph-agnostic metrics to determine important features for a pre-trained GNN~\citep{wang2025InnovativeBiomarkerExploration,basaad2024GraphXNetGraphNeural,chereda2024Stablefeatureselection}. % GNN+SHAP, etc
However, the presence of edges used by the GNN can significantly alter which node features are relevant.
For example, GCNs assume that edges directly indicate nodes that likely belong to the same class.
Figure~\ref{f:motivation} illustrates how this assumption can alter how informative node features are.
As GCNs are best suited to homophilic node features and labels as in Figure~\ref{f:motivation}a, it is common to assess feature quality through its smoothness, that is, how similar feature values are between connected nodes~\citep{zhu2024ImpactFeatureHeterophily}.
However, even with homophilic node labels, a GCN applied to the graph in Figure~\ref{f:motivation}b may not yield sufficiently separable node embeddings~\citep{luan2024WhenGraphNeural}.
Furthermore, if labels are homophilic and node features in different classes follow distinctly different distributions yet exhibit high variance, exemplified in Figure~\ref{f:motivation}c, a graph-agnostic classifier may distinguish classes more easily than a GCN.
Motivated by this consideration, we theoretically characterize how $\bbA$ and $\bbX$ influence GNN performance, which we then empirically verify on real-world graph data.

\subsection{Influence of Graph Data on GCN Performance}\label{Ss:gcn_performance}

Recall that our goal is for our embeddings $\bbZ = f(\bbX;\bbA,\bbTheta)$ to be distinguishable across classes.
A reasonable requirement for this task is that node embeddings exhibit sufficient separation across classes~\citep{tenorio2025adaptingheterophilicgraphdata,nt2021RevisitingGraphNeural}. %sggcn, ng
However, we encounter at least two potential sources of error: noise in features $\bbX$ and in edges $\bbA$.
For the former, we let the \textit{idealized} node features to be $\bbX^* := \bbY\bbP^{-1} \bbY^\top \bbX$, that is, the matrix closest to $\bbX$ whose rows are identical for nodes in the same class, or equivalently,
\alna{
    &\nonumber\\[-.7cm]&
    \bbX^* 
    = 
    \argmin_{\bbX^*}
    \| \bbX^* - \bbX \|_F^2
    ~~~ {\rm s.t.} ~~
    &&
    \bbX_{i,:}^* = \bbX_{j,:}^*
    ~\forall~
    i,j\in[N]
&\nonumber\\[-.17cm]&
    &&
    ~~{\rm s.t.}~~
    y_i = y_j.
    &\label{eq:ideal_feats}\\[-.7cm]& \nonumber}
    % \label{eq:ideal_feats}}
By~\eqref{eq:ideal_feats}, we obtain a notion of feature informativeness: Even if the rows of $\bbX^*$ are equivalent within classes, they may be very similar or even identical across classes, rendering classification effectively infeasible~\citep{nt2021RevisitingGraphNeural}. 
Thus, we consider $\bbX$ to be informative enough if $\bbX^*$ contains distinct rows for different classes, indicating a sufficient shift in feature distributions across classes~\citep{tenorio2025adaptingheterophilicgraphdata}.
Note that we define $\bbX^*$ as above for simplicity, representing the most straightforward relationship between informative features and labels $\bby$; node classes with distribution shifts can still be informative~\citep{luan2024WhenGraphNeural}.
Next, we proceed with the latter source of error, the effect of the graph structure $\bbA$ on GNN performance.

We demonstrate the influence of $\bbA$ for GCNs, as this scenario is well studied and intuitive.
This exemplifies a GNN for which a particular type of structure aids performance, namely, graphs over which $\bby$ is homophilic.
As we will show next, GCNs can perform well on such graphs, even if $\bby$ does not depend on the features $\bbX$.
Beyond GCNs, we also present analogous results in Appendix~\ref{app:gnn_performance}~characterizing the interplay between $\bbA$, $\bbX$, and $\bby$ for other GNNs, as well as examples in which heterophilic edges are informative along with architectures that can exploit them.

It is well established that cross-class edges, that is, those connecting nodes of different classes, mar GCN performance~\citep{zhu2020homophily}.
 % any heterophily paper
    % "Beyond homophily in graph neural networks: Current limitations and effective designs"
Hence, we define the \textit{idealized} graph $\bbA^*$ as having no cross-class edges, where
\alna{
    &\nonumber\\[-.6cm]&
    A^*_{ij}
    &~:=~&
    A_{ij}
    \cdot \mbI( y_i = y_j )
    \quad
    \forall ~
    i,j\in[N].
    &\label{eq:ideal_A}\\[-.6cm]& \nonumber}
% \label{eq:ideal_A}}
% We then let $\bbDelta := \bbA - \bbA^*$ collect all edges between nodes of different classes.
We next characterize the performance of a GCN with respect to features $\bbX$, edges $\bbA$, and labels $\bby$ by comparing our embeddings $\bbZ$ to the \textit{idealized} ones $\bbZ^* := f(\bbX^*;\bbA^*,\bbTheta)$.

\begin{theorem}\label{thm:gcn_error}
    % Let $f:\reals^{N\times M} \rightarrow \reals^{N\times H}$ be a two-layer GCN 
        Let $f:\reals^{N\times M} \rightarrow \reals^{N\times H}$ be an $L$-layer GCN with $\bbZ^{(0)} = \bbX$, $\bbZ = \bbZ^{(L)} = f(\bbX;\bbA,\bbTheta)$, and
        \alna{
            &\nonumber\\[-.7cm]&
            \bbZ^{(\ell)}
            =
            \sigma_{\ell}
            \Big(
                \tbA_{\rm rw}
                \bbZ^{(\ell-1)}
                \bbTheta^{(\ell)}
            \Big)
            \quad
            \forall~
            \ell \in [L]
        &\label{eq:gcn_single_layer}\\[-.7cm]& \nonumber}
        % &\label{eq:gcn_layer}\\[-.7cm]& \nonumber}
        % \label{eq:gcn_single_layer}}
        % 
        % \alna{
        %     &\nonumber\\[-.7cm]&
        %     f(\bbX;\bbA,\bbTheta)
        %     =
        %     \sigma\big(
        %         \tbA_{\rm rw} \sigma\big(
        %             \tbA_{\rm rw}
        %             \bbX
        %             \bbTheta^{(1)}
        %         \big)
        %         \bbTheta^{(2)}
        %     \big)
        % &\label{eq:gcn_single_layer}\\[-.7cm]& \nonumber}
        % \label{eq:gcn_single_layer}}
        % for $\tau$-Lipschitz activation $\sigma$ and weights $\bbTheta = (\bbTheta^{(1)}, \bbTheta^{(2)})$ such that $\| \bbTheta^{(\ell)} \|_2 \leq \omega$ for $\ell=1,2$.
        % Then, with $\bbZ^* = f(\bbX^*; \bbA^*, \bbTheta)$ for $\bbX^*$ in~\eqref{eq:ideal_feats}, $\bbA^*$ in~\eqref{eq:ideal_A}, and $\bbDelta := \bbA - \bbA^*$, 
        for $\tau$-Lipschitz activations $\sigma_{\ell}$ and weights $\bbTheta = \{ \bbTheta^{(\ell)} \}_{\ell=1}^{L}$ such that $\| \bbTheta^{(\ell)} \|_2 \leq \omega$ for $\ell \in [L]$.
        Then, with $\bbZ^* = f(\bbX^*; \bbA^*, \bbTheta)$ for $\bbX^*$ in~\eqref{eq:ideal_feats}, $\bbA^*$ in~\eqref{eq:ideal_A}, and $\bbDelta := \bbA - \bbA^*$,
        \alna{
            &\nonumber\\[-.65cm]&
            \!\!\!\!\!\!
            \| \bbZ^* - \bbZ \|_F
            % &~\leq~&
            \,\leq\,
            % \tau^2 \omega^2
            \tau^{L} \omega^{L}
            \!
            \Bigg[
                % \tau^2 \omega^2
                (L-1)
                (1 + \sqrt{N})
                \|\bbDelta\|_F \|\bbX\|_F
            &\nonumber\\&
            \!\!\!\!\!\!
                ~~
                +
                % \tau^2 \omega^2
                \sum_{c=1}^C
                \sum_{i=1}^N
                \sum_{j=1}^N
                \left|
                    \frac{Y_{ic}Y_{jc}}{p_c}
                    -
                    \frac{A_{ij}}{d_i + 1}
                \right|
                \!\cdot\!
                \| \bbX_{i,:} - \bbX_{j,:} \|_2
            \Bigg].
        &\label{eq:gcn_error}\\[-.65cm]& \nonumber}
        % &\nonumber\\[-.25cm]& 
        % &\label{eq:gcn_error}\\[-.9cm]& \nonumber}
        % \label{eq:gcn_error}}
\end{theorem}

The proof of Theorem~\ref{thm:gcn_error} can be found in Appendix~\ref{app:gcn_error}.
We share results for other GNN architectures in Appendix~\ref{app:gnn_performance}.
Our GCN error bound in~\eqref{eq:gcn_error} depends on cross-class edges in $\bbA$ via the first term and the alignment of labels $\bby$, edges in $\bbA$, and similarity of features in $\bbX$.
First, we discuss when $\bbA$ can render $\bbX$ harmful to GCN accuracy, even if $\bbX$ is informative without $\bbA$.
While the result in~\eqref{eq:gcn_error} does not necessitate unweighted edges, the following discussion assumes $\bbA \in \{ 0,1 \}^{N\times N}$ for ease of interpretation.
Unsurprisingly, GCNs require features $\bbX$ to be highly indicative of $\bby$ if $\bbA$ is sparse or noisy.
More specifically, for any node pair $i$ and $j$ in the same class $c$ that are not connected $(i,j) \notin \ccalE$, we rely on similarity between node features $\|\bbX_{i,:} - \bbX_{j,:}\|_2$ to reduce the second term in~\eqref{eq:gcn_error}.
Thus, $\bbA$ can render $\bbX$ unhelpful even if it is linearly separable between classes when $\bbX$ is not sufficiently well separated.
% if it is not sufficiently separated across classes, even if $\bbX$ is linearly separable between them.
% Thus, if $\bbX$ is not well separated across classes, can become harmful through $\bbA$, even if $\bbX$ is linearly separable between classes.
% an X that is not sufficiently well separated across class becomes harmful through A, even if X linearly separable across classes
% Thus, if X is linearly separable across classes but not sufficiently separated, a graph-agnostic method might deem it relevant 
% Thus, a noisy $\bbA$ can render $\bbX$ harmful even if it is linearly separable across classes, which a graph-agnostic metric may not capture.
However, even if $\bbX$ is highly separated, that is, $\|\bbX_{i,:} - \bbX_{j,:}\|_2$ is much higher when $Y_{ic}Y_{jc} = 0$, we incur greater error from cross-class connections $A_{ij} = 1$.
% However, $\bbA$ can also change the relevance of $\bbX$ for highly separable features, that is, when $\|\bbX_{i,:} - \bbX_{j,:}\|_2$ is higher if $Y_{ic}Y_{jc} = 0$, as we incur greater error from cross-class connections $A_{ij} = 1$.
In this setting, if we disregard $\bbA$ when selecting features, we may retain attributes that are highly separated with respect to classes, causing larger $\|\bbX_{i,:} - \bbX_{j,:}\|_2$ when $Y_{ic}Y_{jc} = 0$ and unknowingly introducing error due to cross-class edges in $\bbA$.

Conversely, the presence of $\bbA$ can also mitigate error due to  noisy features $\bbX$.
In particular, if $\bby$ is sufficiently homophilic with respect to $\bbA$, that is, if $Y_{ic}Y_{jc}=A_{ij}$ holds for sufficiently many node pairs, then we can still achieve a low error via~\eqref{eq:gcn_error}, even if $\bby$ and $\bbX$ are unrelated. 
Moreover, the bound in~\eqref{eq:gcn_error} can be reduced when the variance in $\bbX$ is sufficiently dominated by class sizes $\bbp$ and node degrees $\bbd$, reflecting the intuitive fact that nodes with high degree $d_i$ belonging to a class of large size $p_c$ are easier to predict~\citep{liu2023GeneralizedDegreeFairness,kang2022RawlsGCNRawlsianDifference}. 
Thus, Theorem~\ref{thm:gcn_error} shows that the composition of features and edges can render a feature important or unimportant to GCN performance, regardless of its relevance in the absence of the graph.
Furthermore, while Theorem~\ref{thm:gcn_error} considers GCNs that exploit homophilic labels $\bby$,  Appendix~\ref{app:gnn_performance} includes results with similar conclusions when $\bby$ is heterophilic on $\bbA$.
Therefore, we ought not to rely solely on dependencies between $\bby$ and $\bbX$ to quantify node feature importance.
% Thus, Theorem~\ref{thm:gcn_error} shows that measuring feature importance based solely on dependencies between $\bby$ and $\bbX$ may not be sufficient for GNN feature selection\ibmblue{~\citep{zheng2024WhatMissingGraph}}.
% \ibmblue{More specifically, the bound in~\eqref{eq:gcn_error} reveals that certain compositions of features and edges may render a feature important or unimportant regardless of its relevance in the absence of the graph.}

% ------------------------------------------------------------------------------------------
\renewcommand{\arraystretch}{.7}
\addtolength{\tabcolsep}{-0.2em}
\begin{table*}[t!]
\scriptsize
\centering
\caption{Node classification accuracy under various perturbations. 
(Notation: {\bf Best}, \underline{second best})
% The top performing method is \textbf{boldfaced}, and the secondmost \underline{underlined}.
% \(\tilde{X}:\)\textit{Permuted features}, 
% \(\mathcal{N}:\)\textit{Random features}, 
% \(\tilde{A}:\)\textit{Random ER graph with same number of edges.}
% 100 epochs for the first 3 datasets. 300 epochs for Photo and Computers. GCN for the first 5 datasets. TAG for the last three datasets.
}
\vspace{-.1cm}
% \label{tab:structure-feature-ablation}
\label{t:real_glob_perm}
\begin{tabular}{l|ccc|cc|ccc}
\toprule
\textbf{Setting} 
& \textbf{Cora} 
& \textbf{CiteSeer} 
& \textbf{PubMed} 
& \textbf{Photo} 
& \textbf{Computers} 
& \textbf{Cornell} 
& \textbf{Texas} 
& \textbf{Wisconsin} \\
\midrule
$\text{GNN}(\bbX;\bbA,\bbTheta)$ 
& \textbf{85.83} $\pm$ \tiny{0.46} 
& \textbf{74.38} $\pm$ \tiny{1.09} 
& \textbf{88.85} $\pm$ \tiny{0.42} 
& \textbf{94.04} $\pm$ \tiny{0.69}  
& \textbf{90.58} $\pm$ \tiny{0.79}  
& 74.59 $\pm$ \tiny{7.76}  
& \textbf{82.70} $\pm$ \tiny{4.05}  
& 82.80 $\pm$ \tiny{3.25}  \\
\midrule
$\text{MLP}(\bbX;\bbTheta)$ 
& 74.21 $\pm$ \tiny{1.40} 
& \underline{70.02} $\pm$ \tiny{1.39}
& \underline{88.65} $\pm$ \tiny{0.41}
& 88.73 $\pm$ \tiny{0.73}  
& 81.63 $\pm$ \tiny{0.75}  
& \textbf{78.92} $\pm$ \tiny{5.51}  
& \textbf{82.70} $\pm$ \tiny{2.16}  
& \textbf{83.60} $\pm$ \tiny{6.62}  \\
$\text{GNN}(\tbX;\bbA,\bbTheta)$ 
& 76.53 $\pm$ \tiny{1.12} 
& 63.34 $\pm$ \tiny{1.44} 
& 47.42 $\pm$ \tiny{3.25} 
& 66.76 $\pm$ \tiny{8.58} 
& 51.09 $\pm$ \tiny{8.84}  
& 43.78 $\pm$ \tiny{8.95}  
& 52.97 $\pm$ \tiny{6.07}  
& 46.80 $\pm$ \tiny{6.76}  \\
$\text{GNN}(\bbW;\bbA,\bbTheta)$ 
& \underline{82.92} $\pm$ \tiny{1.54}
& 67.37 $\pm$ \tiny{1.61} 
& 76.23 $\pm$ \tiny{0.53} 
& \underline{89.92} $\pm$ \tiny{0.58}
& \underline{85.49} $\pm$ \tiny{0.44}
& 49.19 $\pm$ \tiny{7.91}  
& 55.68 $\pm$ \tiny{2.76}  
& 47.20 $\pm$ \tiny{7.00}  \\
$\text{GNN}(\bbX;\tbA,\bbTheta)$ 
& 36.97 $\pm$ \tiny{1.80} 
& 35.13 $\pm$ \tiny{2.41} 
& 67.07 $\pm$ \tiny{0.80} 
& 30.52 $\pm$ \tiny{5.77} 
& 37.64 $\pm$ \tiny{0.45}  
& \underline{67.57} $\pm$ \tiny{7.83} 
& \underline{70.27} $\pm$ \tiny{7.05}  
& \underline{78.40} $\pm$ \tiny{6.37}  \\
\bottomrule
\end{tabular}
\vspace{-.4cm}
\end{table*}
\addtolength{\tabcolsep}{0.3em}
\renewcommand{\arraystretch}{1.}
% ------------------------------------------------------------------------------------------

\subsection{GNN Performance Under Perturbations}\label{Ss:real_glob_perm}

We empirically verify this intuition by comparing GNN node classification accuracy on real-world benchmark datasets under various perturbations intended to remove dependencies among $\bbX$, $\bbA$, and $\bby$.
All simulation details can be found in Appendix~\ref{app:exp_details}, which includes dataset details.
To assess the joint influence of a graph and its features, we consider (i) $\text{GNN}(\bbX;\tbA,\bbTheta)$ using an Erdos-Renyi (ER) graph with the same number of edges as $\bbA$, (ii) $\text{MLP}(\bbX;\bbTheta)$, a multilayer perceptron (MLP) that considers no graph, (iii) $\text{GNN}(\bbW;\bbA,\bbTheta)$ using Gaussian noise $\bbW \sim \ccalN(\bbzero,\bbI)$ as node features, and (iv) $\text{GNN}(\tbX;\bbA,\bbTheta)$, where $\tbX$ contains randomly permuted rows of $\bbX$.
We train GCNs for Cora, Citeseer, and PubMed~\citep{sen2008collectiveclassificationnetwork,namata2012QuerydrivenActiveSurveying} and graph isomorphism networks (GINs)~\citep{xu2018howPowerfulGraph} for Photo and Computers~\citep{mcauley2015imagebasedrecommendationsstyles,shchur2018pitfalls}, while for Cornell, Texas, and Wisconsin graphs with heterophilic labels~\citep{pei2020geomgcn}, we consider topology adaptive GCNs (TAGCNs), which can aggregate features of nodes in multi-hop neighborhoods~\citep{du2017topologyadaptivegraph}.

Table~\ref{t:real_glob_perm} demonstrates both the importance of node features along with their dependence on the graph in these datasets.
% demonstrates the importance of both node features and their dependence on the graph for each dataset.
% demonstrates both the importance of node features in these datasets along with their dependence on the associated graph.
For the first five datasets, the respectable performance of $\text{MLP}(\bbX;\bbTheta)$ and $\text{GNN}(\bbW;\bbA,\bbTheta)$ shows that both features $\bbX$ and graph structure $\bbA$ are semantically \mbox{relevant}.
We also observe particularly low accuracy for $\text{GNN}(\bbX;\tbA,\bbTheta)$, reflecting the error due to cross-class edges in~\eqref{eq:gcn_error} caused by the arbitrary connections in $\tbA$, despite the informativeness of $\bbX$.
Furthermore, if $\bbX$ experiences significant shifts across classes, then we expect $\text{GNN}(\tbX;\bbA,\bbTheta)$ with permuted features to perform worse than $\text{GNN}(\bbW;\bbA,\bbTheta)$ for sparse $\bbA$ since summands in the second term of~\eqref{eq:gcn_error} may have large $\| \bbX_{i,:} - \bbX_{j,:} \|_2$ for $Y_{ic}Y_{jc}=1$.
Indeed, $\text{GNN}(\bbW;\bbA,\bbTheta)$ outperforms $\text{GNN}(\tbX;\bbA,\bbTheta)$ for all datasets in Table~\ref{t:real_glob_perm}.
We also corroborate known challenges of graph convolutions for data with heterophilic $\bby$, as $\text{MLP}(\bbX;\bbTheta)$ rivals and can even outperform $\text{GNN}(\bbX;\bbA;\bbTheta)$ for Cornell, Texas, and Wisconsin.
For these datasets, $\text{GNN}(\bbX;\tbA,\bbTheta)$ is far superior to $\text{GNN}(\tbX;\bbA,\bbTheta)$ and $\text{GNN}(\bbW;\bbA,\bbTheta)$, which reflects the difficulty of convolving node features that are both irrelevant to labels $\bby$ and heterophilic on $\bbA$, as even random connections in $\tbA$ yield significantly higher accuracy.
% %%%%%%%%%%%%%%%%%%%%%%%%%%%%%%%%%%%%%%%%%%%%%

% %%%%%%%%%%%%%%%%%%%%%%%%%%%%%%%%%%%%%%%%%%%%%
% \section{Permutation Tests for Node Feature Importance}\label{S:permtest}
\section{Node Feature Permutation Testing}\label{S:permtest}

Inspired by Theorem~\ref{thm:gcn_error} and Table~\ref{t:real_glob_perm}, we propose \emph{\underline{n}ode feature \underline{p}ermutation \underline{t}esting (NPT)} to measure feature importance via \textit{permutation-based scores}~\citep{altmann2010Permutationimportancecorrected,khan2025UsingPermutationBasedFeature,yang2009FeatureSelectionMLP}.
Formally, let $\Pi$ be the set of permutations of $[N]$.
Then, if $\tilde{\bbX}^{(m)}$ denotes $\bbX$ with values of feature $m$ reordered according to some random $\bbpi \in \Pi$, we measure feature importance through permutation tests
\alna{
    &\nonumber\\[-.65cm]&
    \delta_m(\bby,\bbX,\tilde{\bbX}^{(m)})
    % :=
    &~~:=~~&
%     {\rm Acc}(\bby,\bbZ)
%     -
%     {\rm Acc}\big(\bby,\tbZ^{(m)} \big),
% &\nonumber\\&
%     {\rm for}~~~
%     \bbZ := f(\bbX; \bbA, \bbTheta),
%     ~~
%     \tbZ^{(m)} \! := \! f \big(\tbX^{(m)}; \bbA, \bbTheta \big)
    % {\rm Acc}(\bby,f(\bbX;\bbA,\bbTheta))
    % -
    % {\rm Acc}\big(\bby,f( \tilde{\bbX}^{(m)}; \bbA,\bbTheta) \big),
    {\rm Acc}(\bby,f(\bbX;\bbA,\bbTheta))
&\nonumber\\&
    &&
    ~~~~
    -
    {\rm Acc}\big(\bby,f( \tilde{\bbX}^{(m)}; \bbA,\bbTheta) \big),
&\label{eq:perm}\\[-.65cm]&
\nonumber}
% \label{eq:perm}}
where ${\rm Acc}(\bby,\bbZ)$ measures the accuracy of embeddings $\hby = g(\bbZ)$ for classifier $g$.
With some abuse of notation, we let $\delta_m(\bby_{\rm train},\bbX,\tilde{\bbX}^{(m)})$ denote the accuracy for the subset of nodes corresponding to observed training nodes, with analogous definitions for other subsets of nodes.
The formulation in~\eqref{eq:perm} evaluates how much model performance suffers by removing the effect of a feature, as permuting the $m$-th feature in $\bbX$ breaks its correlation with the labels $\bby$, graph structure $\bbA$, and remaining features.
% \mad{Sentence about how the purpose is to evaluate how well a model performs with and without the effect of a feature. If difference is small, then the feature appears not to be relevant, but if large, then error increased upon permuting, indicating its importance.}
Permutation tests are a classical approach to isolate the effects of a feature~\citep{brieman2001RandomForests,toth2020PermutationTestComplex,altmann2010Permutationimportancecorrected}. However, we must validate that permutation testing~\eqref{eq:perm} is valid in the presence of node dependencies. In fact, we will show that it can be particularly informative in the presence of $\bbA$.
% Permutation tests are a classical approach to isolate the effects of a feature~\citep{brieman2001RandomForests,toth2020PermutationTestComplex,altmann2010Permutationimportancecorrected}, and we next show that it can be particularly informative in the presence of $\bbA$.
% \mad{Maybe sentence instead saying "Permutation tests are a classical approach to isolate the effects of a feature~\citep{brieman2001RandomForests,toth2020PermutationTestComplex,altmann2010Permutationimportancecorrected}. However, we must validate that permutation testing~\eqref{eq:perm} is valid in the presence of node dependencies. In fact, we will show that it can be particularly informative in the presence of $\bbA$."}
To this end, we validate that permuting columns of $\bbX$ indeed decouples node features from $\bby$ and $\bbA$, which verifies that $\delta_m$ reflects feature influence for GCN predictions, supporting the results in Table~\ref{t:real_glob_perm}.

% \vspace{-.2cm}

\begin{theorem}\label{thm:perm_gcn_error}
    % \mad{To be updated for arbitrary layers}
    Consider $\tbX \in \reals^{N\times M}$ such that $\tbX_{i,:} = \bbX_{\pi(i),:}$ for all $i\in[N]$ and some permutation $\bbpi \in \Pi$ chosen randomly.
    For the GCN in~\eqref{eq:gcn_single_layer}, let $\tbZ^* := f(\tbX^*;\bbA^*,\bbTheta)$ for $\tbX^* := \bbY\bbP^{-1}\bbY^\top\tbX$, $\bbA^*$ in~\eqref{eq:ideal_A}, and $\bbDelta = \bbA - \bbA^*$.
    Furthermore, if $\alpha := \max_{m\in[M],i,j \in [N]} (X_{im} - X_{jm})^2$, then with probability at least $1-e^{-t^2 / 2 + \log M}$ for any $t \geq \sqrt{2\log M}$,
    \alna{
        &\nonumber\\[-.65cm]&
        \!\!\!\!\!
        \| \tbZ^* - \tbZ \|_F
        \leq
        % \tau^2 \omega^2
        % \!
        \tau^{L} \omega^{L}
        \Big[
            % \tau^2 \omega^2
            % \tau^{L} \omega^{L}
            (L-1)
            (1 + \sqrt{N})
            \|\bbDelta\|_F \|\bbX\|_F
        &\nonumber\\&
            \qquad\qquad\quad
            +
            % \tau^2 \omega^2
            \sqrt{\gamma}
            \big\| \vect\big(\bbY\bbP^{-1}\bbY^\top - \tilde{\bbD}^{-1}\bbA\big) \big\|_1
        \Big],
    &\nonumber\\&
        \!\!\!\!\!
        \text{where}~
        \gamma
        % \,:=\,
        :=
        % \frac{2}{N-1}
        {\textstyle\frac{2}{N-1}}
        \Big(
            \| \bbX \|_F^2
            -
            % \frac{1}{N}
            {\textstyle\frac{1}{N}}
            \| \bbX^\top \bbone \|_2^2
        \Big)
        +
        \alpha t M\sqrt{N}.
    % &\nonumber\\[-.25cm]& 
    &\label{eq:pert_gcn_error}\\[-.7cm]& \nonumber}
    % &\label{eq:pert_gcn_error}\\[-.7cm]&
    % \nonumber}
\end{theorem}

We prove Theorem~\ref{thm:perm_gcn_error} in Appendix~\ref{app:perm_gcn_error}.
The bound in~\eqref{eq:pert_gcn_error} reveals how comparing accuracy with and without permuting node features reveals the influence of $\bbX$.
More specifically, analogously to computing~\eqref{eq:perm}, we can compare GCN error with and without permutations by taking the difference in the upper bounds of~\eqref{eq:pert_gcn_error} and~\eqref{eq:gcn_error},
\alna{
    &\nonumber\\[-.6cm]&
    \tau^{L}
    \omega^{L}
    \!
    % \sum_{c=1}^C
    % \sum_{i,j=1}^N
    \sum_{c,i,j}
    \left|
        \frac{Y_{ic}Y_{jc}}{p_c}
        -
        \frac{A_{ij}}{d_i + 1}
    \right|
    \!
    \big(
        \sqrt{\gamma} - \| \bbX_{i,:} \!- \bbX_{j,:} \|_2
    \big),
    &\label{eq:error_diff}\\[-.6cm]& \nonumber}
% \label{eq:error_diff}}
which we can assess to determine when $\delta_m$ will likely be small or large.
First, we note in~\eqref{eq:pert_gcn_error} that because the heterophily of $\bby$ encoded in $\| \vect(\bbY\bbP^{-1}\bbY^\top - \tilde{\bbD}^{-1} \bbA) \|_1$ can no longer be mitigated by node feature similarities as in the original bound~\eqref{eq:gcn_error}, permuting features will likely worsen GCN performance if $\bbX$ is informative.
Indeed, if class sizes $\bbp$ and node degrees $\bbd$ are relatively similar,~\eqref{eq:error_diff} will increase if $\bbX$ has high overall variance $\gamma$ and $\| \bbX_{i,:} - \bbX_{j,:} \|_2$ is small when $Y_{ic}Y_{jc}=1$ and $A_{ij}=0$, implying that $\| \bbX_{i,:} - \bbX_{j,:} \|_2$ is large when $Y_{ic}Y_{jc}=0$ and $A_{ij} =1$.
Thus, as the number of heterophilic edges increase and homophilic edges decrease, features become more important if they show high within-class similarity and across-class dissimilarity.
However, observe that a highly homophilic $\bby$ can reduce the difference in~\eqref{eq:error_diff}, implying that features are less informative.
In fact, when $\bby$ is highly homophilic over $\bbA$, any feature can be deemed important if the features possess a high enough variance, that is, when $\gamma$ is large enough.
Conversely, we observe that features with low variance will reduce $\gamma$ and therefore~\eqref{eq:error_diff}, as expected since features that exhibit smaller differences are likely to be less informative, even if they are separable across classes.
Theorem~\ref{thm:perm_gcn_error} therefore supports evaluating GNN performance before and after permutations to quantify feature importance.
% %%%%%%%%%%%%%%%%%%%%%%%%%%%%%%%%%%%%%%%%%%%%%

%%%%%%%%%%  ALGORITHM  %%%%%%%%%%
\algsetup{indent=1em, linenosize=\small}
% \begin{figure*}[t!]
\begin{figure}[t!]
% \begin{wrapfigure}[28]{r}{0.70\textwidth}
    \centering
    \vspace{-.3cm}
    \scalebox{.84}{
    \begin{minipage}{0.55\textwidth}
    % \begin{minipage}{0.95\textwidth}
        \begin{algorithm}[H]
        \begin{algorithmic}[1]
        % \small
        \REQUIRE 
            Step size $\lambda > 0$,~~
            $T_{\rm burn}, T \in \naturals$,~~
            $K \in \naturals$,~~
            $r\in(0,1)$
        \STATE 
            Initialize
            % $\hbX = \bbX$, feature mask $\bbb \in \{0,1\}^M$, counter $t=1$.
            % \revise{feature mask $\bbb = \bbone$ for $\hbX := \bbX\diag(\bbb)$, counter $t=1$.}
            $\hbX = \bbX$, feature mask $\bbb = \bbone$, counter $t=1$.
        \WHILE{Stopping criteria not met}
            \STATE 
                Gradient update: 
                $\bbTheta \leftarrow \bbTheta - \lambda \nabla_{\bbTheta} \ccalL(\bby_{\rm train}, f(\hbX;\bbA,\bbTheta))$. 
            \STATE 
                Update $t \leftarrow t + 1$. 
            \IF{$t > \max(T,T_{\rm burn})$}
                \STATE Reset $t \leftarrow 1$. 
                \FOR{{$m \in \{ \ell ~|~ b_{\ell}=1, ~ \ell\in[M] \}$}}
                    \STATE Initialize average score $\hat{\delta}_m = 0$.
                    \FOR{$k \in [K]$}
                        \STATE Sample random permutation $\bbpi \sim \Pi$. 
                        \STATE Permute $\hbX_{:,m}$ for $\tbX^{(m)}$ such that 
                                % ~ $\tilde{X}_{im}^{(m)} = \hat{X}_{\pi(i),m}$ and $\tilde{X}_{i\ell}^{(m)} = \hat{X}_{i\ell}$ 
                                $\tilde{X}_{im}^{(m)} = \hat{X}_{\pi(i),m}$ and $\tilde{X}_{i\ell}^{(m)} = \hat{X}_{i\ell}$ 
                                ~ $\forall i\in[N],\ell\in[M]\backslash\{m\}$. 
                        \STATE Update $\hat{\delta}_m \leftarrow \hat{\delta}_m + \frac{1}{K}\delta_m(\bby_{\rm val},\hbX,\tbX^{(m)})$ via~\eqref{eq:perm}.
                    \ENDFOR
        %             % \If{\red{$\hat{\delta}_m \leq 0$}}{
        %             %     Prune unimportant feature $b_m \leftarrow 0$
        %             % }
                \ENDFOR
            \ENDIF
            \STATE Compute $r$-quantile $\delta^{(r)}$ from $\big\{ \hat{\delta}_m ~|~ b_m=1, m\in[M] \big\}$.
            \FOR{$m \in \{ \ell ~|~ b_{\ell}=1, ~ \ell\in[M] \}$}
                \IF{$ \hat{\delta}_m < \delta^{(r)} $}
                    \STATE Prune unimportant feature $b_m \leftarrow 0$, $\hbX \leftarrow \hbX \diag(\bbb)$.
                    % ~~ $\hat{\bbX}_{:,m} \leftarrow b_m \hat{\bbX}_{:,m}$.
                \ENDIF
            \ENDFOR
        \ENDWHILE
        % \RETURN Model $f(\cdot;\bbA;\bbTheta)$, pruned features $\hbX$, mask $\bbb$, scores $\hbdelta$
        \RETURN Model $f(\cdot;\bbA;\bbTheta)$, pruned features $\hbX$, scores $\hbdelta$
        % \caption{\small Adaptive node feature selection via NPT.}\label{alg:nfpt}
        \caption{Adaptive node feature selection via NPT.}\label{alg:nfpt}
        \end{algorithmic}
        \end{algorithm}
    \end{minipage}
    }
% 
% \end{wrapfigure}
\vspace{-0.4cm}
% \end{figure*}
\end{figure}
%%%%%%%%%%  END ALGORITHM  %%%%%%%%%%

% %%%%%%%%%%%%%%%%%%%%%%%%%%%%%%%%%%%%%%%%%%%%%
\subsection{Adaptive Node Feature Selection}\label{Ss:method}

Comparing Theorems~\ref{thm:gcn_error} and~\ref{thm:perm_gcn_error} in~\eqref{eq:error_diff} reveals the value of permutation tests for node feature importance.
More specifically, we found that NPT in~\eqref{eq:perm} can identify important features for node classification accuracy, even if the feature is not relevant without the graph structure $\bbA$, and vice versa.
Because~\eqref{eq:perm} is agnostic to the model, data, and task, it is particularly suited to an adaptive approach to evaluate node feature importance, eliminating the need to measure feature relevance as a pre-processing step.
Thus, we propose an adaptive feature selection method in Algorithm~\ref{alg:nfpt} to identify and remove unnecessary features during training.
% Given the value of permutation tests for node feature importance, we propose an adaptive feature selection method in Algorithm~\ref{alg:nfpt} to identify and remove unnecessary features during training.

The matrix $\hbX$ in Algorithm~\ref{alg:nfpt} denotes the pruned feature matrix with masked columns corresponding to $\bbb \in \{0,1\}^M$, representing selected features.
After the model $f$ has been trained for $T_{\rm burn}$ epochs, we periodically compute the empirical average $\hat{\delta}_m$ of the NPT importance score $\delta_m(\bby_{\rm val},\hbX,\tilde{\bbX}^{(m)})$ for every $m \in [M]$ over $K$ random permutations $\bbpi \in \Pi$ (lines 7-15).
We then keep the top $r$-th percentile of features based on $\hbdelta$ by setting $b_m = 0$ for the remaining ones.
% \santi{Somewhere early in the paper you talk about not having to drop a proportion of the features but rather those that have negative delta in accuracy ... but here you talk about dropping based on percentile. These seem to be at odds. I understand that you can do both, but at the very least I would mention that somewhere here to make it consistent with the previous mention.}
% \red{If any feature is deemed unhelpful $\delta_m \leq 0$, we eliminate it from training by setting $b_m = 0$ in lines 16-17.}
We then continue training to update model parameters given the new subset of features.
Algorithm~\ref{alg:nfpt} thus yields a \emph{single} process to both train a GNN $f$ and successively prune unnecessary features.
The most complex step of Algorithm~\ref{alg:nfpt} occurs at the first checkpoint when $t = T+1$, where all $M$ features must be permuted $K$ times, resulting in $O(KNM)$.
However, at $nT+1$ for $n > 1$, we need only permute $r^n M < M$ features, so we may choose $r \in (0,1)$ with no cost to theoretical complexity.
% however, even for only the first checkpoint, measuring the scores of every feature via uniform permutation can be computationally prohibitive for very high-dimensional features. while we 
Nevertheless, to mitigate costs for very high-dimensional features, Algorithm~\ref{alg:nfpt} can accommodate more efficient albeit less statistically grounded metrics, such as importance sampling variance reduction techniques.
In fact, our theoretical results still hold as long as the alternative approach yields the same mean and covariance as node feature permutations.

% \revise{While the dominant cost of Algorithm~\ref{alg:nfpt} lies in this first checkpoint, this overhead can be mitigated through scalable approximations of the permutation step.
% In particular, importance sampling or variance reduction techniques can be used to reduce the effective number of permutations $K$ required to obtain reliable importance estimates.}
% We visualize permutation time empirically in Figure~\ref{f:ablation}, which we discuss further in Section~\ref{S:sim}.
% by comparing the time to permute node features for each dataset at each checkpoint $nT$ for $n\in\naturals$.

Besides its convenience, another critical advantage of Algorithm~\ref{alg:nfpt} is its flexibility.
% Other advantages of Algorithm~\ref{alg:nfpt} include flexibility to graph data, architecture choice, and more.
Indeed, since the metric $\delta_m$ is defined by changes in performance, we may replace accuracy ${\rm Acc}$ in~\eqref{eq:perm} with any quality to which features ought to contribute, such as promoting fairness~\citep{little2024FairFeatureImportance,navarro2024fairglasso,navarro2024mitigating}.
Therefore, with \emph{no prior assumptions on the graph} and \emph{no restrictions on architecture}~\citep{maurya2022Simplifyingapproachnode,maurya2023FeatureselectionKey}, the model $f$ adapts to the learning task by the definition of $\delta_m$.
% Thus, the model $f$ adapts to the learning task by the definition of $\delta_m$ without requiring prior assumptions on the graph, nor are we restricted to particular architectures~\citep{maurya2022Simplifyingapproachnode,maurya2023FeatureselectionKey}. 
As we show in Section~\ref{S:sim}, this means that Algorithm~\ref{alg:nfpt} is amenable to multiple scenarios, including heterophilic labels $\bby$ or features $\bbX$.
Furthermore, observe that we can compute line 14 using \emph{any} feature importance score that may be particularly suited to the task.
Thus, our algorithm constitutes a flexible approach that can be completely data agnostic or tailored to the task given prior knowledge.
Crucially, many prior graph-based metrics do not account for model behavior~\citep{mahmoud2023Nodeclassificationgraph,zheng2025letyourfeatures}, whereas $\delta_m$ explicitly aims to promote the accuracy of $f$, making permutation tests an advantageous general choice, which we illustrate empirically.
% since $\delta_m$ explicitly aims to promote the accuracy of $f$.
% whereas $\delta_m$ explicitly aims to promote the accuracy of $f$, rendering it an appropriate general choice.
% while we espouse permutation tests due to our results in Theorems~\ref{thm:gcn_error} and~\ref{thm:perm_gcn_error},

% % Finally, observe that line line 14 may be computed using any feature importance score, as another may be particularly suited to the task given prior knowledge.
% Finally, while we espouse permutation tests due to our results in Theorems~\ref{thm:gcn_error} and~\ref{thm:perm_gcn_error}, line 14 may be computed using any feature importance score, as another may be particularly suited to the task given prior knowledge.
% However, many prior graph-based metrics do not account for model behavior~\citep{mahmoud2023Nodeclassificationgraph,zheng2025letyourfeatures}, whereas $\delta_m$ explicitly aims to promote the accuracy of $f$, rendering it an appropriate general choice.

% %%%%%%%%%%%%%%%%%%%%%%%%%%%%%%%%%%%%%%%%%%%%%

% %%%%%%%%%%%%%%%%%%%%%%%%%%%%%%%%%%%%%%%%%%%%%
\section{Numerical experiments}\label{S:sim}

We next evaluate our importance scores and algorithm based on node feature permutation tests.
We consider the same datasets and architectures as in Table~\ref{t:real_glob_perm}, with minimal details explained below.
Dataset statistics, along with other dataset details, are included in Appendix~\ref{app:exp_details}.
Code to reproduce all experiments is available at {\url{https://github.com/madnavarro/node_feature_importance}}.

% \medskip 
% \smallskip
\vspace{-.35cm}

% \textbf{Datasets.}
% \vspace{-.1cm}
\begin{itemize}[left= 2pt .. 10pt, noitemsep]
    \item \textbf{Citation networks:}
        Cora, Citeseer, and PubMed consist of papers as nodes, connected based on citations~\citep{sen2008collectiveclassificationnetwork,namata2012QuerydrivenActiveSurveying}. 
        The goal is to predict paper topic $\bby$ from bag-of-words paper embeddings $\bbX$.
        % We additionally evaluate on the large-scale ogbn-arxiv dataset from the Open Graph Benchmark~\citep{NEURIPS2020_fb60d411}, where nodes represent arXiv CS papers connected by citation links. 
        % Each node is assigned one of 40 subject-area classes, and input features \(\bbX\) correspond to averaged word embeddings of the paper title and abstract.
    \item \textbf{Co-purchase graphs:}
        Photo and Computers represent Amazon goods as nodes that are connected if frequently purchased together, with $\bby$ as product category and $\bbX$ as word embeddings of product reviews~\citep{mcauley2015imagebasedrecommendationsstyles,shchur2018pitfalls}.
    \item \textbf{Webpage graphs:}
        Cornell, Texas, and Wisconsin connect linked webpages of individuals in computer science departments across various universities~\citep{pei2020geomgcn}.
        Labels $\bby$ represent the role of individuals to be predicted from webpage word embeddings $\bbX$.    
\end{itemize}

% \medskip 
% \smallskip
\vspace{-.35cm}

\textbf{Architectures.}
Cora, Citeseer, and PubMed are trained with GCNs~\citep{kipf2017SemisupervisedClassificationGraph}, whereas we use TAGCNs for Cornell, Texas, and Wisconsin~\citep{du2017topologyadaptivegraph}.
As for Photo and Computers, frequently co-purchased items likely indicate similar product categories, so $\bby$ is homophilic on $\bbA$.
However, while reviews contain valuable keywords for prediction, positive and negative reviews may contain different words despite products belonging in the same category.
Thus, since features $\bbX$ may exhibit both homophily and heterophily, we employ a GIN model, which can extract complex interactions of features~\citep{xu2018howPowerfulGraph}.

% \medskip 
% \smallskip
\vspace{-.15cm}
% \vspace{-.05cm}

\textbf{Metrics.}
% \santi{You should probably introduce this name and acronym earlier in the paper. Well, I see the idea of introducint the bolded acronym here. I would maybe mention it in the method section and then repeat here with this bolding, but only once (see next comment)}
We compare our {\bf NPT} scores to alternative feature importance metrics, listed in Appendix~\ref{app:table2_info}.
% These include both graph-agnostic and graph-specific measurements.
% The full list can be found in Appendix~\ref{app:table2_info}.

% %%%%%%%%%%%%%%%%%%%%%%%%%%%%%%%%%%%%%%%%%%%%%

% ---------------------------------------
\begin{figure*}[b!]
    \centering
    % --------------------------------
    \vspace{-.3cm}
    \scalebox{.92}{\includegraphics[width=\textwidth]{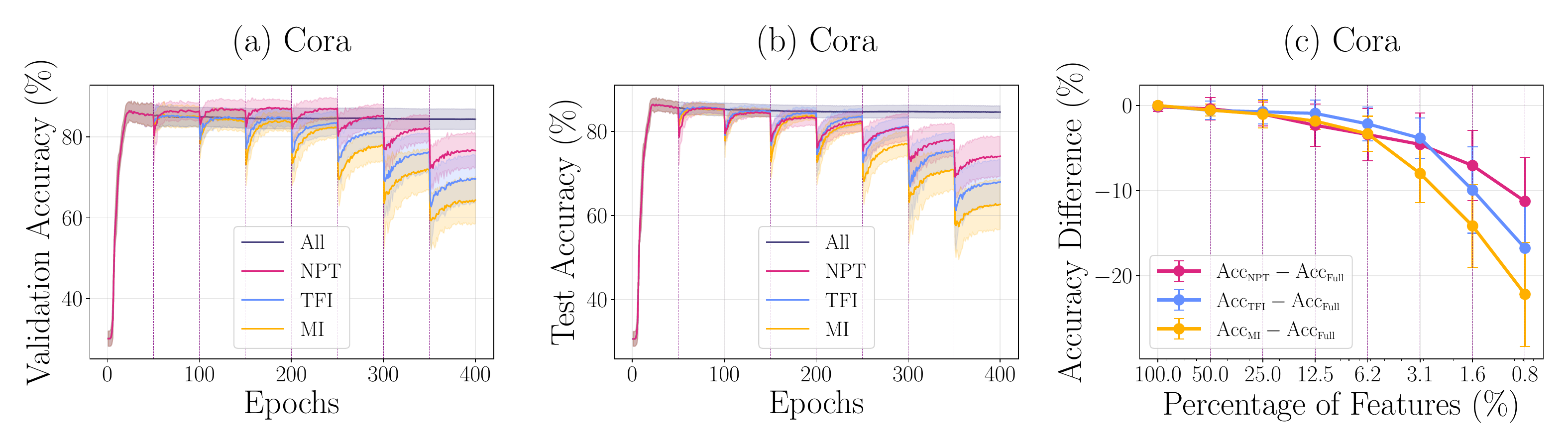}}
    \vspace{-.3cm}
    % \includegraphics[width=\textwidth]{img/Plots_Rebuttal/Fig2_AccCurves_Cora.pdf}
    % \vspace{-.7cm}
    % --------------------------------
    \caption{
        Node classification accuracy during GCN training for Cora using Algorithm~\ref{alg:nfpt} with different feature importance metrics.
        (a) Validation accuracy for models trained using all features versus {\bf NPT}, {\bf TFI}, and {\bf MI}.
        (b) Test accuracy for models trained using all features versus {\bf NPT}, {\bf TFI}, and {\bf MI}.
        (c) The difference in test accuracy between the full model and the model trained with Algorithm~\ref{alg:nfpt}.
    \label{f:cora_algo}
    }
    \vspace{-.25cm}
\end{figure*}
% ---------------------------------------

\renewcommand{\arraystretch}{.7}
\addtolength{\tabcolsep}{-0.1em}
\begin{table*}[t!]
\scriptsize
\centering
% \vspace{-.1cm}
\caption{Node classification accuracy with feature selection. 
% The top performing method is \textbf{boldfaced}, and the secondmost \underline{underlined}.
(Notation: {\bf Best}, \underline{second best})
}
\vspace{-.1cm}
% \caption{Node classification accuracy on real-world datasets with feature selection. Using only the top $2\%$ ($5\%$ for PubMed, Photo, and Computers) important features detected by each method.}
\label{t:real_fs}
\begin{tabular}{l|ccc|cc|ccc}
\toprule
\textbf{Method} 
& \textbf{Cora} 
& \textbf{CiteSeer} 
& \textbf{PubMed} 
& \textbf{Photo} 
& \textbf{Computers} 
& \textbf{Cornell} 
& \textbf{Texas} 
& \textbf{Wisconsin} \\
\midrule
All features 
& {85.83} $\pm$ \tiny{0.46} 
& {74.38} $\pm$ \tiny{1.09} 
& {88.85} $\pm$ \tiny{0.42} 
& {94.04} $\pm$ \tiny{0.69}  
& {90.58} $\pm$ \tiny{0.79}  
& 74.59 $\pm$ \tiny{7.76}  
& {82.70} $\pm$ \tiny{4.05}  
& 82.80 $\pm$ \tiny{3.25}   \\
\midrule
$\text{NPT}$ 
& \textbf{79.19} $\pm$ \tiny{2.45} 
& \textbf{69.35} $\pm$ \tiny{1.49} 
& \textbf{87.11} $\pm$ \tiny{0.75} 
& \textbf{93.59} $\pm$ \tiny{0.79} 
& \underline{90.09} $\pm$ \tiny{0.51} 
& \textbf{69.73} $\pm$ \tiny{6.26} 
& \textbf{72.97} $\pm$ \tiny{9.21} 
& \textbf{73.20} $\pm$ \tiny{6.88}  \\
$\text{NPT-gaussian}$ 
& \underline{78.86} $\pm$ \tiny{1.42} 
& \underline{68.48} $\pm$ \tiny{1.23} 
& \underline{86.69} $\pm$ \tiny{0.72} 
& 93.49 $\pm$ \tiny{0.67} 
& 89.85 $\pm$ \tiny{0.17} 
& \underline{65.41} $\pm$ \tiny{9.76} 
& \underline{68.65} $\pm$ \tiny{8.48} 
& 68.80 $\pm$ \tiny{8.26}  \\
$\text{NPT-mask}$ 
& {76.05} $\pm$ \tiny{1.08} 
& {68.12} $\pm$ \tiny{1.69} 
& {86.11} $\pm$ \tiny{0.82} 
& 93.48 $\pm$ \tiny{0.59} 
& 89.94 $\pm$ \tiny{0.22} 
& 63.24 $\pm$ \tiny{4.39} 
& 64.86 $\pm$ \tiny{5.13} 
& \underline{72.40} $\pm$ \tiny{7.31}  \\
TFI
& 72.73 $\pm$ \tiny{5.47} 
& 65.77 $\pm$ \tiny{2.04} 
& 83.80 $\pm$ \tiny{0.92} 
& 93.02 $\pm$ \tiny{0.68} 
& \underline{90.09} $\pm$ \tiny{0.23} 
& 61.62 $\pm$ \tiny{7.13} 
& 61.08 $\pm$ \tiny{4.39} 
& 52.40 $\pm$ \tiny{3.44}  \\
MI 
& 66.83 $\pm$ \tiny{3.68} 
& 63.79 $\pm$ \tiny{1.02} 
& 85.96 $\pm$ \tiny{1.00} 
& \underline{93.56} $\pm$ \tiny{0.61} 
& \textbf{90.33} $\pm$ \tiny{0.32} 
& {63.78} $\pm$ \tiny{5.82} 
& {65.41} $\pm$ \tiny{7.33} 
& {69.60} $\pm$ \tiny{5.99}  \\
$h_{\rm attr}$
& 39.96 $\pm$ \tiny{1.00} 
& 22.59 $\pm$ \tiny{1.01} 
& 78.85 $\pm$ \tiny{0.21} 
& 93.53 $\pm$ \tiny{0.46} 
& \underline{90.09} $\pm$ \tiny{0.46} 
& 55.14 $\pm$ \tiny{5.01} 
& 58.38 $\pm$ \tiny{5.01} 
& 45.60 $\pm$ \tiny{6.62}  \\
$h_{\rm Euc}$
& 32.77 $\pm$ \tiny{1.66} 
& 22.62 $\pm$ \tiny{1.03} 
& 74.37 $\pm$ \tiny{0.60} 
& \textbf{93.59} $\pm$ \tiny{0.50} 
& 89.19 $\pm$ \tiny{0.51} 
& 52.43 $\pm$ \tiny{4.05} 
& 57.30 $\pm$ \tiny{3.15} 
& 44.00 $\pm$ \tiny{7.48}  \\
$h_{\rm GE}$
& 31.44 $\pm$ \tiny{1.35} 
& 22.47 $\pm$ \tiny{1.01} 
& 70.52 $\pm$ \tiny{0.55} 
& 93.41 $\pm$ \tiny{0.46} 
& 89.24 $\pm$ \tiny{0.25} 
& 52.43 $\pm$ \tiny{4.05} 
& 57.30 $\pm$ \tiny{3.15} 
& 44.00 $\pm$ \tiny{7.48}  \\
Rnd.
& 39.76 $\pm$ \tiny{1.22} 
& 34.39 $\pm$ \tiny{4.07} 
& 70.71 $\pm$ \tiny{1.08} 
& 91.99 $\pm$ \tiny{0.49} 
& 88.27 $\pm$ \tiny{0.26} 
& 56.65 $\pm$ \tiny{4.69} 
& 58.49 $\pm$ \tiny{3.23} 
& 57.04 $\pm$ \tiny{4.49}  \\
\bottomrule
\end{tabular}
\vspace{-.3cm}
\end{table*}
\addtolength{\tabcolsep}{0.2em}
\renewcommand{\arraystretch}{1.}

% %%%%%%%%%%%%%%%%%%%%%%%%%%%%%%%%%%%%%%%%%%%%%
\subsection{Node feature selection comparison}\label{Ss:featsel}

To validate our importance metric $\delta_m$ in~\eqref{eq:perm}, we train a GNN with the full set of features per dataset in Table~\ref{t:real_fs}, which we compare to GNNs trained with a subset of features selected based on \textbf{NPT} and other feature selection baselines in Appendix~\ref{app:table2_info}.
For each method, we select the top $r\%$ of features ranked by the importance metric and retrain the GNN using only these features, with $r = 5\%$ for PubMed, Photo, and Computers and $r = 2\%$ otherwise.
% \mad{Something like ``This is analogous to $n=1$'' or whatever it needs to be}
In terms of Algorithm~\ref{alg:nfpt}, this is equivalent to one-shot feature selection, where we first train a GNN on the full feature set for $T_{\rm burn}$ epochs, apply feature selection once using the resulting importance scores, and then continue training the model on the reduced feature set for an additional $T_{\rm burn}$ epochs.
Thus, $r$ denotes the fraction of features retained after this \textit{single} pruning step rather than a recurring per-checkpoint pruning rate, which we provide in Section~\ref{sec:adaptive}.

We also introduce two variants of {\bf NPT} included in Table~\ref{t:real_fs}.
Instead of permuting columns of $\bbX$, {\bf NPT-mask} masks a feature by setting its values to zero.
Masking is straightforward and can be suitable when zero is a naturally occurring value, such as for binary features, but in general, it likely will not preserve the original feature distribution.
Hence, we also include {\bf NPT-gaussian}, which replaces a feature with i.i.d. Gaussian noise matching its empirical mean and variance.
However, of the three variants, only permuting via {\bf NPT} preserves the full marginal distribution of the feature.

% For the graphs with homophilic $\bby$ (Cora, Citeseer, and PubMed), both {\bf NPT} and its masking variant {\bf NPT-mask} outperform the rest, even the GCN-specific {\bf TFI}.
% This aligns with our expectations from Theorems~\ref{thm:gcn_error} and~\ref{thm:perm_gcn_error}.
% We noted that GCNs exhibit lower error for features that are informative of classes, which was confirmed in Table~\ref{t:real_glob_perm} for these datasets.
% Moreover, Theorem~\ref{thm:perm_gcn_error} and Table~\ref{t:real_fs} showed that permuting informative features will likely increase GCN error more, but reducing feature variance may not, hence {\bf NPT} consistently outperforming {\bf NPT-mask} .

We observe that, in all cases, {\bf NPT} achieves either the highest or among the highest accuracy compared to other importance scores, even though the top baseline differs across datasets.
For the graphs with homophilic $\bby$ (Cora, Citeseer, and PubMed), all three variants of {\bf NPT} outperform the rest, even the GCN-specific {\bf TFI}.
This aligns with our expectations from Theorems~\ref{thm:gcn_error} and~\ref{thm:perm_gcn_error}.
We noted that GCNs exhibit lower error for features that are informative of classes, which was confirmed in Table~\ref{t:real_glob_perm} for these datasets. 
Moreover, Theorem~\ref{thm:perm_gcn_error} and Table~\ref{t:real_fs} showed that permuting informative features will likely increase GCN error more, but reducing feature variance may not, hence {\bf NPT} and {\bf NPT-gaussian} consistently outperforming {\bf NPT-mask} for homophilic datasets.

We similarly find that {\bf NPT} performs best for the graphs Cornell, Texas, and Wisconsin with heterophilic labels, while {\bf NPT-mask} and {\bf NPT-gaussian} worsen in performance.
Again, this follows our intuition from~\eqref{eq:gcn_error}, which shows that for GCNs, reducing the variance for heterophilic features may actually decrease error, so masking features that are relevant to $\bby$ by setting them to zero may underestimate their importance for settings of heterophily.
More specifically,~\eqref{eq:error_diff} indicates that a low variance may result in smaller differences in GCN performance before and after permuting features, therefore even if a feature is correlated with $\bby$, reducing its variance by masking may yield an erroneously small $\delta_m$.
On the contrary, \textbf{MI} performs second best, outperforming even graph-specific metrics.

Finally, for Photo and Computers, we observe smaller performance gaps across all methods, even relative to randomly chosen features \textbf{Rnd}, as expected since Table~\ref{t:real_glob_perm} indicated that these labels are highly homophilic.
We observe {\bf NPT} rivaling the top performing baselines, homophily metrics and {\bf MI}.
Indeed, review text can contain informative features such as keywords related to product category $\bby$ as well as words unrelated to $\bby$ such as sentiment.
Product-related keywords are also likely to be homophilic, hence the improved performance of homophily-based scores for these two datasets.
Similarly, {\bf MI} ignores graph structure and thus may be better able to identify review content that is unrelated to $\bby$ and $\bbA$.
Thus, in all cases, {\bf NPT} performs as well as or better than the top performing baseline: {\bf TFI} for homophilic $\bby$, {\bf MI} for heterophilic $\bby$, and homophily metrics and {\bf MI} for co-purchase graphs.
This generality comes without any prior assumptions, despite the top baselines being fundamentally different across scenarios.

% %%%%%%%%%%%%%%%%%%%%%%%%%%%%%%%%%%%%%%%%%%%%%

% %%%%%%%%%%%%%%%%%%%%%%%%%%%%%%%%%%%%%%%%%%%%%
\subsection{Adaptive node feature selection}\label{Ss:conv}
\label{sec:adaptive}
Next, we assess our adaptive node feature selection approach in Algorithm~\ref{alg:nfpt}, training GNNs while also dropping less important features.
To evaluate the tradeoff between maintaining performance and improving model efficiency, we compare Algorithm~\ref{alg:nfpt} to using the full dataset as the model is trained.
Every $T = 50$ epochs, we drop $50\%$ ($r=0.5$) of features based on the scores $\delta_m$ in~\eqref{eq:perm}.
We also apply tailored versions of Algorithm~\ref{alg:nfpt} with \textbf{TFI} and \textbf{MI} in place of $\delta_m$ in line 14.
Figures~\ref{f:cora_algo}a,b depict accuracy during training to show how models perform for the same number of features.
For each checkpoint, we measure the difference between test accuracy using feature selection and the full dataset, shown in Figure~\ref{f:cora_algo}c.

We present results for Cora in Figure~\ref{f:cora_algo}, with results for the remaining datasets in Appendix~\ref{app:data_info}.
In Figure~\ref{f:cora_algo}, we observe that {\bf NPT} preserves higher accuracy than {\bf MI} and even the GCN-specific {\bf TFI} at low $r$, mirroring results in Table~\ref{t:real_fs}.
We also observe smaller drops in accuracy for {\bf NPT} as features are eliminated, as expected since our method adapts to GNN performance, allowing the model to focus on the importance of only the remaining features.
For the other datasets, analogous figures of accuracy during training can be found in Appendix~\ref{app:adapt_plots}, while a subset of the results are shared in Table~\ref{t:adaptive_fs}.
This includes training GraphSAGE models~\citep{hamilton2017InductiveRepresentationLearning} on the ArXiv citation network from the Open Graph Benchmark~\citep{NEURIPS2020_fb60d411} with word embeddings as features and paper subjects as labels, for which we use $r = 0.4$.
We similarly find {\bf NPT} to demonstrate a competitive or superior ability to identify the most relevant features during training.
Thus, given prior information, we could select {\bf MI}, {\bf TFI}, or any other feature importance metric for our proposed, adaptive Algorithm~\ref{alg:nfpt}, but in real-world scenarios, it is likely that we will not know which metric is appropriate.
% ought to perform better.
In these cases, {\bf NPT} provides an effective approach for general scenarios due to its adaptability to the learning task.
% but if we do not know which ought to perform better, then 

% We find similar comparisons of accuracy during training for the remaining datasets with {\bf NPT} consistently demonstrating a competitive or superior ability to identify the most relevant features.
% Analogous figures of accuracy during training can be found in Appendix~\ref{app:adapt_plots}, while a subset of the results are shared in Table~\ref{t:adaptive_fs}.
% We find {\bf NPT} effective for selecting important features during training, while {\bf MI} is competitive for Cornell, Texas, and Wisconsin, similarly to Table~\ref{t:real_fs}.
% Moreover, {\bf TFI} and {\bf MI} appear to be effective importance metrics in our algorithm for Photo and Computers, aligning with our intuition about these datasets.
% Thus, with prior information, our algorithm can be further improved with an appropriate choice of metric, while permutation-based tests remain effective for general scenarios.
% %%%%%%%%%%%%%%%%%%%%%%%%%%%%%%%%%%%%%%%%%%%%%

\renewcommand{\arraystretch}{.7}
\addtolength{\tabcolsep}{-0.3em}
\begin{table*}[t!]
\scriptsize
\centering
% \vspace{-.2cm}
\caption{Node classification accuracy with adaptive feature selection via Algorithm~\ref{alg:nfpt}. 
(Notation: {\bf Best per ratio}, ``\textendash''~denotes OOM)
% The top performing method per ratio is \textbf{boldfaced}.
% , and the second best is \underline{underlined}.
% \(\tilde{X}:\)\textit{Permuted features}, 
% \(\mathcal{N}:\)\textit{Random features}, 
% \(\tilde{A}:\)\textit{Random ER graph with same number of edges.}
% 100 epochs for the first 3 datasets. 300 epochs for Photo and Computers. GCN for the first 5 datasets. TAG for the last three datasets.
}
\vspace{-.1cm}
% \label{tab:structure-feature-ablation}
\label{t:adaptive_fs}
\begin{tabular}{cc|ccc|ccc|ccc}
\toprule
\textbf{$\%$} 
& \textbf{\tiny Method}
& \textbf{\tiny Cora} 
& \textbf{\tiny CiteSeer} 
& \textbf{\tiny PubMed} 
& \textbf{\tiny Photo} 
& \textbf{\tiny Computers} 
& \textbf{\tiny ArXiv} 
& \textbf{\tiny Cornell} 
& \textbf{\tiny Texas} 
& \textbf{\tiny Wisconsin} \\
\midrule
\multirow{3}{*}{\tiny $6.25$}
& {\tiny \bf NPT}
& {\bf 82.47} \tiny{$\pm$ 1.68}
& {\bf 71.82} \tiny{$\pm$ 1.48}
& {\bf 87.06} \tiny{$\pm$ 0.89}
& 83.54 \tiny{$\pm$ 5.12}
& 81.00 \tiny{$\pm$ 2.42}
& {\bf 40.84} \tiny{$\pm$ 0.44}
& 63.78 \tiny{$\pm$ 3.67}
& {\bf 72.43} \tiny{$\pm$ 5.51}
& 74.00 \tiny{$\pm$ 6.07} \\
& {\tiny \bf TFI}
& 81.40 \tiny{$\pm$ 1.47}
& 70.02 \tiny{$\pm$ 2.04}
& 84.25 \tiny{$\pm$ 1.34}
& {\bf 91.95} \tiny{$\pm$ 1.10}
& {\bf 84.47} \tiny{$\pm$ 1.73}
& $-$          
& 63.24 \tiny{$\pm$ 8.48}
& 61.08 \tiny{$\pm$ 6.96}
& 64.40 \tiny{$\pm$ 6.86} \\
& {\tiny \bf MI}
& 78.23 \tiny{$\pm$ 0.96}
& 68.66 \tiny{$\pm$ 1.98}
& 86.48 \tiny{$\pm$ 1.10}
& 91.06 \tiny{$\pm$ 1.06}
& 83.82 \tiny{$\pm$ 3.15}
& 40.26 \tiny{$\pm$ 0.18}
& {\bf 66.49} \tiny{$\pm$ 3.67}
& 69.19 \tiny{$\pm$ 3.67}
& {\bf 78.00} \tiny{$\pm$ 1.79} \\
\midrule
\multirow{3}{*}{\tiny $3.13$}
& {\tiny \bf NPT}
& {\bf 81.88} \tiny{$\pm$ 2.65}
& {\bf 70.14} \tiny{$\pm$ 1.50}
& {\bf 86.51} \tiny{$\pm$ 0.84}
& 89.12 \tiny{$\pm$ 2.29}
& 86.92 \tiny{$\pm$ 1.61}
& {\bf 35.54} \tiny{$\pm$ 0.63}
& {\bf 67.57} \tiny{$\pm$ 4.52}
& 69.73 \tiny{$\pm$ 8.44}
& {\bf 69.60} \tiny{$\pm$ 4.96} \\
& {\tiny \bf TFI}
& 77.60 \tiny{$\pm$ 1.13}
& 68.30 \tiny{$\pm$ 1.58}
& 81.56 \tiny{$\pm$ 1.66}
& {\bf 93.05} \tiny{$\pm$ 0.61}
& {\bf 88.27} \tiny{$\pm$ 0.74}
& $-$
& 63.24 \tiny{$\pm$ 9.61}
& 61.08 \tiny{$\pm$ 5.57}
& 58.00 \tiny{$\pm$ 2.19} \\
& {\tiny \bf MI}
& 71.88 \tiny{$\pm$ 1.48}
& 65.11 \tiny{$\pm$ 1.87}
& 85.16 \tiny{$\pm$ 1.01}
& 92.39 \tiny{$\pm$ 0.97}
& 87.67 \tiny{$\pm$ 1.65}
& 35.30 \tiny{$\pm$ 0.10}
& 63.24 \tiny{$\pm$ 4.05}
& {\bf 70.81} \tiny{$\pm$ 6.92}
& 68.80 \tiny{$\pm$ 5.74} \\
\midrule
\multirow{3}{*}{\tiny $1.56$}
& {\tiny \bf NPT}
& {\bf 78.52} \tiny{$\pm$ 2.17}
& {\bf 69.08} \tiny{$\pm$ 2.26}
& {\bf 84.88} \tiny{$\pm$ 0.69}
& 88.71 \tiny{$\pm$ 1.16}
& 80.92 \tiny{$\pm$ 4.33}
& {\bf 30.58} \tiny{$\pm$ 0.49}
& {\bf 67.57} \tiny{$\pm$ 3.82}
& {\bf 70.27} \tiny{$\pm$ 4.19}
& {\bf 68.80} \tiny{$\pm$ 5.46} \\
& {\tiny \bf TFI}
& 71.73 \tiny{$\pm$ 4.72}
& 65.02 \tiny{$\pm$ 1.82}
& 79.38 \tiny{$\pm$ 0.33}
& 91.41 \tiny{$\pm$ 0.94}
& 86.65 \tiny{$\pm$ 1.07}
& $-$          
& 55.68 \tiny{$\pm$ 6.30}
& 60.54 \tiny{$\pm$ 3.67}
& 49.20 \tiny{$\pm$ 5.46} \\
& {\tiny \bf MI}
& 63.51 \tiny{$\pm$ 3.43}
& 62.17 \tiny{$\pm$ 0.49}
& 83.41 \tiny{$\pm$ 0.39}
& {\bf 92.14} \tiny{$\pm$ 0.55}
& {\bf 86.87} \tiny{$\pm$ 1.57}
& 28.79 \tiny{$\pm$ 1.32}
& 60.54 \tiny{$\pm$ 8.65}
& 61.08 \tiny{$\pm$ 10.76}
& 66.00 \tiny{$\pm$ 5.93} \\
\bottomrule
\end{tabular}
\vspace{-.3cm}
\end{table*}
\addtolength{\tabcolsep}{0.4em}
\renewcommand{\arraystretch}{1.}

\begin{figure*}[b!]
    \centering
    % --------------------------------
    \vspace{-.1cm}
    \scalebox{.93}{\includegraphics[width=\textwidth]{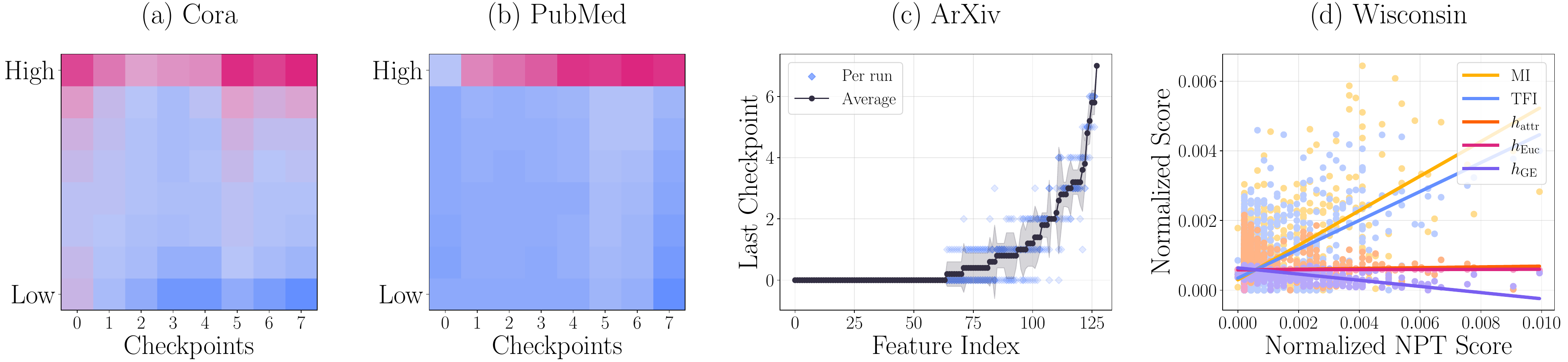}}
    % --------------------------------
    \vspace{-.05cm}
    \caption{
        Analysis of feature importance scores obtained from Algorithm~\ref{alg:nfpt}.
        % (a,b) Heatmaps of feature importance scores $\delta_m$ (high $\delta_m$ is {\color{darkpink}red} and low $\delta_m$ is {\color{darkblue}blue}).
        (a) Heatmap of feature importance $\delta_m$ during training for a GCN trained on Cora (high $\delta_m$ is {\color{darkpink}red} and low $\delta_m$ is {\color{darkblue}blue}).
        (b) Heatmap of feature importance $\delta_m$ during training for a GCN trained on PubMed.
        (c) Last checkpoint each feature is kept before dropping for GraphSAGE trained on ArXiv.
        (d) Normalized importance scores per baseline versus normalized NPT scores for a TAGCN trained on Wisconsin.
        % Heatmap of average last checkpoint features are kept for each feature ranking, that is, features are grouped by most frequent last checkpoint across 20 independent runs.
        % NPT feature importance analysis of various datasets.
        % (a) Heat map of feature importance during training for Cora.
        % (b) Heat map of feature importance during training for PubMed.
        % (c) Last checkpoint kept during training for each feature in ArXiv.
        % (d) Importance scores via baselines versus NPT importance scores for Wisconsin.
    }
    \label{f:heatmaps}
    \vspace{-.25cm}
\end{figure*}

% %%%%%%%%%%%%%%%%%%%%%%%%%%%%%%%%%%%%%%%%%%%%%
\subsection{Feature importance analysis}\label{Ss:heatmaps}

We demonstrate our ability to dynamically track feature relevance during training, confirming that features are properly dropped before the model is fully trained.
We exemplify periodically monitoring the scores $\delta_m$ in~\eqref{eq:perm} for Cora and PubMed in Figure~\ref{f:heatmaps}a,b. 
At each checkpoint ($50$ epochs), we compute \textbf{NPT} importance scores.
After training, we sort features by the scores at the final checkpoint, corresponding to the fully trained model. 
We fix this feature ordering and partition the ordered features into bins for each checkpoint.
Thus, each row of each heatmap in Figure~\ref{f:heatmaps} represents the same set of features for the corresponding dataset, allowing us to track the average $\delta_m$ per bin over time.
For both datasets, we indeed identify relevant features as early as the first checkpoint, as the ranking of features is relatively consistent throughout training.
This validates that our adaptive approach can identify and preserve the relative importance of features before full convergence, as the importance trends remain consistent over the course of training.

To validate the consistency of {\bf NPT} feature selection, we also compute the average last checkpoint in which each ArXiv feature is kept before being dropped in Figure~\ref{f:heatmaps}c.
We find that the features of highest importance are consistently ranked high, while the least important features are always dropped early.
For more concrete verification that {\bf NPT} can identify importance in a controlled setting, Figure~\ref{f:fi_synth} in Appendix~\ref{app:feat_import} visualizes importance scores using synthetic graph data, comparing scores obtained from {\bf NPT}, {\bf TFI}, {\bf MI}, and {\bf PT} (permutation testing via an MLP instead of a GNN).

We also analyze the types of features deemed important by {\bf NPT} across datasets, further verifying the generality of {\bf NPT} as the metric in Algorithm~\ref{alg:nfpt}.
To this end, Figure~\ref{f:heatmaps}d plots normalized importance scores computed from baseline metrics versus {\bf NPT} scores for Wisconsin.
We observe that {\bf NPT} tends to rank Wisconsin features as more important with higher {\bf MI} and lower homophily $h_{\rm GE}$, as expected for data with heterophilic labels.
Table~\ref{t:fi_correlation} expands on this by listing the linear correlation between {\bf NPT} and baseline scores for all datasets.
% To expand on this analysis, Table~\ref{t:fi_correlation} lists the linear correlation between {\bf NPT} scores and scores from each baseline for all datasets.
In all cases, {\bf NPT} attains its highest correlation with the metrics that performed best in Table~\ref{t:real_fs}.
This result indicates two takeaways.
First, {\bf NPT} indeed identifies feature importance in based on relevant data properties {\it without requiring prior information}.
Second, because {\bf NPT} detects relevant characteristics and attains competitive performance across datasets of various types, our approach is {\it a theoretically valid and empirically effective general choice for node feature selection}.
Analogous plots of Figures~\ref{f:heatmaps}c,d for the remaining datasets are in Appendix~\ref{app:feat_import}.
% %%%%%%%%%%%%%%%%%%%%%%%%%%%%%%%%%%%%%%%%%%%%%

\begin{figure*}[b!]
    \centering
    % --------------------------------
    \scalebox{.92}{\includegraphics[width=\textwidth]{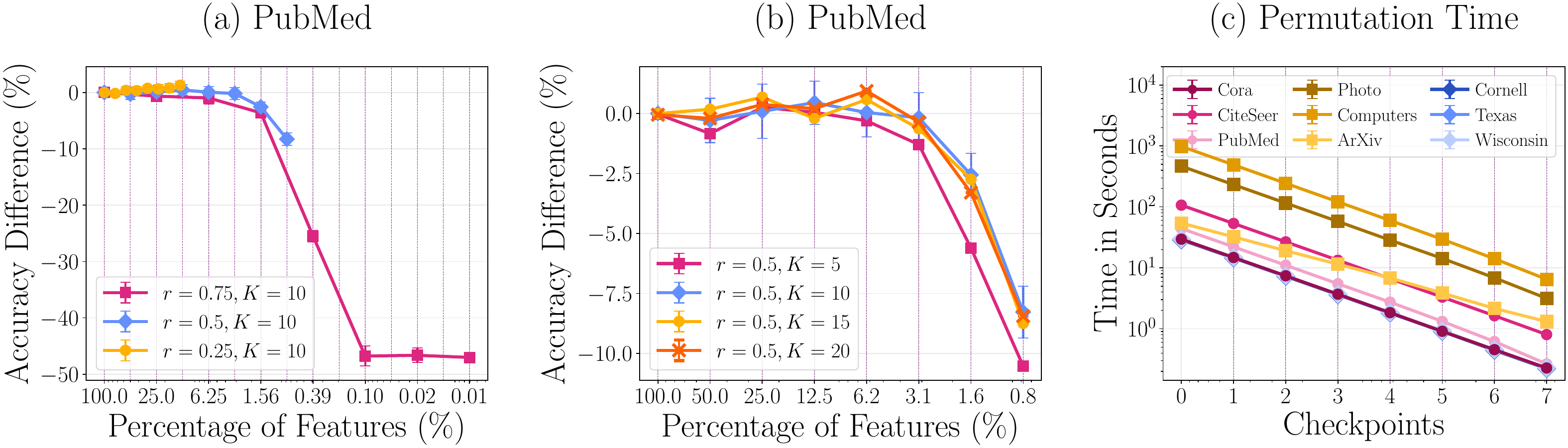}}
    % --------------------------------
    \vspace{-.1cm}
    \caption{
        Evaluation of Algorithm~\ref{alg:nfpt} in various scenarios.
        (a) GCN performance on PubMed for fixed $K=10$ and varying $r \in \{ 0.25, 0.5, 0.75 \}$.
        (b) GCN performance on PubMed for fixed $r=0.5$ and varying $K \in \{ 5, 10, 15, 20 \}$.
        (c) Time to permute node features for each checkpoint of Algorithm~\ref{alg:nfpt}, that is, for $t = nT$ for $n\in\naturals$, across multiple datasets.
    }
    \label{f:ablation}
    \vspace{-.3cm}
\end{figure*}

% %%%%%%%%%%%%%%%%%%%%%%%%%%%%%%%%%%%%%%%%%%%%%
\subsection{Method performance analysis}\label{Ss:perform}

% We next our method in Algorithm~\ref{alg:nfpt} using {\bf NPT} scores in various settings.
We next demonstrate the performance of Algorithm~\ref{alg:nfpt} using {\bf NPT} for PubMed while varying either $r$ or $K$, shown in Figure~\ref{f:ablation}a,b.
% First, we demonstrate the performance of a GCN on PubMed while varying either $r$ or $K$, shown in Figure~\ref{f:ablation}a,b.
Further results can be found in Appendix~\ref{app:model_perf}.
As we drop more features via larger $r\%$, we naturally experience an increasing drop in accuracy.
However, dropping features more slowly with $r = 0.25$ may improve performance, although we retain more features for the same number of training iterations.
Moreover, we require a large enough $K$ to perform enough permutations for a statistically relevant result.
In Figure~\ref{f:ablation}b, increasing $K$ above our choice of $10$ in previous simulations does not drastically change results, but lower $K$ can have negative effects on performance, as expected.
See Proposition~\ref{prop:sample_bound} in Appendix~\ref{app:featperm} for a statistical lower bound on the smallest $K$ required for desirable performance.
% In Prop, we provide a bound on the smallest K for desired performance
% For a statistical choice of $K$, see Proposition~\ref{prop:sample_bound} in Appendix~\ref{app:featperm}.
% 
Finally, we measure the additional cost of permuting during training in Figure~\ref{f:ablation}c.
We observe the exact decay in permutation time as discussed in Section~\ref{Ss:method}, where the cost of permutations is largest at the first checkpoint, but subsequent checkpoints decrease exponentially in duration. 
Moreover, as expected, graph size $N$ and the number of features $M$ control how costly computation will be, with dataset details listed in Appendix~\ref{app:data_info}.
% %%%%%%%%%%%%%%%%%%%%%%%%%%%%%%%%%%%%%%%%%%%%%

\renewcommand{\arraystretch}{.7}
% \addtolength{\tabcolsep}{-0.3em}
\begin{table*}[t!]
\scriptsize
\centering
% \vspace{-.2cm}
\caption{ 
% \small
Pearson correlation coefficient between NPT and other importance scores.
% The top performing method is \textbf{boldfaced}, and the second best is \underline{underlined}.
(Notation: {\bf Best}, \underline{second best}, ``\textendash''~denotes OOM)
%     Node classification accuracy for multiple datasets under various perturbations. 
% The top performing method is \textbf{boldfaced}, and the secondmost \underline{underlined}.
% \(\tilde{X}:\)\textit{Permuted features}, 
% \(\mathcal{N}:\)\textit{Random features}, 
% \(\tilde{A}:\)\textit{Random ER graph with same number of edges.}
% 100 epochs for the first 3 datasets. 300 epochs for Photo and Computers. GCN for the first 5 datasets. TAG for the last three datasets.
}
\vspace{-.1cm}
% \label{tab:structure-feature-ablation}
\label{t:fi_correlation}
\begin{tabular}{l|ccc|ccc|ccc}
\toprule
\textbf{Method} 
& \textbf{Cora} 
& \textbf{CiteSeer} 
& \textbf{PubMed} 
& \textbf{Photo} 
& \textbf{Computers} 
& \textbf{ArXiv} 
& \textbf{Cornell} 
& \textbf{Texas} 
& \textbf{Wisconsin} \\
\midrule
TFI 
& {\bf 0.6234}
& {\bf 0.5032}
& \underline{0.2618}
& 0.3098
& 0.3510
& $-$
& \underline{0.3659}
& \underline{0.3048}
& \underline{0.5225} \\
MI
& \underline{0.6178}
& \underline{0.4969}
& {\bf 0.6026}
& {\bf 0.5897}
& {\bf 0.6380}
& {\bf 0.5872}
& {\bf 0.4314}
& {\bf 0.4142}
& {\bf 0.5658} \\
$h_{\rm attr}$
& 0.1800
& 0.0373
& 0.2192
& \underline{0.5260}
& \underline{0.6245}
& \underline{0.5656}
& 0.0417
& -0.0093
& 0.0335 \\
$h_{\rm Euc}$
& 0.0208
& 0.0266
& 0.1116
& 0.4152
& 0.2715
& 0.4523
& 0.0016
& -0.0171
& 0.0370 \\
$h_{\rm GE}$
& -0.6479
& -0.5349
& -0.2285
& 0.1541
& 0.3072
& 0.0411
& -0.5712
& -0.5204
& -0.6635 \\
\bottomrule
\end{tabular}
\vspace{-.3cm}
\end{table*}
% \addtolength{\tabcolsep}{0.3em}
\renewcommand{\arraystretch}{1.}

% %%%%%%%%%%%%%%%%%%%%%%%%%%%%%%%%%%%%%%%%%%%%%
\section{Conclusion}\label{S:conc}

In this work, we presented permutation tests for node feature importance.
We verified the use of permutation-based importance scores for GCNs, both theoretically and empirically.
We thus exploit a well-established statistical metric, but we also verified that it returns relevant information that is unique to GNNs for graph-structured data.
% Our approach allows us to exploit a well-established statistical metric, but we also verified that it returns relevant information that is unique to GNNs for graph-structured data.
In particular, we applied our permutation scores for feature selection and found them to either rival or outperform other, data-specific metrics, regardless of graph dataset.
Furthermore, we presented an adaptive algorithm to eliminate features during training.
We can train a model and select features in a single process with our algorithm using any feature importance metric, but our permutation-based scores remained a theoretically well-established yet empirically effective choice for general scenarios with no prior information.

% We demonstrated that our algorithm can identify irrelevant features during training using multiple importance metrics, where our permutation scores either rival or outperform other, data-specific scores.
% We compared our permutation scores to other importance metrics for feature selection.
% We also demonstrated the effectiveness of our algorithm on multiple datasets, where we compared using permutation-based feature importance versus other metrics for adaptive feature selection.

\subsection{Scope and Limitations}\label{Ss:scope}

We also share limitations of this work to inspire future directions.
We require no assumptions on graph data, but performance-based metrics such as ours necessitate an appropriate selection of the GNN architecture.
While a reasonable requirement, the interpretation of the importance scores may change depending on the model used.
Furthermore, we demonstrated our approach only for node classification, but as $\delta_m$ can be employed to evaluate the effect of node features on any quantity, future work will see feature selection for link prediction and graph classification.
Moreover, while permutation tests are typically found to be very effective~\citep{khan2025UsingPermutationBasedFeature}, permuting features that are correlated may result in overestimated importance scores~\citep{hooker2021Unrestrictedpermutationforces}.
Thus, we plan to explore conditional permutation tests for the explainability of graph data.
Additionally, we chose a fixed pruning ratio $r$ for a fair comparison with {\bf TFI} and {\bf MI}, as these baselines are computed once before training and therefore do not naturally permit a dynamic feature selection approach. 
However, we expect that the performance of our algorithm can be improved further by adaptively eliminating features for which $\delta_m \leq 0$, which we explore in future work.

Observe that we analyzed how graph structure affects feature importance using well-established permutations, which are not specific to graphs.
Our result in Section~\ref{S:permtest} shows that this can identify {\it which} features are important but not necessarily {\it why}.
For example, a feature may be highly correlated with labels or highly dependent on graph structure.
Thus, it would be valuable to develop novel, graph-dependent methods, such as incorporating edge rewiring or performing conditional permutation tests that depend on node degrees or local neighborhoods.
Such a pursuit would complement and build on our analyses in Section~\ref{S:permtest} and Appendix~\ref{app:gnn_performance}, which imply the need for a more complex and structured approach to node feature importance evaluation.

\clearpage

\section*{Impact Statement}

In this paper, we propose a statistically well-motivated, permutation-based approach to measuring feature importance as well as an adaptive algorithm to drop features during GNN training based on importance scores.
These contributions promote more efficient model design, as we need not perform feature selection prior to training a GNN.
Moreover, the development of graph-based feature importance scores is relevant yet understudied.
This is particularly critical as the use of black-box GNNs remains prevalent, despite the increasing need for more interpretable methodologies.
Feature importance is a fundamental approach to understanding such model decisions.
Furthermore, our focus on feature importance for graphs will allow practitioners to promote novel insights into important variables in real-world applications using graph-structured data.

\section*{Acknowledgments}
This work was supported by the NSF under grants EF-2126387 and CCF-2340481. 
Research was sponsored by the Army Research Office and was accomplished under Grant Number W911NF-17-S-0002. 
The views and conclusions contained in this document are those of the authors and should not be interpreted as representing the official policies, either expressed or implied, of the Army Research Office or the U.S. Army or the U.S. Government. 
The U.S. Government is authorized to reproduce and distribute reprints for
Government purposes notwithstanding any copyright notation herein.

% \ibmyellow{
% Authors are \textbf{required} to include a statement of the potential broader
% impact of their work, including its ethical aspects and future societal
% consequences. This statement should be in an unnumbered section at the end of
% the paper (co-located with Acknowledgements -- the two may appear in either
% order, but both must be before References), and does not count toward the paper
% page limit. In many cases, where the ethical impacts and expected societal
% implications are those that are well established when advancing the field of
% Machine Learning, substantial discussion is not required, and a simple
% statement such as the following will suffice:

% ``This paper presents work whose goal is to advance the field of Machine
% Learning. There are many potential societal consequences of our work, none
% which we feel must be specifically highlighted here.''

% The above statement can be used verbatim in such cases, but we encourage
% authors to think about whether there is content which does warrant further
% discussion, as this statement will be apparent if the paper is later flagged
% for ethics review.
% }


\bibliography{citations}
\bibliographystyle{icml2026}

%%%%%%%%%%%%%%%%%%%%%%%%%%%%%%%%%%%%%%%%%%%%%%%%%%%%%%%%%%%%%%%%%%%%%%%%%%%%%%%
%%%%%%%%%%%%%%%%%%%%%%%%%%%%%%%%%%%%%%%%%%%%%%%%%%%%%%%%%%%%%%%%%%%%%%%%%%%%%%%
% APPENDIX
%%%%%%%%%%%%%%%%%%%%%%%%%%%%%%%%%%%%%%%%%%%%%%%%%%%%%%%%%%%%%%%%%%%%%%%%%%%%%%%
%%%%%%%%%%%%%%%%%%%%%%%%%%%%%%%%%%%%%%%%%%%%%%%%%%%%%%%%%%%%%%%%%%%%%%%%%%%%%%%
\newpage
\appendix
\onecolumn
\section{Related Work}\label{app:related}

Measuring variable importance is a fundamental task in several fields such as machine learning, statistics, and signal processing~\citep{fisher2019AllModelsWrong,mandler2024ReviewBenchmarkFeature}.
Classical techniques for classification tasks seek to identify correlations between features and labels to be predicted, such as their mutual information~\citep{theng2024Feature}.
Simpler, interpretable models such as linear regression and decision trees can be used as surrogate models to explain sample or feature relevance~\citep{ribeiro2016why}.
To avoid training simple, albeit cheap, models, one of the most common approaches is to apply perturbations, where model inputs or parameters are perturbed and the change in output measured~\citep{datta2016AlgorithmicTransparencyQuantitative,fisher2019AllModelsWrong,covert2021explainingremoving}.
Seminal examples include feature occlusion~\citep{feng2013efficient,lei2018distributionfree}, permutation~\citep{altmann2010Permutationimportancecorrected,brieman2001RandomForests,datta2016AlgorithmicTransparencyQuantitative}, and Shapley values~\citep{lundberg2018consistent,chen2018lshapley}. 
Scores based on measuring model outcomes under perturbations may require training multiple models to be used for feature selection~\citep{wang2024Featureselectionstrategies}.
% More specifically, perturbation-based scores require that the change in output must be measured using a trained model.
% However, this results in an inefficient process if the scores are intended to select features for training~\precite.
Not only is this potentially infeasible computationally, but for optimizing models with nonconvex losses, differences in performance for models trained on perturbed data may be misleading.
% Alternatively, other works have sought to learn feature importance via an auxiliary model that returns masks for feature selection~\precite, for which many tend to use uninterpretable models.

Past works such as~\citep{luan2024WhenGraphNeural,zheng2024WhatMissingGraph} investigated how the relationship among graph structure, node features, and node labels influences GNN performance.
These works applied parametric models for the \emph{global} relationships between entities and did not provide the specificity to describe the influence of a single feature, but the results more than justify the need for graph-specific feature importance scores and feature selection methods.
For graph-structured data, a plethora of works seek to identify the contribution of nodes or edges to particular GNN predictions~\citep{alkhoury2025Improvinggraphneural,akkas2024GNNShapScalableAccurate,chen2024Identifyinginfluentialnodes,huang2023GraphLIMELocalInterpretable}.
Among these, some works consider node feature relevance, albeit primarily as they pertain to structural importance~\citep{fang2023CooperativeExplanationsGraph,chen2024MotifGraphNeural}.
Feature importance methods have been proposed specifically for graphs~\citep{zheng2025letyourfeatures}, which often require assumptions about the type of graph data~\citep{mahmoud2023Nodeclassificationgraph,shao2024reviewfeatureselection}.
For example, as GCNs are a highly popular family of GNNs, the homophily of node features has been explored as relevance measurements~\citep{zhu2024ImpactFeatureHeterophily}.
The score proposed in~\citep{zheng2025letyourfeatures} computes the mutual information between labels and node features passed through a linear low-pass filter, implying relevance for a GCN.
Authors considered all features informative, and their metric was used to identify which features ought to be trained with a GNN versus an MLP. 
Thus, they did not evaluate their metric for eliminating features to reduce model complexity or to remove unhelpful features.
Conversely, several works aim to select node features during training, albeit without returning importance scores~\citep{maurya2023FeatureselectionKey,jiang2023Sparsenormregularized,acharya2020FeatureSelectionExtraction,lin2020GraphNeuralNetworks,zheng2020GSSAPayattention}.
    % Feature selection: Key to enhance node classification with graph neural networks
    % Sparse norm regularized attribute selection for graph neural networks
    % Feature Selection and Extraction for Graph Neural Networks
    % Graph Neural Networks Including Sparse Interpretability
    % GSSA: Pay attention to graph feature importance for GCN via statistical self-attention
Moreover, these methods learn which features to eliminate via an auxiliary model, for which many tend to use uninterpretable models.
Furthermore, even if some return learned masks that indicate important features, the need to train additional submodules introduces more parameters, which undermines the goal of reducing dimensionality~\citep{maurya2022Simplifyingapproachnode,acharya2020FeatureSelectionExtraction,lin2020GraphNeuralNetworks,zheng2020GSSAPayattention}.
% Additionally, while some of these train submodules to learn masks for identifying important features, these approaches can require learning additional parameters, undermining the goal of reducing dimensionality~\citep{maurya2022Simplifyingapproachnode,acharya2020FeatureSelectionExtraction,lin2020GraphNeuralNetworks,zheng2020GSSAPayattention}.

% \section{Perturbation Results for Synthetic Data}\label{app:synth_glob_perm}

\section{Proof of Theorem~\ref{thm:gcn_error}}\label{app:gcn_error}

The following proof is inspired by that of~\citep{tenorio2025adaptingheterophilicgraphdata}, which was itself motivated by~\citep{nt2021RevisitingGraphNeural} for evaluating GCN dependence on homophily.
We prove the bound in~\eqref{eq:gcn_error} recursively.
We first bound the difference in outputs of the first layer $\| \bbZ^{*(1)} - \bbZ^{(1)} \|_F$, from which we can bound $\| \bbZ^{*(\ell)} - \bbZ^{(\ell)} \|_F$ for any number of layers $\ell \in \naturals$.
By the definitions of $\bbZ^*$, $\bbZ$, the GCN in~\eqref{eq:gcn_single_layer}, and the fact that $\bbX^* = \tbA^*_{\rm rw} \bbX^*$, we can bound the outputs of the first GCN layer $\ell=1$ for
\alna{
    \big\| \bbZ^{*(1)} - \bbZ^{(1)} \big\|_F
    &~=~&
    \Big\|
        \sigma_1\big( \tbA^*_{\rm rw} \bbX^* \bbTheta^{(1)} \big)
        -
        \sigma_1\big( \tbA_{\rm rw} \bbX \bbTheta^{(1)} \big)
    \Big\|_F
&\nonumber\\&
    &~\leq~&
    \tau \omega
    \big\|
        \tbA^*_{\rm rw} \bbX^*
        -
        \tbA_{\rm rw} \bbX
    \big\|_F
&\nonumber\\&
    &~=~&
    \tau \omega
    \big\|
        \bbX^*
        -
        \tbA_{\rm rw} \bbX
    \big\|_F
&\nonumber\\&
    &~=~&
    \tau \omega
    \big\|
        \bbY\bbP^{-1}\bbY^\top \bbX
        -
        \tbA_{\rm rw} \bbX
    \big\|_F,
\nonumber}
where the inequality is due to the $\tau$-Lipschitzness of $\sigma_1$ and the boundedness of the weights $\bbTheta^{(1)}$, and the final equality is by the definition of the idealized features $\bbX^* = \bbY\bbP^{-1}\bbY^\top$.
% Next, let $\tbD^* := \diag( (\bbA^* + \bbI) \bbone )$, analogous to $\tbD = \diag(\bbd + \bbone)$, and recall that $\tbA_{\rm rw} = \tbD^{-1}(\bbA + \bbI)$ and $\tbA^*_{\rm rw} = (\tbD^*)^{-1}(\bbA^* + \bbI)$.
% Then, by the definition of the idealized features $\bbX^* = \bbY\bbP^{-1}\bbY^\top$, we can rewrite the upper bound above as
Then, recalling that $\tbD = \diag(\bbd + \bbone)$ and $\tbA_{\rm rw} = \tbD^{-1}(\bbA + \bbI)$, we rewrite the upper bound and apply the triangle inequality for a node-wise error bound, that is,
\alna{
    \big\| \bbZ^{*(1)} - \bbZ^{(1)} \big\|_F
    &~\leq~&
    \tau \omega
    \big\|
        \bbY\bbP^{-1}\bbY^\top \bbX
        -
        \bbP^{-1}\bbP \bbX
        +
        \tbD^{-1} \tbD \bbX
        -
        \tbA_{\rm rw} \bbX
    \big\|_F
&\nonumber\\&
    &~\leq~&
    \tau \omega
    \sum_{i=1}^N
    \bigg\|
        \Big[
            \big(
                \bbY\bbP^{-1}\bbY^\top - \bbP^{-1}\bbP 
            \big)
            \bbX
        \Big]_{i,:}
        +
        \Big[
            \tbD^{-1}
            \big(
                \bbA - \bbD
            \big)
            \bbX
        \Big]_{i,:}
    \bigg\|_2.
\nonumber}
Then, to bound the error with respect to pairs of nodes, we rewrite the above and again apply the triangle inequality for
\alna{
    \big\| \bbZ^{*(1)} - \bbZ^{(1)} \big\|_F
    &~\leq~&
    \tau \omega
    \sum_{i=1}^N
    \Bigg\|
        \sum_{j=1}^N
        \frac{ [\bbY\bbY^\top]_{ij} }{ p_{y_i} }
        \bbX_{j,:}
        -
        \frac{p_{y_i}}{p_{y_i}}
        \bbX_{i,:}
        -
        \frac{1}{d_i+1}
        \Big(
            A_{ij} \bbX_{j,:}
            -
            d_i \bbX_{i,:}
        \Big)
    \Bigg\|_2
&\nonumber\\&
    &~=~&
    \tau \omega
    \sum_{i=1}^N
    \Bigg\|
        \sum_{j=1}^N
        \frac{ [\bbY\bbY^\top]_{ij} }{ p_{y_i} }
        \big( \bbX_{j,:} - \bbX_{i,:} \big)
        -
        \frac{A_{ij}}{ d_i+1 }
        \big( \bbX_{j,:} - \bbX_{i,:} \big)
    \Bigg\|_2
&\nonumber\\&
    &~\leq~&
    \tau \omega
    \sum_{i=1}^N
    \sum_{j=1}^N
    \Bigg\|
        \bigg(
            \frac{ [\bbY\bbY^\top]_{ij} }{ p_{y_i} }
            -
            \frac{A_{ij}}{ d_i+1 }
        \bigg)
        \big( \bbX_{j,:} - \bbX_{i,:} \big)
    \Bigg\|_2,
\nonumber}
where we recall that $[\bbY\bbY^\top]_{ij} = 1$ if and only if nodes $i$ and $j$ belong to the same class, that is, $y_i = y_j$, otherwise $[\bbY\bbY^\top]_{ij} = 0$.
We can express this via the following sum
\alna{
    \big\| \bbZ^{*(1)} - \bbZ^{(1)} \big\|_F
    &~\leq~&
    \tau \omega
    \sum_{c=1}^C
    \sum_{i=1}^N
    \sum_{j=1}^N
    \mbI( y_i = c )
    \mbI( y_j = c )
    \cdot
    \Bigg\|
        \bigg(
            \frac{ [\bbY\bbY^\top]_{ij} }{ p_{y_i} }
            -
            \frac{A_{ij}}{ d_i+1 }
        \bigg)
        \big( \bbX_{j,:} - \bbX_{i,:} \big)
    \Bigg\|_2
&\nonumber\\&
    &~=~&
    \tau \omega
    \sum_{c=1}^C
    \sum_{i=1}^N
    \sum_{j=1}^N
    \Bigg\|
        \bigg(
            \frac{ Y_{ic} Y_{jc} }{ p_{c} }
            -
            \frac{A_{ij}}{ d_i+1 }
        \bigg)
        \big( \bbX_{j,:} - \bbX_{i,:} \big)
    \Bigg\|_2
&\nonumber\\&
    &~=~&
    \tau \omega
    \sum_{c=1}^C
    \sum_{i=1}^N
    \sum_{j=1}^N
    \bigg|
        \frac{ Y_{ic} Y_{jc} }{ p_{c} }
        -
        \frac{A_{ij}}{ d_i+1 }
    \bigg|
    \cdot
    \big\|
        \bbX_{j,:} - \bbX_{i,:}
    \big\|_2,
\label{eq:gcn_layer1}}
which characterizes the error for the output of GCN layer $\ell=1$.

Analogously, for an arbitrary layer $\ell \in [L]$, we apply the $\tau$-Lipschitzness of $\sigma_{\ell}$ and the boundedness of $\bbTheta^{(\ell)}$ for
\alna{
    \big\| \bbZ^{*(\ell)} - \bbZ^{(\ell)} \big\|_F
    &~\leq~&
    \tau \omega
    \big\|
        \tbA^*_{\rm rw}
        \bbZ^{*(\ell-1)}
        -
        \tbA_{\rm rw}
        \bbZ^{(\ell-1)}
    \big\|_F
&\nonumber\\&
    &~\leq~&
    \tau \omega
    \Big\|
        \big(
            \tbA^*_{\rm rw} - \tbA_{\rm rw}
        \big)
        \bbZ^{*(\ell-1)}
    \Big\|_F
    +
    \tau \omega
    \Big\|
        \tbA_{\rm rw}
        \big(
            \bbZ^{*(\ell-1)} - \bbZ^{(\ell-1)}
        \big)
    \Big\|_F.
\nonumber}
% yielding the error bound in~\eqref{eq:gcn_error}, as desired.
Then, we can apply the Cauchy-Schwarz inequality to both terms of the upper bound for
\alna{
    \big\| \bbZ^{*(\ell)} - \bbZ^{(\ell)} \big\|_F
    &~\leq~&
    \tau \omega
    \big\|
        \tbA^*_{\rm rw} - \tbA_{\rm rw}
    \big\|_F
    \cdot
    \big\| \bbZ^{*(\ell-1)} \big\|_F
    +
    \tau \omega
    \big\|
        \bbZ^{*(\ell-1)} - \bbZ^{(\ell-1)}
    \big\|_F
\label{eq:gcn_bnd_before}}
since $\| \tbA_{\rm rw} \|_2 = 1$.
We next bound the adjacency matrix discrepancy $\|\tbA_{\rm rw}^* - \tbA_{\rm rw}\|_F$.
We first let $\tbD^* := \diag( (\bbA^* + \bbI) \bbone )$, analogous to $\tbD = \diag(\bbd + \bbone)$, and recall that $\tbA_{\rm rw} = \tbD^{-1}(\bbA + \bbI)$ and $\tbA^*_{\rm rw} = (\tbD^*)^{-1}(\bbA^* + \bbI)$.
Then, we have that
\alna{
    \big\| \tbA_{\rm rw}^* - \tbA_{\rm rw} \big\|_F
    &~=~&
    \big\|
        \tbA^*_{\rm rw}
        -
        \tbD^{-1}
        (\bbA^* + \bbI)
        +
        \tbD^{-1}
        (\bbA^* + \bbI)
        -
        \tbD^{-1}
        (\bbA + \bbI)
    \big\|_F
&\nonumber\\&
    &~\leq~&
    \big\|
        \tbA^*_{\rm rw}
        -
        \tbD^{-1}
        (\bbA^* + \bbI)
    \big\|_F
    +
    \big\|
        \tbD^{-1}
        \big( \bbA^* - \bbA \big)
    \big\|_F
&\nonumber\\&
    &~\leq~&
    \big\|
        \big(\bbI - \tbD^{-1} \tbD^*\big)
        \tbA^*_{\rm rw}
    \big\|_F
    +
    \big\|
        \tbD^{-1}
        \bbDelta
    \big\|_F,
\nonumber}
where $\tbD^{-1}(\bbA^* + \bbI) = \tbD^{-1}\tbD^* \tbA^*_{\rm rw}$.
Then, with $\| \tbA^*_{\rm rw} \|_2 = 1$, we obtain our adjacency matrix error bound as
\alna{
    \big\| \tbA_{\rm rw}^* - \tbA_{\rm rw} \big\|_F
    &~\leq~&
    \big\|
        \tbD^{-1}
        \big(\tbD - \tbD^*\big)
    \big\|_F
    +
    \|
        \bbDelta
    \|_F
&\nonumber\\&
    &~\leq~&
    \| \bbD - \bbD^* \|_F
    +
    \| \bbDelta \|_F
&\nonumber\\&
    &~=~&
    \| \diag(\bbDelta \bbone) \|_F
    +
    \| \bbDelta \|_F
&\nonumber\\&
    &~\leq~&
    \| \bbDelta \bbone \|_2
    +
    \| \bbDelta \|_F
&\nonumber\\&
    &~\leq~&
    \sqrt{N}
    \| \bbDelta \|_F
    +
    \| \bbDelta \|_F.
\label{eq:thm1_Adiff_bound}}
In addition to the above inequality, we must also bound the embeddings $\| \bbZ^{*(\ell-1)} \|_F$, for which we have that
\alna{
    \big\|
        \bbZ^{*(\ell-1)}
    \big\|_F
    &~\leq~&
    \tau \omega
    \big\|
        \tbA^*_{\rm rw}
        \bbZ^{*(\ell-2)}
    \big\|_F
    ~\leq~
    \tau \omega
    \big\|
        \bbZ^{*(\ell-2)}
    \big\|_F
    ~\leq~
    (\tau\omega)^{\ell-1}
    \| \bbX \|_F,
\label{eq:gcn_output_bound}}
which, together with~\eqref{eq:thm1_Adiff_bound}, we substitute into~\eqref{eq:gcn_bnd_before} for
\alna{
    \big\| \bbZ^{*(\ell)} - \bbZ^{(\ell)} \big\|_F
    &~\leq~&
    \tau^{\ell} \omega^{\ell}
    (1 + \sqrt{N})
    \| \bbDelta \|_F
    \cdot
    \|\bbX\|_F
    +
    \tau\omega
    \big\|
        \bbZ^{*(\ell-1)} - \bbZ^{(\ell-1)}
    \big\|_F.
\nonumber}
We expand the above relationship recursively to obtain 
\alna{
    \big\| \bbZ^{*(\ell)} - \bbZ^{(\ell)} \big\|_F
    &~\leq~&
    \tau^{\ell} \omega^{\ell}
    (\ell-1)
    (1 + \sqrt{N})
    \| \bbDelta \|_F
    \cdot
    \|\bbX\|_F
    +
    (\tau\omega)^{\ell-1}
    \big\| \bbZ^{*(1)} - \bbZ^{(1)} \big\|_F.
\nonumber}
Since $\bbZ = \bbZ^{(L)}$ and $\bbZ^* = \bbZ^{*(L)}$, combining the above with~\eqref{eq:gcn_layer1} for $\ell = L$ yields the desired error bound in~\eqref{eq:gcn_error}.  $\hfill\blacksquare$

\section{Proof of Theorem~\ref{thm:perm_gcn_error}}\label{app:perm_gcn_error}

Observe that we may repeat the steps of the proof of Theorem~\ref{thm:gcn_error} in Appendix~\ref{app:gcn_error} to obtain
\alna{
    \| \tbZ^* - \tbZ \|_F
    &~\leq~&
    \tau^\ell \omega^\ell
    \!
    \Bigg[
        (\ell-1)
        (1 + \sqrt{N})
        \|\bbDelta\|_F \|\bbX\|_F
        +
        \sum_{i=1}^N
        \sum_{j=1}^N
        \left|
            \frac{\bbY_{i,:}\bbY_{j,:}^\top}{p_{y_i}}
            -
            \frac{A_{ij}}{d_i + 1}
        \right|
        \cdot
        \| \tbX_{i,:} - \tbX_{j,:} \|_2
    \Bigg].
\label{eq:pert_error_init}}
Thus, for the remainder of the proof, we need only obtain a bound for $\| \tbX_{i,:} - \tbX_{j,:} \|_2$.
We proceed with bounding $(\tilde{X}_{im} - \tilde{X}_{jm})^2$ for any $m \in [M]$.
For a given permutation $\bbpi \in \Pi$ sampled uniformly at random, we define the function
\alna{
    \psi(\bbpi)
    :=
    \big( 
        X_{\pi(i),m} - X_{\pi(j),m} 
    \big)^2
\label{eq:psi}}
with expected value
\alna{
    \mbE[\psi(\bbpi)]
    &~=~&
    \mbE\big[
    \big( 
        X_{\pi(i),m} - X_{\pi(j),m} 
    \big)^2
    \big]
&\nonumber\\&
    &~=~&
    \mbE\big[ X_{\pi(i),m}^2 \big]
    -
    2
    \mbE\big[ X_{\pi(i),m} X_{\pi(j),m} \big]
    +
    \mbE\big[ X_{\pi(j),m}^2 \big].
\label{eq:exp_psi_init}}
Then, for any $i,j \in [N]$ such that $i\neq j$ and $m \in [M]$,
\alna{
    \mbE[ X_{\pi(i),m}^2 ]
    &~=~&
    \frac{1}{N!}
    \sum_{\bbpi \in \Pi}
    X_{\pi(i),m}^2
    ~=~
    \frac{1}{N!}
    \sum_{j=1}^N
    \sum_{\bbpi \in \Pi}
    X_{jm}^2
    \mbI( \pi(i) = j )
    ~=~
    \frac{1}{N}
    \sum_{j=1}^N
    X_{jm}^2
    ~=~
    \frac{1}{N}
    \| \bbX_{:,m} \|_2^2
\nonumber}
and
\alna{
    \mbE[ X_{\pi(i),m} X_{\pi(j),m} ]
    &~=~&
    \frac{1}{N!}
    \sum_{\bbpi \in \Pi}
    X_{\pi(i),m} X_{\pi(j),m}
&\nonumber\\&
    &~=~&
    \frac{1}{N!}
    \sum_{k=1}^N
    \sum_{\ell =1}^N
    \sum_{\bbpi \in \Pi}
    X_{km} X_{\ell m}
    \mbI(\pi(i) = k)
    \mbI(\pi(j) = \ell)
&\nonumber\\&
    &~=~&
    \frac{1}{N!}
    \sum_{k\neq \ell}
    \sum_{\bbpi \in \Pi}
    X_{km} X_{\ell m}
    (N-2)!
&\nonumber\\&
    &~=~&
    \frac{1}{N(N-1)}
    \sum_{k=1}^N
    \sum_{\ell=1}^N
    X_{km} X_{\ell m}
    -
    \frac{1}{N(N-1)}
    \sum_{k=1}^N
    X_{km}^2
&\nonumber\\&
    &~=~&
    \frac{1}{N(N-1)}
    \Big(
        (\bbone^\top \bbX_{:,m})^2
        -
        \| \bbX_{:,m} \|_2^2
    \Big),
\nonumber}
which we substitute into~\eqref{eq:exp_psi_init} for
\alna{
    \mbE[\psi(\bbpi)]
    &~=~&
    \frac{2}{N-1}
    \bigg(
        \| \bbX_{:,m} \|_2^2
        -
        \frac{1}{N} (\bbone^\top\bbX_{:,m})^2
    \bigg).
\label{eq:exp_psi}}
Then, we define the Doob martingale $\{ Q_k \}_{k=0}^N$ such that $Q_0 = \mbE[\psi(\bbpi)]$ and 
\alna{
    Q_k = \mbE[
        \psi(\bbpi) ~|~ \pi(1),\dots,\pi(k-1)
    ]
    \quad
    \forall k=1,\dots,N,
\nonumber}
thus $Q_N = \psi(\bbpi)$.
Additionally, we have that
\alna{
    \mbE[\psi(\bbpi) ~|~ \pi(1),\dots,\pi(k-1)]
    &~=~&
    \sum_{\ell=k}^N
    \frac{1}{ N - k + 1 }
    \mbE[\psi( (k\ell) \bbpi) \,|\, \pi(1),\dots,\pi(k)],
\nonumber}
where $\psi((k\ell)\bbpi)$ denotes $\psi$ given $\bbpi$ with elements $k$ and $\ell$ swapped.
Then, we bound the following differences
\alna{
    | Q_k - Q_{k-1} |
    &~=~&
    \big|
        \mbE[ \psi(\bbpi) \,|\, \pi(1),\dots,\pi(k) ]
        -
        \mbE[ \psi(\bbpi) \,|\, \pi(1),\dots,\pi(k-1) ]
    \big|
&\nonumber\\&
    &~=~&
    \bigg|
        \mbE[ \psi(\bbpi) \,|\, \pi(1),\dots,\pi(k) ]
        -
        \sum_{\ell=k}^N
        \frac{1}{ N - k + 1 }
        \mbE[\psi( (k\ell) \bbpi) \,|\, \pi(1),\dots,\pi(k)]
    \bigg|
&\nonumber\\&
    &~=~&
    \bigg|
        \frac{1}{ N - k + 1 }
        \sum_{\ell=k}^N
        \mbE[ \psi(\bbpi) \,|\, \pi(1),\dots,\pi(k) ]
        -
        \mbE[\psi( (k\ell) \bbpi) \,|\, \pi(1),\dots,\pi(k)]
    \bigg|
&\nonumber\\&
    &~\leq~&
    \frac{1}{ N - k + 1 }
    \sum_{\ell=k}^N
    \Big|
        \mbE[ \psi(\bbpi) - \psi( (k\ell) \bbpi) \,|\, \pi(1),\dots,\pi(k) ]
    \Big|.
\label{eq:Qdiff_bound}}
Then, with $\tbX_{:,m}^{(k\ell)}$ representing $\tbX_{:,m}$ with elements $k$ and $\ell$ swapped, we have
\alna{
    |\psi(\bbpi) - \psi((k\ell)\bbpi)|
    &~=~&
    \big|
        \big(
            \tilde{X}_{im}
            -
            \tilde{X}_{jm}
        \big)^2
        -
        \big(
            \tilde{X}_{im}^{(k\ell)}
            -
            \tilde{X}_{jm}^{(k\ell)}
        \big)^2
    \big|,
\nonumber}
which is zero for $i=k,j=\ell$ or $i\neq k,j \neq \ell$, but for $k = i,j \neq \ell$, we instead have
\alna{
    |\psi(\bbpi) - \psi((k\ell)\bbpi)|
    &~=~&
    \big|
        \big(
            \tilde{X}_{im}
            -
            \tilde{X}_{\ell m}
        \big)
        \big(
            \tilde{X}_{im}
            +
            \tilde{X}_{\ell m}
            -
            2\tilde{X}_{jm}
        \big)
    \big|
&\nonumber\\&
    &~=~&
    \big|
        \tilde{X}_{im} - \tilde{X}_{\ell m}
    \big|
    \cdot
    \big|
        \tilde{X}_{im}
        +
        \tilde{X}_{\ell m}
        -
        2\tilde{X}_{jm}
    \big|
&\nonumber\\&
    &~\leq~&
    \big|
        \tilde{X}_{im} - \tilde{X}_{\ell m}
    \big|
    \cdot
    \Big(
        \big|
            \tilde{X}_{im}
            -
            \tilde{X}_{jm}
        \big|
        +
        \big|
            \tilde{X}_{\ell m}
            -
            \tilde{X}_{jm}
        \big|
    \Big)
&\nonumber\\&
    &~\leq~&
    2 \alpha
\nonumber}
by our assumption that $\max_{m\in[M]} \max_{k,\ell \in [N]} ( X_{km} - X_{\ell m} )^2 = \alpha$.
With~\eqref{eq:psi},~\eqref{eq:exp_psi}, and~\eqref{eq:Qdiff_bound}, we apply the Azuma–Hoeffding inequality~\citep{azuma1967Weightedsumscertain,hoeffding1963ProbabilityInequalitiesSums} for
\alna{
    \mbP\left[
        \psi(\bbpi) - \mbE[\psi(\bbpi)] \geq \eta
    \right]
    \leq
    \exp\left(
        -
        \frac{ \eta^2 }{2 N \alpha^2 }
    \right),
\nonumber}
and with $t := \eta / (\alpha \sqrt{N})$, we obtain
\alna{
    \big( 
        X_{\pi(i),m} - X_{\pi(j),m} 
    \big)^2
    \leq
    \frac{2}{N-1}
    \bigg(
        \| \bbX_{:,m} \|_2^2
        -
        \frac{1}{N} (\bbone^\top\bbX_{:,m})^2
    \bigg)
    +
    \alpha t \sqrt{N}
\nonumber}
with probability at least $1-e^{-t^2/2}$.
Thus, by Boole's inequality, we then have that
\alna{
    \mbP\Bigg[
        \big\| \bbX_{\pi(i),:} - \bbX_{\pi(j),:} \big\|_2^2
        \leq
        \sum_{m=1}^M
            \bigg[
            \frac{2}{N-1}
            \Big(
                \| \bbX_{:,m} \|_2^2
                -
                \frac{1}{N}
                (\bbone^\top \bbX_{:,m})^2
            \Big)
            +
            \alpha t \sqrt{N}
            \bigg]
    \Bigg]
&\nonumber\\&
    \qquad\qquad\qquad
    ~=~
    \mbP\Bigg[
        \big\| \bbX_{\pi(i),:} - \bbX_{\pi(j),:} \big\|_2^2
        \leq
        \frac{2}{N-1}
        \bigg(
            \| \bbX \|_F^2
            -
            \frac{1}{N} 
            \| \bbX^\top \bbone \|_2^2
        \bigg)
        +
        \alpha M t \sqrt{N}
    \Bigg]
&\nonumber\\&
    \qquad\qquad\qquad
    ~\geq~
    1-M e^{-t^2 /2}
    ~=~
    1 - e^{-t^2/2 + \log M}.
\nonumber}
Thus, with probability at least $1-e^{-t^2/2 + \log M}$,
\alna{
    \big\| \bbX_{\pi(i),:} - \bbX_{\pi(j),:} \big\|_2
    &~\leq~&
    \sqrt{
        \bigg(
            \| \bbX \|_F^2
            -
            \frac{1}{N} 
            \| \bbX^\top \bbone \|_2^2
        \bigg)
        +
        % \alpha M (t + \sqrt{2 \log M}) \sqrt{N}
        \alpha M t \sqrt{N}
    },
\nonumber}
% 
% which yields
% \alna{
%     \| \tbX_{i,:} - \tbX_{j,:} \|_2
%     &~=~&
%     \sqrt{ \left[
%         \sum_{m=1}^M
%         \frac{2}{N-1}
%         \bigg(
%             \| \bbX_{:,m} \|_2^2
%             -
%             \frac{1}{N} (\bbone^\top\bbX_{:,m})^2
%         \bigg)
%         +
%         \alpha t \sqrt{N}
%     \right] }
% &\nonumber\\&
%     &~\leq~&
%     \sqrt{
%         \frac{2}{N-1}
%         \bigg(
%         \| \bbX \|_F^2
%         -
%         \frac{1}{N}
%         \| \bbX^\top \bbone \|_2^2
%         \bigg)
%         +
%         M
%         \alpha t \sqrt{N}
%     },
% \nonumber}
which we substitute into~\eqref{eq:pert_error_init} for the error bound in~\eqref{eq:pert_gcn_error}, as desired.
$\hfill\blacksquare$

\section{Influence of Graph Data on Other GNNs}\label{app:gnn_performance}

We next introduce results analogous to Theorems~\ref{thm:gcn_error} and~\ref{thm:perm_gcn_error} for other GNNs besides GCNs.
The proofs for all of the results in this section can be found in Appendix~\ref{app:gnn_bound_proofs}.
For alternative architectures, we consider the GIN and TAGCN, as applied in all of the empirical results for node classification.

\subsection{GIN Performance}\label{Ss:gin_performance}

We begin with characterizing the error for node classification using GINs.

\begin{theorem}\label{thm:gin_error}
    Let $f: \reals^{N\times M} \rightarrow \reals^{N\times H}$ be an $L$-layer GIN with $\bbZ^{(0)} = \bbX$, $\bbZ = \bbZ^{(L)} = f(\bbX; \bbA, \bbTheta)$, and
    \alna{
        \bbZ^{(\ell)}
        &~=~&
        \sigma_{\ell}
        \Big(
            (\epsilon\bbI + \tbA_{\rm rw})
             \bbZ^{(\ell-1)} \bbTheta^{(\ell)}
        \Big)
        \quad
        \forall \ell \in [L]
    \label{eq:gin_single_layer}}
    for (potentially learnable) $\epsilon > 0$, $\tau$-Lipschitz activations $\sigma_{\ell}$, and weights $\bbTheta = \{ \bbTheta^{(\ell)} \}_{\ell=1}^L$ such that $\| \bbTheta^{(\ell)} \|_2 \leq \omega$ for $\ell \in [L]$.
    Then, with $\bbZ^* = f(\bbX^*; \bbA^*, \bbTheta)$ for $\bbX^*$ in~\eqref{eq:ideal_feats}, $\bbA^*$ in~\eqref{eq:ideal_A}, and $\bbDelta := \bbA - \bbA^*$,
    \alna{
        % \!\!\!\!\!\!
        \| \bbZ^* - \bbZ \|_F
        &~\leq~&
        % \,\leq\,
        % \tau^2 \omega^2
        \tau^{L} \omega^{L}
        (1 + \epsilon)^{L-1}
        \!
        \Bigg[
            % \tau^2 \omega^2
            (L-1)
            (1 + \sqrt{N})
            \|\bbDelta\|_F \|\bbX\|_F
        &\nonumber\\&
        % \!\!\!\!\!\!
            % ~~
            &&
            +
            % \tau^2 \omega^2
            \sum_{c=1}^C
            \sum_{i=1}^N
            \sum_{j=1}^N
            \left|
                (1 + \epsilon)
                \frac{Y_{ic}Y_{jc}}{p_c}
                -
                \frac{A_{ij}}{d_i + 1}
            \right|
            \!\cdot\!
            \| \bbX_{i,:} - \bbX_{j,:} \|_2
        \Bigg].
    &\label{eq:gin_error}}
\end{theorem}
Observe that we consider the same $\epsilon$ for every layer for simplicity, but it is straightforward to obtain similar results if $\epsilon$ differs per layer.
The formulation of the GIN layer in~\eqref{eq:gin_single_layer} allows the model to prioritize either a node's own features (larger $\epsilon$) or aggregating neighboring features (smaller $\epsilon$), where $\epsilon$ can either be learned or fixed prior to training.
Hence, when $\epsilon$ is larger, we require $\bbX$ to be more indicative of class labels $\bby$, that is, we require features to be more similar within classes and more different across classes, as $\epsilon$ outweighs the effect of neighboring nodes. 
Naturally, when $\epsilon$ is small, the GIN behaves more similarly to the GCN in~\eqref{eq:gcn_single_layer}, and therefore we have the same conclusions for GIN performance as for the GCN.

\subsection{TAGCN Performance}\label{Ss:tagcn_performance}

We also provide a node classification error bound for TAGCNs.
\begin{theorem}\label{thm:tagcn_error}
    Let $f: \reals^{N\times M} \rightarrow \reals^{N\times H}$ be an $L$-layer TAGCN with $\bbZ^{(0)} = \bbX$, $\bbZ = \bbZ^{(L)} = f(\bbX; \bbA, \bbTheta)$, and
    \alna{
        \bbZ^{(\ell)}
        &~=~&
        \sigma_{\ell}
        \Bigg(
            \sum_{k=0}^K
            \tbA_{\rm rw}^k
             \bbZ^{(\ell-1)} \bbTheta^{(\ell)}
        \Bigg)
        \quad
        \forall \ell \in [L]
    \label{eq:tagcn_single_layer}}
    for matrix polynomial order $K \in \naturals$, $\tau$-Lipschitz activations $\sigma_{\ell}$, and weights $\bbTheta = \{ \bbTheta^{(\ell)} \}_{\ell=1}^L$ such that $\| \bbTheta^{(\ell)} \|_2 \leq \omega$ for $\ell \in [L]$.
    Then, with $\bbZ^* = f(\bbX^*; \bbA^*, \bbTheta)$ for $\bbX^*$ in~\eqref{eq:ideal_feats}, $\bbA^*$ in~\eqref{eq:ideal_A}, and $\bbDelta := \bbA - \bbA^*$,
    \alna{
        \| \bbZ^* - \bbZ \|_F
        &~\leq~&
        \tau^{L} \omega^{L}
        (K+1)^{L-1}
        \!
        \Bigg[
            % \tau^2 \omega^2
            {\frac{1}{2}}
            % \cdot
            K(K+1)
            L
            (1 + \sqrt{N})
            \|\bbDelta\|_F \|\bbX\|_F
        &\nonumber\\&
        % \!\!\!\!\!\!
            % ~~
            &&
            +
            % \tau^2 \omega^2
            \sum_{c=1}^C
            \sum_{i=1}^N
            \sum_{j=1}^N
            \Bigg(
                \frac{Y_{ic} Y_{jc}}{p_c}
                +
                K
                \Bigg|
                    \frac{Y_{ic}Y_{jc}}{p_c}
                    -
                    \frac{A_{ij}}{d_i + 1}
                \Bigg|
            \Bigg)
            \!\cdot\!
            \| \bbX_{i,:} - \bbX_{j,:} \|_2
        \Bigg].
    &\label{eq:tagcn_error}}
\end{theorem}
As before, the effect of the graph connections give us an error bound for TAGCNs that relies on the homophily of the node labels $\bby$ over $\bbA$. 
Because the TAGCN layer in~\eqref{eq:tagcn_single_layer} requires aggregating node features within a $K$-hop neighborhood, where all hops are weighted equally, it is more critical that $\bby$ be homophilic or $\bbX$ be informative, as any discrepancies yield a higher TAGCN error bound in~\eqref{eq:tagcn_error} than they would in the GCN error bound in~\eqref{eq:gcn_error} due to the increased scale of $\big| \frac{Y_{ic}Y_{jc}}{p_c} - \frac{A_{ij}}{d_i + 1} \big|$ based on the number of hops $K$.
% Furthermore, note that its scale is augmented by $\| \bbDelta \|_F$, as we would expect since message passing over cross-class edges affect all neighbors in a $K$-hop radius, not just immediate neighbors as in~\eqref{eq:gcn_error}.
In addition, the $0$-th power of $\tbA_{\rm rw}$ in~\eqref{eq:tagcn_single_layer} is analogous to the presence of $\epsilon$ for the GIN layer in~\eqref{eq:gin_single_layer}, which emphasizes the role of the input node embeddings rather than those of neighboring nodes. 
Then, node features that are highly informative, that is, well separated across classes with low within-class variance, can reduce TAGCN error.

\subsection{GNN Performance under Label Heterophily}\label{Ss:hetgnn_performance}

The performance of all three GNNs thus far, the GCN in~\eqref{eq:gcn_single_layer}, the GIN in~\eqref{eq:gin_single_layer}, and the TAGCN in~\eqref{eq:tagcn_single_layer}, exhibit lower error for node classification when the node labels $\bby$ are homophilic over the graph $\bbA$.
However, it is not uncommon in real-world graph data for the labels $\bby$ to be heterophilic, where we expect that edges indicate nodes that belong to {\it different} classes, where $A_{ij} \neq 0$ is more likely to imply that $y_i \neq y_j$, a known challenge for message-passing GNNs~\cite{zhu2020homophily}.
We thus present similar results for GNNs under the assumption that the node labels are {\it heterophilic} over the graph.
As there are a plethora ways for $\bby$ to be heterophilic over $\bbA$, we present our theoretical framework in ways that suit a few particular, prominent scenarios of heterophilic node labels.

First, one approach to address label heterophily is neighborhood discovery~\cite{tenorio2025adaptingheterophilicgraphdata}, which corresponds to finding a new graph on which $\bby$ is homophilic.
Naturally, our original analyses from Sections~\ref{S:nfi} and~\ref{S:permtest} as well as Theorems~\ref{thm:gin_error} and~\ref{thm:tagcn_error} apply to this situation if we let the given graph $\bbA$ instead denote the newly obtained one.
Second, even if the edges in $\bbA$ do not directly indicate which nodes belong to the same class, the graph itself may still be informative.
For example, we may have that (i) nodes with shared labels may be far apart on the graph or (ii) cross-class edges may be more informative than within-class edges.
In both of these scenarios, naively applying a GNN with direct message passing for adjacent nodes may hinder classification accuracy, but we will discuss other ways we can still exploit the graph structure to improve performance.
In either case, we need only redefine the ideal graph $\bbA^*$ for our prior analyses to be applied, which we discuss in the sequel.

For the former, we consider an ideal graph $\bbA^*$ such that neighboring nodes at particular hops are most relevant.
We let $\tbDelta^{(k)} := \tbA^{k}_{\rm rw} - \tbA^{*k}_{\rm rw}$ denote the difference in random-walk adjacency matrices at the $k$-th power for $k \in \naturals$.
Then, we can define $\bbA^*$ and $\bba\in\reals^{K}$ such that 
\alna{
    \big\|
        \tbDelta^{(k)}
    \big\|_F
    &~\leq~&
    a_k
    \quad
    \forall~
    k\in [K],
\label{eq:aagcn_ideal_graph}}
where the magnitude of $a_k \geq 0$ describes how informative $k$-hop neighbors are for node classification.
% $\| \tbDelta^{(k)} \|_F \leq a_k$ for each hop $k\in[K]$.
As our GNN of choice for this setting, we exploit a modification of the TAGCN layer, 
\alna{
    \bbZ^{(\ell)}
    &~=~&
    \sigma_{\ell}
    \Bigg(
        \sum_{k=0}^K
        b_k \tbA_{\rm rw}^k
        \bbZ^{(\ell-1)}
        \bbTheta^{(\ell)}
    \Bigg)
    \quad
    % \forall~\ell \in [L],
\label{eq:aagcn_single_layer}}
which aggregates node features across neighbors at different hops, but each is assigned a different weight from $\bbb := [b_0,\dots, b_K]^\top \in \reals^{K+1}$, which reflect the learned emphasis of the model at each hop.
The use of matrix polynomials is common for allowing a message-passing GNN to adapt to long-range interactions.
Our particular formulation in~\eqref{eq:aagcn_single_layer} corresponds to an instantiation of a filter-bank GCN~\cite{gama2020GraphsConvolutionsNeuralNetworksGraph} or AAGCN~\cite{rey2025RedesigningGraphFilterBased}, as well as being similar in concept to other polynomial-based models such as APPNP~\cite{gasteiger2019Predict} or ChebNet~\cite{luan2024WhenGraphNeural}.
We can then obtain a performance bound for the polynomial-based GCN in~\eqref{eq:aagcn_single_layer} as follows.
\begin{theorem}\label{thm:aagcn_error}
    Let $f: \reals^{N\times M} \rightarrow \reals^{N\times H}$ be an $L$-layer modified TAGCN with $\bbZ^{(0)} = \bbX$, $\bbZ = \bbZ^{(L)} = f(\bbX; \bbA, \bbTheta)$, and layers defined in~\eqref{eq:aagcn_single_layer} for all $\ell \in [L]$
    for learnable weights $\bbb \in \reals^{K+1}$, matrix polynomial order $K \in \naturals$, $\tau$-Lipschitz activations $\sigma_{\ell}$, and weights $\bbTheta = \{ \bbTheta^{(\ell)} \}_{\ell=1}^L$ such that $\| \bbTheta^{(\ell)} \|_2 \leq \omega$ for $\ell \in [L]$.
    Then, with $\bbZ^* = f(\bbX^*; \bbA^*, \bbTheta)$ for $\bbX^*$ in~\eqref{eq:ideal_feats}, $\bbA^*$ satisfying~\eqref{eq:aagcn_ideal_graph}, and $\tbDelta^{(k)} := \tbA^k_{\rm rw} - \tbA^{*k}_{\rm rw}$ such that $\|\tbDelta^{(k)}\|_F\leq a_k$ for every $k\in[K]$,
    \alna{
        \| \bbZ^* - \bbZ \|_F
        &~\leq~&
        \tau^{L} \omega^{L}
        (\bbb^\top \bbone)^{L-1}
        \!
        \Bigg[
            L
            \|\bbX\|_F
            \sum_{k=1}^K
            a_k
            b_k
        % &\nonumber\\&
        % % \!\!\!\!\!\!
        %     % ~~
        %     &&
            +
            % \tau^2 \omega^2
            (\bbb^\top \bbone)
            \sum_{c=1}^C
            \sum_{i=1}^N
            \sum_{j=1}^N
                \frac{Y_{ic} Y_{jc}}{p_c}
            \!\cdot\!
            \| \bbX_{i,:} - \bbX_{j,:} \|_2
        \Bigg].
    &\label{eq:aagcn_error}}
\end{theorem}
Theorem~\ref{thm:aagcn_error} presents the performance for a polynomial-based GCN that can adapt to long-range interactions for node classification.
If we train a modified TAGCN in~\eqref{eq:aagcn_single_layer} such that $b_k$ is small when $a_k$ is high, then we can reduce the error bound in~\eqref{eq:aagcn_error}.
This reflects the effectiveness of such GNNs.
Unlike the error bound for GCNs in~\eqref{eq:gcn_error}, we can reduce the dependence of model performance of directly adjacent neighbors by reducing $b_k$ for lower $k$ while emphasizing higher-order neighbors by encouraging larger $b_k$ for higher $k$.

Finally, we exemplify the scenario where $\bbA$ contains heterophilic edges, but their presence can be informative for node classification.
In particular, we will provide an example for which a GCN can achieve a lower error when we have more heterophilic edges and fewer homophilic edges.
Given the number of classes $C$, let $\phi:[C]\rightarrow[C]$ be a bijective mapping between classes.
Then, we can define the ideal graph $\bbA^*$ such that
\alna{
    A_{ij}^*
    ~:=~
    A_{ij}
    \cdot
    \mbI\big( y_j = \phi(y_i) \big)
    \quad
    \forall ~
    i,j\in[N].
\label{eq:het_ideal_graph}}
Let us also consider a GCN without self-loops, that is, where the $\ell$-th layer is defined as
\alna{
    \bbZ^{(\ell)}
    &~=~&
    \sigma_{\ell}
    \Big(
        \bbA_{\rm rw} 
        \bbZ^{(\ell-1)}
        \bbTheta^{(\ell)}
    \Big)
    \quad
    \forall~\ell \in [L]
\label{eq:hetgcn_single_layer}}
for the random-walk adjacency matrix $\bbA_{\rm rw} = \bbD^{-1} \bbA$ {\it without} self-loops.
Recall that the rows of $\bbX^* = \bbY\bbP^{-1} \bbY^\top$ as defined in~\eqref{eq:ideal_feats} are distinct across classes.
Then, for any $k\in\naturals$, the transformation $\tbA^{*k}_{\rm rw} \bbX^*$ will still be distinct across classes.
For instance, if $C=2$, then $\bbA^*$ corresponds to a bipartite graph, and every odd power of $\bbA^*_{\rm rw}$ sets the node embeddings of class $c=1$ as the average features of class $c=2$ and vice versa.
Thus, applying the GCN defined by~\eqref{eq:hetgcn_single_layer} can achieve lower node classification error as we observe more heterophilic edges in $\bbA$, as shown formally in the following theorem.
\begin{theorem}\label{thm:hetgcn_error}
    Let $f: \reals^{N\times M} \rightarrow \reals^{N\times H}$ be an $L$-layer GCN without self-loops with $\bbZ^{(0)} = \bbX$, $\bbZ = \bbZ^{(L)} = f(\bbX; \bbA, \bbTheta)$, and layers defined in~\eqref{eq:hetgcn_single_layer} for all $\ell \in [L]$
    for $\tau$-Lipschitz activations $\sigma_{\ell}$ and weights $\bbTheta = \{ \bbTheta^{(\ell)} \}_{\ell=1}^L$ such that $\| \bbTheta^{(\ell)} \|_2 \leq \omega$ for $\ell \in [L]$.
    Then, with $\bbZ^* = f(\bbX^*; \bbA^*, \bbTheta)$ for $\bbX^*$ in~\eqref{eq:ideal_feats}, $\bbA^*$ in~\eqref{eq:het_ideal_graph}, and $\bbDelta := \bbA - \bbA^*$,
    \alna{
        \| \bbZ^* - \bbZ \|_F
        &~\leq~&
        \tau^{L} \omega^{L}
        \!
        \Bigg[
            (L-1)
            (1 + \sqrt{N})
            \|\bbDelta\|_F
            \|\bbX\|_F
        % &\nonumber\\&
        % % \!\!\!\!\!\!
        %     % ~~
        %     &&
            +
            % \tau^2 \omega^2
            \sum_{c=1}^C
            \sum_{i=1}^N
            \sum_{j=1}^N
                \left|
                    \frac{Y_{ic}Y_{j\phi(c)}}{p_{\phi(c)}}
                    -
                    \frac{A_{ij}}{d_i}
                \right|
                \!\cdot\!
                \| \bbX_{j,:} \|_2
        \Bigg].
    &\label{eq:hetgcn_error}}
\end{theorem}
According to Theorem~\ref{thm:hetgcn_error}, we observe a {\it larger} error bound if we have edges between nodes in the {\it same} class.
Note that since we do not implement self-loops with the GCN in~\eqref{eq:hetgcn_single_layer}, its performance does not depend on similarities between node features.
By our result in~\eqref{eq:hetgcn_error}, if nodes in class $c$ are connected to one and only one class $\phi(c)$, and no other class connects to class $c$, then we observe a lower error for GCN predictions.
Critically, this indicates that there are scenarios where heterophilic graphs are feasible and could even be preferable for GCNs, as long as the nodes do not connect to more than one class, even if the class they connect to is not their own.
Our results in Theorems~\ref{thm:aagcn_error} and~\ref{thm:hetgcn_error} thus exemplify extensions of our original analysis to heterophilic graph data.

\subsection{GNN Performance under Permutations}\label{ss:pert_gnn_error}

In addition to the above GNN error bounds given the original adjacency matrix $\bbA$ and features $\bbX$, we also extend our results to consider model performance under feature permutations, analogously to Theorem~\ref{thm:perm_gcn_error}.
For the GIN in Section~\ref{Ss:gin_performance}, the TAGCN in Section~\ref{Ss:tagcn_performance}, and the modified TAGCN in Section~\ref{Ss:hetgnn_performance}, observe that the error bounds under perturbed node features $\tbX$ as defined in Section~\ref{S:nfi} follow directly from the steps in the proof of Theorem~\ref{thm:perm_gcn_error}, which give a probabilistic bound for the feature difference between any two nodes from the permuted feature matrix $\| \tbX_{i,:} - \tbX_{j,:} \|_2$.
Thus, we have the following result for GIN performance under feature permutations.

\begin{corollary}\label{cor:perm_gin_error}
    Consider $\tbX \in \reals^{N\times M}$ such that $\tbX_{i,:} = \bbX_{\pi(i),:}$ for all $i\in[N]$ and some permutation $\bbpi \in \Pi$ chosen randomly.
    For the GIN in~\eqref{eq:gin_single_layer}, let $\tbZ^* = f(\tbX^*; \bbA^*, \bbTheta)$ for $\tbX^* := \bbY\bbP^{-1} \bbY^\top \tbX$, $\bbA^*$ in~\eqref{eq:ideal_A}, and $\bbDelta := \bbA - \bbA^*$.
    Then, with $\alpha$ and $\gamma$ defined as in Theorem~\ref{thm:perm_gcn_error} and with probability at least $e^{-t^2/2+\log M}$ for any $t \geq \sqrt{2\log M}$,
    \alna{
        \!\!\!\!\!
        \| \tbZ^* - \tbZ \|_F
        \leq
        % \tau^2 \omega^2
        % \!
        \tau^{L} \omega^{L}
        (1 + \epsilon)^{L-1}
        \Big[
            % \tau^2 \omega^2
            % \tau^{L} \omega^{L}
            (L-1)
            (1 + \sqrt{N})
            \|\bbDelta\|_F \|\bbX\|_F
        % &\nonumber\\&
        %     \qquad\qquad\quad
            +
            % \tau^2 \omega^2
            \sqrt{\gamma}
            \big\| \vect\big((1+\epsilon)\bbY\bbP^{-1}\bbY^\top - \tilde{\bbD}^{-1}\bbA\big) \big\|_1
        \Big].
    % &\nonumber\\&
    %     \!\!\!\!\!
    %     \text{where}~
    %     \gamma
    %     % \,:=\,
    %     :=
    %     % \frac{2}{N-1}
    %     {\textstyle\frac{2}{N-1}}
    %     \Big(
    %         \| \bbX \|_F^2
    %         -
    %         % \frac{1}{N}
    %         {\textstyle\frac{1}{N}}
    %         \| \bbX^\top \bbone \|_2^2
    %     \Big)
    %     +
    %     \alpha t M\sqrt{N}.
    % &\nonumber\\[-.25cm]& 
    \label{eq:perm_gin_error}}
    % for $\gamma$ defined in~\eqref{eq:pert_gcn_error}.
\end{corollary}
The primary difference in conclusion between Corollary~\ref{cor:perm_gin_error} and the result for GCNs in Theorem~\ref{thm:perm_gcn_error} is the presence of $\epsilon$.
When $\epsilon$ is large, reducing the error bound in~\eqref{eq:perm_gin_error} may require a larger $p_{y_i}$ relative to $d_i$, or equivalently, the number of nodes in a class $c$ is expected to outweight the degrees of the nodes in that class.
This corresponds to a GIN emphasizing the importance of node features rather than graph connectivity.
Similarly, we provide a result for TAGCN performance under node feature permutations.
\begin{corollary}\label{cor:perm_tagcn_error}
    Consider $\tbX \in \reals^{N\times M}$ such that $\tbX_{i,:} = \bbX_{\pi(i),:}$ for all $i\in[N]$ and some permutation $\bbpi \in \Pi$ chosen randomly.
    For the TAGCN in~\eqref{eq:tagcn_single_layer}, let $\tbZ^* = f(\tbX^*; \bbA^*, \bbTheta)$ for $\tbX^* := \bbY\bbP^{-1} \bbY^\top \tbX$, $\bbA^*$ in~\eqref{eq:ideal_A}, and $\bbDelta := \bbA - \bbA^*$.
    Then, with $\alpha$ and $\gamma$ defined as in Theorem~\ref{thm:perm_gcn_error} and with probability at least $e^{-t^2/2+\log M}$ for any $t \geq \sqrt{2\log M}$,
    \alna{
        \| \tbZ^* - \tbZ \|_F
        &~\leq~&
        \tau^{L} \omega^{L}
        (K+1)^{L-1}
        \!
        \Bigg[
            % \tau^2 \omega^2
            {\textstyle \frac{(K-1)(K-2)}{2}}
            \!\cdot\!
            L
            (1 + \sqrt{N})
            \|\bbDelta\|_F \|\bbX\|_F
        &\nonumber\\&
        % \!\!\!\!\!\!
            % ~~
            &&
            \quad\qquad\qquad
            +
            \sqrt{\gamma}
            \bigg(
                N + 
                \Big(
                    1 + 
                    (1 + \sqrt{N})
                    (K-1)
                    \| \bbDelta \|_F
                \Big)
                \Big\| \vect\big(\bbY\bbP^{-1}\bbY^\top - \tilde{\bbD}^{-1}\bbA\big) \Big\|_1
            \bigg)
            % \tau^2 \omega^2
            % \Bigg(
            %     \frac{Y_{ic} Y_{jc}}{p_c}
            %     +
            %     \Big(
            %         1 + 
            %         (1 + \sqrt{N})
            %         (K-1)
            %         \| \bbDelta \|_F
            %     \Big)
            %     \Bigg|
            %         \frac{Y_{ic}Y_{jc}}{p_c}
            %         -
            %         \frac{A_{ij}}{d_i + 1}
            %     \Bigg|
            % \Bigg)
        \Bigg].
    &\label{eq:perm_tagcn_error}}
    % for $\gamma$ defined in~\eqref{eq:pert_gcn_error}.
\end{corollary}
The conclusions of Corollary~\ref{cor:perm_tagcn_error} regarding the overall versus within-class variance of $\bbX$ is similar to our discussion for Theorem~\ref{thm:perm_gcn_error}.
However, a critical difference for the TAGCN in Corollary~\ref{cor:perm_tagcn_error} is that the increase in the error bound due to label heterophilc is more dramatic in~\eqref{eq:perm_tagcn_error}, dependent on the number of nodes via $(1 + \sqrt{N})$, the matrix polynomial order via $(K-1)$, and the presence of cross-class edges via~$\|\bbDelta\|_F$.
Indeed, as node features are propagated within a $K$-hop radius, perturbations such as $\tbX$ will have a compoundingly large effect as the number of hops increases.
We find a similar result in the following corollary for the modified TAGCN in~\eqref{eq:aagcn_single_layer} applied to heterophilic data.
\begin{corollary}\label{cor:perm_aagcn_error}
    Consider $\tbX \in \reals^{N\times M}$ such that $\tbX_{i,:} = \bbX_{\pi(i),:}$ for all $i\in[N]$ and some permutation $\bbpi \in \Pi$ chosen randomly.
    For the modified TAGCN in~\eqref{eq:aagcn_single_layer}, let $\tbZ^* = f(\tbX^*; \bbA^*, \bbTheta)$ for $\tbX^* := \bbY\bbP^{-1} \bbY^\top \tbX$, $\bbA^*$ such that~\eqref{eq:aagcn_ideal_graph}, and $\bbDelta := \bbA - \bbA^*$.
    Then, with $\alpha$ and $\gamma$ defined as in Theorem~\ref{thm:perm_gcn_error} and with probability at least $e^{-t^2/2+\log M}$ for any $t \geq \sqrt{2\log M}$,
    \alna{
        \| \tbZ^* - \tbZ \|_F
        &~\leq~&
        \tau^{L} \omega^{L}
        (\bbb^\top \bbone)^{L-1}
        % \!
        \Bigg[
            L \|\bbX\|_F
            \sum_{k=1}^K
            a_k b_k
            +
            N
            \sqrt{\gamma}
        \Bigg].
    &\label{eq:perm_aagcn_error}}
    % for $\gamma$ defined in~\eqref{eq:pert_gcn_error}.
\end{corollary}
Thus, the error bound of the modified TAGCN under feature permutations still depends on the discrepancy between the true and observed random-walk adjacency matrices at different powers $\tbA^k_{\rm rw} - \tbA^{*k}_{\rm rw}$, the errors of which are scaled the learned weights $\bbb$.
Thus, $\bbb$ can still mitigate the increase in error due to permuting the features $\bbX$, but the presence of $\sqrt{\gamma}$ reflects the potential source of error due to permuting features that are well separated across classes. 

Finally, we present the performance of the GCN without self-loops in~\eqref{eq:hetgcn_single_layer} under node feature permutations.
\begin{corollary}\label{cor:perm_hetgcn_error}
    Consider $\tbX \in \reals^{N\times M}$ such that $\tbX_{i,:} = \bbX_{\pi(i),:}$ for all $i\in[N]$ and some permutation $\bbpi \in \Pi$ chosen randomly.
    For the GCN without self-loops in~\eqref{eq:hetgcn_single_layer}, let $\tbZ^* = f(\tbX^*; \bbA^*, \bbTheta)$ for $\tbX^* := \bbY\bbP^{-1} \bbY^\top \tbX$, $\bbA^*$ in~\eqref{eq:het_ideal_graph}, and $\bbDelta := \bbA - \bbA^*$.
    Then, with $\alpha$ and $\gamma$ defined as in Theorem~\ref{thm:perm_gcn_error} and with probability at least $e^{-t^2/2+\log M}$ for any $t \geq \sqrt{2\log M}$,
    \alna{
        \| \tbZ^* - \tbZ \|_F
        ~\leq~
        \tau^{L} \omega^{L}
        % \!
        \Bigg[
            (L-1)
            (1 + \sqrt{N})
            \| \bbDelta \|_F
            \|\bbX\|_F
            +
            \sqrt{\xi}
            \cdot
            \Big\| \vect\big(\bbY\bbPhi\bbP^{-1}\bbY^\top - \bbD^{-1}\bbA\big) \Big\|_1
        \Bigg],
    &\nonumber\\&
        \text{where}~
        \xi
        :=
        \frac{1}{N} \|\bbX\|_F^2 + \alpha M t \sqrt{N}
    &\label{eq:perm_hetgcn_error}}
    and $\bbPhi \in \{0,1\}^{C\times C}$ is the permutation matrix such that $\Phi_{ij} := \mbI(y_i = \phi(y_j))$.
\end{corollary}
Thus, the bound for the GCN applied for node classification with heterophilic node labels closely resembles that of the GCN applied to homophilic data in Theorem~\ref{thm:gcn_error}.
However, since we do not apply self-loops, the error bound in~\eqref{eq:perm_hetgcn_error} does not rely on the variance of the node features $\bbX$ as in the previous results, as we combine the embeddings of strictly neighboring nodes.
Then, the worsening error bound due to homophilic edges depends on the overall magnitude of the node features.

\section{Proofs of Additional Results}\label{app:gnn_bound_proofs}

We provide proofs for the results shared in Appendix~\ref{app:gnn_performance}.
Corollaries~\ref{cor:perm_gin_error},~\ref{cor:perm_tagcn_error}, and~\ref{cor:perm_aagcn_error} are trivial and merely require the result in the proof of Theorem~\ref{thm:perm_gcn_error}, while we provide brief steps to attain an analogous result for Corollary~\ref{cor:perm_hetgcn_error}.

% \begin{proof}[Proof of Theorem~\ref{thm:gin_error}]
% \end{proof}
\subsection{Proof of Theorem~\ref{thm:gin_error}}\label{Ss:gin_bound_proof}

Just as for the proof of Theorem~\ref{thm:gcn_error}, we first bound the difference $\|\bbZ^{*(1)} - \bbZ^{(1)}\|_F$ for the first layer then recursively for the $\ell$-th layer $\| \bbZ^{*(\ell)} - \bbZ^{(\ell)} \|_F$.
First, by the definition of the GIN layer in~\eqref{eq:gin_single_layer}, we have that
\alna{
    \big\| \bbZ^{*(1)} - \bbZ^{(1)} \big\|_F
    &~\leq~&
    \tau \omega
    \Big\|
        \big(
            \epsilon \bbI + \tbA^*_{\rm rw}
        \big)
        \bbX^*
        -
        \big(
            \epsilon \bbI + \tbA_{\rm rw}
        \big)
        \bbX
    \Big\|_F
&\nonumber\\&
    &~=~&
    \tau \omega
    \Big\|
        (1 + \epsilon)
        \bbX^*
        -
        \big(
            \epsilon \bbI + \tbA_{\rm rw}
        \big)
        \bbX
    \Big\|_F
\nonumber}
since $\bbX^* = \tbA^*_{\rm rw} \bbX^*$.
Then, we add and subtract $\bbX$ inside the norm for
\alna{
    \big\| \bbZ^{*(1)} - \bbZ^{(1)} \big\|_F
    &~\leq~&
    \tau \omega
    \Big\|
        (1 + \epsilon)(\bbX^*-\bbX)
        +
        \bbX
        -
        \tbA_{\rm rw} \bbX
    \Big\|_F.
\nonumber}
With the definition of $\tbA_{\rm rw} = \tbD^{-1} (\bbI + \bbI)$, we have that
\alna{
    \big\| \bbZ^{*(1)} - \bbZ^{(1)} \big\|_F
    &~\leq~&
    \tau \omega
    \Big\|
        (1 + \epsilon)(\bbX^*-\bbX)
        +
        \tbD^{-1}
        (\bbD + \bbI - \bbA - \bbI)
        \bbX
    \Big\|_F
&\nonumber\\&
    &~=~&
    \tau \omega
    \Big\|
        (1 + \epsilon)(\bbX^*-\bbX)
        -
        \tbD^{-1}
        (\bbA - \bbD)
        \bbX
    \Big\|_F
&\nonumber\\&
    &~=~&
    \tau \omega
    \Big\|
        (1 + \epsilon)
        (\bbY\bbP^{-1}\bbY^\top - \bbI)
        \bbX
        -
        \tbD^{-1}
        (\bbA - \bbD)
        \bbX
    \Big\|_F
&\nonumber\\&
    &~=~&
    \tau \omega
    \sum_{i=1}^N
    \sqrt{
        \Bigg\|
            (1+\epsilon)
            \bigg(
                \frac{1}{p_{y_i}}[\bbY\bbY^\top]_{i,:}
                \bbX
                -
                \bbX_{i,:}
            \bigg)
            -
            \frac{1}{d_i+1}
            \Big(
                \bbA_{i,:}\bbX
                -
                d_i
                \bbX_{i,:}
            \Big)
        \Bigg\|_2^2
    }
&\nonumber\\&
    &~=~&
    \tau \omega
    \sum_{i=1}^N
    \sqrt{
        \Bigg\|
            \sum_{j=1}^N
            (1+\epsilon)
            \bigg[
                \frac{1}{p_{y_i}}
                \Big(
                    [\bbY\bbY^\top]_{ij}
                    \bbX_{j,:}
                    -
                    [\bbY\bbY^\top]_{i,j}
                    \bbX_{i,:}
                \Big)
                -
                \frac{1}{d_i+1}
                \Big(
                    A_{ij}\bbX_{j,:}
                    -
                    A_{ij}
                    \bbX_{i,:}
                \Big)
            \bigg]
        \Bigg\|_2^2
    }
&\nonumber\\&
    &~\leq~&
    \tau \omega
    \sum_{i=1}^N
    \sum_{j=1}^N
    \Bigg\|
        (1+\epsilon)
        \bigg[
                \frac{1}{p_{y_i}}
                [\bbY\bbY^\top]_{i,j}
            \big(
                \bbX_{j,:}
                -
                \bbX_{i,:}
            \big)
            -
            \frac{1}{d_i+1}
            A_{ij}
            \big(
                \bbX_{j,:}
                -
                \bbX_{i,:}
            \big)
        \bigg]
    \Bigg\|_2,
\nonumber}
where the final inequality is a straightforward application of the triangle inequality.
Thus, we have that
\alna{
    \big\| \bbZ^{*(1)} - \bbZ^{(1)} \big\|_F
    &~\leq~&
    \tau \omega
    \sum_{c=1}^C
    \sum_{i=1}^N
    \sum_{j=1}^N
    \Bigg|
        (1+\epsilon)
        \frac{Y_{ic}Y_{jc}}{p_c}
        -
        \frac{A_{ij}}{d_i+1}
    \Bigg|
    \cdot
    \big\|
        \bbX_{i,:}
        -
        \bbX_{j,:}
    \big\|_2.
\label{eq:gin_layer1}}
This we combine with the bound on $\| \bbZ^{*(\ell)} - \bbZ^{*(\ell)} \|_F$ to obtain our desired result.
We first require a bound on $\|\bbZ^{*(\ell-1)}\|_F$, where we have that
\alna{
    \big\|
        \bbZ^{*(\ell-1)}
    \big\|_F
    &~\leq~&
    \tau \omega
    \Big\|
        (\epsilon \bbI + \tbA_{\rm rw}^*)
        \bbZ^{*(\ell-2)}
    \Big\|_F
    =
    \tau \omega
    (1 + \epsilon)
    \big\|
        \bbZ^{*(\ell-2)}
    \big\|_F
    \leq
    \big( \tau \omega (1 + \epsilon) \big)^{\ell-1}
    \| \bbX \|_F.
\label{eq:gin_single_var}}
Then, we obtain our bound for the output of the $\ell$-th layer as
\alna{
    \big\| \bbZ^{*(\ell)} - \bbZ^{(\ell)} \big\|_F
    &~\leq~&
    \tau \omega
    \Big\|
        (\epsilon \bbI + \tbA^*_{\rm rw}) \bbZ^{*(\ell-1)}
        -
        (\epsilon \bbI + \tbA_{\rm rw}) \bbZ^{(\ell-1)}
    \Big\|_F
&\nonumber\\&
    &~\leq~&
    \tau \omega
    \big\|
        (\tbA^*_{\rm rw} - \tbA_{\rm rw}) \bbZ^{*(\ell-1)}
    \big\|_F
    +
    \tau\omega
    \big\|
        (\epsilon \bbI + \tbA_{\rm rw}) 
        (\bbZ^{*(\ell-1)} - \bbZ^{(\ell-1)})
    \big\|_F
&\nonumber\\&
    &~\leq~&
    \tau\omega 
    (1 + \sqrt{N})
    \| \bbZ^{*(\ell-1)} \|_F
    +
    \tau\omega(1 + \epsilon)
    \| \bbZ^{*(\ell-1)} - \bbZ^{(\ell-1)} \|_F
&\nonumber\\&
    &~\leq~&
    \tau^{\ell}\omega^{\ell}
    (1 + \epsilon)^{\ell-1}
    (1 + \sqrt{N})
    \| \bbX \|_F
    +
    \tau\omega(1 + \epsilon)
    \| \bbZ^{*(\ell-1)} - \bbZ^{(\ell-1)} \|_F,
\nonumber}
where we apply~\eqref{eq:gin_single_var} for the last inequality.
Applied recursively, we have that
\alna{
    \big\| \bbZ^{*(\ell)} - \bbZ^{(\ell)} \big\|_F
    &~\leq~&
    \tau^{\ell}\omega^{\ell}
    (1 + \epsilon)^{\ell-1}
    (\ell-1)
    (1 + \sqrt{N})
    \| \bbX \|_F
    +
    \tau^{\ell-1}\omega^{\ell-1}
    (1 + \epsilon)^{\ell-1}
    \| \bbZ^{*(1)} - \bbZ^{(1)} \|_F,
\nonumber}
which, combined with~\eqref{eq:gin_layer1} for $\ell = L$, yields the result in~\eqref{eq:gin_error} as desired.

\subsection{Proof of Theorem~\ref{thm:tagcn_error}}\label{Ss:tagcn_bound_proof}

We proceed as for the proofs of Theorems~\ref{thm:gcn_error} and~\ref{thm:gin_error} by bounding the output error for the first TAGCN layer followed by that of an arbitrary $\ell$-th layer.
First, given the TAGCN layer in~\eqref{eq:tagcn_single_layer}, we obtain
\alna{
    \big\| \bbZ^{*(1)} - \bbZ^{(1)} \big\|_F
    &~\leq~&
    \tau \omega
    \Bigg\|
        \sum_{k=0}^K
        \Big(
            \tbA^{*k}_{\rm rw}
            \bbX^*
            -
            \tbA^{k}_{\rm rw}
            \bbX
        \Big)
    \Bigg\|_F
&\nonumber\\&
    &~=~&
    \tau \omega
    \Bigg\|
        \bbX^*-\bbX
        +
        \bbX^*-\tbA_{\rm rw}\bbX
        +
        \sum_{k=2}^K
        \Big(
            \tbA^{*k}_{\rm rw}
            \bbX^*
            -
            \tbA^{k}_{\rm rw}
            \bbX
        \Big)
    \Bigg\|_F
&\nonumber\\&
    &~\leq~&
    \tau \omega
    \big\|
        \bbX^* - \bbX
    \big\|_F
    +
    \tau \omega
    \Bigg\|
        \bbX^*-\tbA_{\rm rw}\bbX
        +
        \sum_{k=2}^K
        \Big(
            \tbA^{*k}_{\rm rw}
            \bbX^*
            -
            \tbA^{k}_{\rm rw}
            \bbX
        \Big)
    \Bigg\|_F
\label{eq:tagcn_layer1_init}}
by the triangle inequality.
Then, we can rewrite the sum of differences as
\alna{
    \sum_{k=2}^K
    \Big(
        \tbA_{\rm rw}^{*k}
        \bbX^*
        -
        \tbA_{\rm rw}^k
        \bbX
    \Big)
    &~=~&
    \sum_{k=2}^K
    \Big(
        \big(
            \tbA_{\rm rw}^{*k-1}
            -
            \tbA_{\rm rw}^{k-1}
        \big)
        \tbA_{\rm rw}^*
        \bbX^*
        +
        \tbA_{\rm rw}^{k-1}
        \tbA_{\rm rw}^*
        \bbX^*
        -
        \tbA_{\rm rw}^k
        \bbX
    \Big)
&\nonumber\\&
    &~=~&
    \sum_{k=2}^K
    \Big(
        \big(
            \tbA_{\rm rw}^{*k-1}
            -
            \tbA_{\rm rw}^{k-1}
        \big)
        \bbX^*
        +
        \tbA_{\rm rw}^{k-1}
        \big(
            \bbX^* - \tbA_{\rm rw}\bbX
        \big)
    \Big).
\label{eq:sum_diff_pows}}
% Furthermore, recall that $\tbDelta^{(1)} = \tbA_{\rm rw} - \tbA_{\rm rw}^*$.
For convenience, we let $\barbDelta := \tbA^*_{\rm rw} - \tbA_{\rm rw}$.
Then, we have that
\alna{
    \sum_{k=2}^K
    \Big(
        \tbA_{\rm rw}^{*k-1}
        -
        \tbA_{\rm rw}^{k-1}
    \Big)
    &~=~&
    \sum_{k=1}^{K-1}
    \Big(
        \tbA_{\rm rw}^{*k}
        -
        \tbA_{\rm rw}^{k}
    \Big)
&\nonumber\\&
    &~=~&
    \sum_{k=1}^{K-1}
    \Big(
        \tbA_{\rm rw}^*
        \tbA_{\rm rw}^{*k-1}
        -
        \tbA_{\rm rw}
        \tbA_{\rm rw}^{*k-1}
        +
        \tbA_{\rm rw}
        \tbA_{\rm rw}^{*k-1}
        -
        \tbA_{\rm rw}
        \tbA_{\rm rw}^{k-1}
    \Big)
&\nonumber\\&
    &~=~&
    \sum_{k=1}^{K-1}
    \bigg(
        \big(
            \tbA_{\rm rw}^* - \tbA_{\rm rw}
        \big)
        \tbA_{\rm rw}^{*k-1}
        +
        \tbA_{\rm rw}
        \Big(
            \tbA_{\rm rw}^{*k-1} - \tbA_{\rm rw}^{k-1}
        \Big)
    \bigg)
&\nonumber\\&
    &~=~&
    \sum_{k=1}^{K-1}
    \bigg(
        \barbDelta
        \tbA_{\rm rw}^{*k-1}
        +
        \tbA_{\rm rw}
        \Big(
            \tbA_{\rm rw}^{*k-1} - \tbA_{\rm rw}^{k-1}
        \Big)
    \bigg)
&\nonumber\\&
    &~=~&
    \barbDelta
    +
    \sum_{k=2}^{K-1}
    \bigg(
        \barbDelta
        \tbA_{\rm rw}^{*k-1}
        +
        \tbA_{\rm rw}
        \Big(
            \tbA_{\rm rw}^{*k-1} - \tbA_{\rm rw}^{k-1}
        \Big)
    \bigg)
&\nonumber\\&
    &~=~&
    \barbDelta
    +
    \sum_{k=2}^{K-1}
    \bigg(
        \barbDelta
        \tbA_{\rm rw}^{*k-1}
        +
        \tbA_{\rm rw}
        \Big(
            \barbDelta
            \tbA_{\rm rw}^{*k-2}
            +
            \tbA_{\rm rw}
            \big(
                \tbA_{\rm rw}^{*k-2} - \tbA_{\rm rw}^{k-2}
            \big)
        \Big)
    \bigg),
\nonumber}
which we can continue to expand for
\alna{
    \sum_{k=2}^K
    \Big(
        \tbA_{\rm rw}^{*k-1}
        -
        \tbA_{\rm rw}^{k-1}
    \Big)
    &~=~&
    \sum_{j=0}^{K-2}
    \sum_{k=0}^{K-j-2}
    \tbA_{\rm rw}^k
    \barbDelta
    \tbA_{\rm rw}^{*j}.
\nonumber}
This substituted into~\eqref{eq:sum_diff_pows} can subsequently be substituted into~\eqref{eq:tagcn_layer1_init} for
\alna{
    \big\| \bbZ^{*(1)} - \bbZ^{(1)} \big\|_F
    &~\leq~&
    \tau \omega
    \| \bbX^* - \bbX \|_F
    +
    \tau \omega
    \Bigg\|
        \bbX^* - \tbA_{\rm rw}\bbX
        +
        \sum_{k=2}^{K}
            \big( \tbA_{\rm rw}^{*k-1} - \tbA_{\rm rw}^{k-1} \big)
        \bbX^*
        +
        \sum_{k=2}^{K}
        \tbA_{\rm rw}^{k-1}
        \big(
            \bbX^* - \tbA_{\rm rw}\bbX
        \big)
    \Bigg\|_F
&\nonumber\\&
    &~=~&
    \tau \omega
    \| \bbX^* - \bbX \|_F
    +
    \tau \omega
    \Bigg\|
        \bbX^* - \tbA_{\rm rw}\bbX
        +
        \sum_{k=1}^{K-1}
            \big( \tbA_{\rm rw}^{*k} - \tbA_{\rm rw}^k \big)
        \bbX^*
        +
        \sum_{k=1}^{K-1}
        \tbA_{\rm rw}^k
        \big(
            \bbX^* - \tbA_{\rm rw}\bbX
        \big)
    \Bigg\|_F
&\nonumber\\&
    &~=~&
    \tau \omega
    \| \bbX^* - \bbX \|_F
    +
    \tau \omega
    \Bigg\|
        \sum_{k=1}^{K-1}
            \big( \tbA_{\rm rw}^{*k} - \tbA_{\rm rw}^k \big)
        \bbX^*
        +
        \sum_{k=0}^{K-1}
        \tbA_{\rm rw}^k
        \big(
            \bbX^* - \tbA_{\rm rw}\bbX
        \big)
    \Bigg\|_F
&\nonumber\\&
    &~=~&
    \tau \omega
    \| \bbX^* - \bbX \|_F
    +
    \tau \omega
    \Bigg\|
        \sum_{j=0}^{K-2}
        \sum_{k=0}^{K-j-2}
        \tbA_{\rm rw}^k
        \barbDelta
        \tbA_{\rm rw}^{*j}
        \bbX^*
        +
        \sum_{k=0}^{K-1}
        \tbA_{\rm rw}^k
        \big(
            \bbX^* - \tbA_{\rm rw}\bbX
        \big)
    \Bigg\|_F
&\nonumber\\&
    &~=~&
    \tau \omega
    \| \bbX^* - \bbX \|_F
    +
    \tau \omega
    \Bigg\|
        \sum_{j=0}^{K-2}
        \sum_{k=0}^{K-j-2}
        \tbA_{\rm rw}^k
        \barbDelta
        \bbX^*
        +
        \sum_{k=0}^{K-1}
        \tbA_{\rm rw}^k
        \big(
            \bbX^* - \tbA_{\rm rw}\bbX
        \big)
    \Bigg\|_F.
\nonumber}
We again employ the triangle inequality and $\| \tbA_{\rm rw} \|_2 = 1$ for
\alna{
    \big\| \bbZ^{*(1)} - \bbZ^{(1)} \big\|_F
    &~\leq~&
    \tau \omega
    \| \bbX^* - \bbX \|_F
    +
    \tau\omega
    \sum_{j=0}^{K-2}
    \sum_{k=0}^{K-j-2}
    \big\|
        \barbDelta
        \bbX^*
    \big\|_F
    +
    \tau\omega
    \sum_{k=0}^{K-1}
    \big\|
        \tbA_{\rm rw}^k
        \big(
            \bbX^* - \tbA_{\rm rw}\bbX
        \big)
    \big\|_F
&\nonumber\\&
    &~\leq~&
    \tau \omega
    \| \bbX^* - \bbX \|_F
    +
    \frac{K(K-1)}{2}
    \cdot
    \tau\omega
    (1 + \sqrt{N})
    \| \bbDelta \|_F
    \| \bbX \|_F
    +
    K
    \tau\omega
    \big\|
        \bbX^* - \tbA_{\rm rw}\bbX
    \big\|_F.
\nonumber}
Then, with the definition $\bbX^* = \bbY \bbP^{-1} \bbY^\top$, we apply similar steps as in the proofs of Theorems~\ref{thm:gcn_error} and~\ref{thm:gin_error} for
\alna{
    \big\| \bbZ^{*(1)} - \bbZ^{(1)} \big\|_F
    &~\leq~&
    \tau \omega
    \Bigg[
        {\textstyle\frac{K(K-1)}{2}}
        (1 + \sqrt{N})
        \| \bbDelta \|_F
        \| \bbX \|_F
        +
        \sum_{c=1}^C
        \sum_{i,j=1}^N
        \!
        \bigg[
            \frac{Y_{ic}Y_{jc}}{p_c}
            +
            K
            \bigg|
                \frac{Y_{ic}Y_{jc}}{p_c}
                -
                \frac{A_{ij}}{d_i + 1}
            \bigg|
        \bigg]
        \big\| \bbX_{i,:} - \bbX_{j,:} \big\|_2
    \Bigg].
\label{eq:tagcn_layer1}}
Then, to bound the difference in outputs for the $\ell$-th layer, we need the following output bound
\alna{
    \big\|
        \bbZ^{*(\ell-1)}
    \big\|_F
    &~\leq~&
    \tau\omega
    \Bigg\|
        \sum_{k=0}^K
        \tbA_{\rm rw}^{*k}
        \bbZ^{*(\ell-1)}
    \Bigg\|_F
    ~\leq~
    \tau\omega
    (K+1)
    \big\|
        \bbZ^{*(\ell-1)}
    \big\|_F
    ~\leq~
    \Big(
        \tau\omega (K+1)
    \Big)^{\ell-1}
    \| \bbX \|_F,
\nonumber}
which we apply for the following bound
\alna{
    \big\| \bbZ^{*(\ell)} - \bbZ^{(\ell)} \big\|_F
    &~\leq~&
    \tau \omega
    \Bigg\|
        \sum_{k=0}^K
        \Big(
            \tbA_{\rm rw}^{*k}
            \bbZ^{*(\ell-1)}
            -
            \tbA_{\rm rw}^k
            \bbZ^{(\ell-1)}
        \Big)
    \Bigg\|_F
&\nonumber\\&
    &~\leq~&
    \tau \omega
    \Bigg\|
        \sum_{k=0}^K
        \Big(
            \tbA_{\rm rw}^{*k}
            -
            \tbA_{\rm rw}^{k}
        \Big)
        \bbZ^{*(\ell-1)}
    \Bigg\|_F
    +
    \tau \omega
    \Bigg\|
        \sum_{k=0}^K
        \tbA_{\rm rw}^k
        \big(
            \bbZ^{*(\ell-1)}
            -
            \bbZ^{(\ell-1)}
        \big)
    \Bigg\|_F
&\nonumber\\&
    &~\leq~&
    \tau \omega
    \Bigg\|
        \sum_{k=1}^{K}
        \Big(
            \tbA_{\rm rw}^{*k}
            -
            \tbA_{\rm rw}^{k}
        \Big)
        \bbZ^{*(\ell-1)}
    \Bigg\|_F
    +
    \tau \omega
    (K+1)
    \big\|
        \bbZ^{*(\ell-1)} - \bbZ^{(\ell-1)}
    \big\|_F
&\nonumber\\&
    &~=~&
    \tau \omega
    \Bigg\|
        \sum_{j=0}^{K-1}
        \sum_{k=0}^{K-j-1}
        \tbA_{\rm rw}^k
        \barbDelta
        \tbA_{\rm rw}^{*j}
        \bbZ^{*(\ell-1)}
    \Bigg\|_F
    +
    \tau \omega
    (K+1)
    \big\|
        \bbZ^{*(\ell-1)} - \bbZ^{(\ell-1)}
    \big\|_F
&\nonumber\\&
    &~\leq~&
    {\textstyle\frac{K(K+1)}{2}} \cdot
    \tau \omega
    \big\|
        \barbDelta
        \bbZ^{*(\ell-1)}
    \big\|_F
    +
    \tau \omega
    (K+1)
    \big\|
        \bbZ^{*(\ell-1)} - \bbZ^{(\ell-1)}
    \big\|_F
&\nonumber\\&
    &~\leq~&
    {\textstyle\frac{K(K+1)}{2}} \cdot
    \tau \omega
    (1 + \sqrt{N})
    \| \bbDelta \|_F
    \big\| \bbZ^{*(\ell-1)} \big\|_F
    +
    \tau \omega
    (K+1)
    \big\|
        \bbZ^{*(\ell-1)} - \bbZ^{(\ell-1)}
    \big\|_F
&\nonumber\\&
    &~\leq~&
    {\textstyle\frac{K(K+1)}{2}} \cdot
    \tau^{\ell} \omega^{\ell}
    (K+1)^{\ell-1}
    (1 + \sqrt{N})
    \| \bbDelta \|_F
    \| \bbX \|_F
    +
    \tau \omega
    (K+1)
    \big\|
        \bbZ^{*(\ell-1)} - \bbZ^{(\ell-1)}
    \big\|_F.
\nonumber}
Then, applied recursively, we have that
\alna{
    \big\| \bbZ^{*(\ell)} - \bbZ^{(\ell)} \big\|_F
    &~\leq~&
    {\textstyle\frac{K(K+1)}{2}}
    \tau^\ell \omega^{\ell} (K+1)^{\ell-1}
    (\ell-1)
    (1 + \sqrt{N})
    \| \bbDelta \|_F
    \| \bbX \|_F
    +
    \Big(
        \tau \omega (K + 1)
    \Big)^{\ell-1}
    \big\|
        \bbZ^{*(1)} - \bbZ^{(1)}
    \big\|_F,
% &\nonumber\\&
%     &~\leq~&
%     {\textstyle\frac{K(K+1)}{2}}
%     \tau^\ell \omega^{\ell} (K+1)^{\ell-1}
%     \ell
%     (1 + \sqrt{N})
%     \| \bbDelta \|_F
%     \| \bbX \|_F
%     +
%     \tau^{\ell} \omega^{\ell}
%     (K + 1)^{\ell-1}
\nonumber}
which we can combine with~\eqref{eq:tagcn_layer1} for the bound in~\eqref{eq:tagcn_error}, as desired.

\subsection{Proof of Theorem~\ref{thm:aagcn_error}}\label{Ss:aagcn_bound_proof}

We first obtain a bound for the difference in outputs of the first layer, that is,~\eqref{eq:aagcn_single_layer} for $\ell=1$,
\alna{
    \big\| \bbZ^{*(1)} - \bbZ^{(1)} \big\|_F
    &~\leq~&
    \tau \omega
    \Bigg\|
        \sum_{k=0}^K
        b_k
        \Big(
            \tbA_{\rm rw}^{*k}
            \bbX^*
            -
            \tbA_{\rm rw}^{k}
            \bbX
        \Big)
    \Bigg\|_F
&\nonumber\\&
    &~\leq~&
    \tau \omega b_0
    \big\| \bbX^* - \bbX \big\|_F
    +
    \tau\omega
    \sum_{k=1}^K
    b_k
    \Big\|
        \tbA_{\rm rw}^{*k}
        \bbX^*
        -
        \tbA_{\rm rw}^{k}
        \bbX
    \Big\|_F
&\nonumber\\&
    &~\leq~&
    \tau \omega b_0
    \big\| \bbX^* - \bbX \big\|_F
    +
    \tau\omega
    \sum_{k=1}^K
    b_k
    \bigg(
        \Big\|
            \big(
                \tbA_{\rm rw}^{*k}
                -
                \tbA_{\rm rw}^{k}
            \big)
            \bbX^*
        \Big\|_F
        +
        \big\| \bbX^* - \bbX \big\|_F
    \bigg)
&\nonumber\\&
    &~\leq~&
    \tau \omega (\bbb^\top \bbone)
    \big\| \bbX^* - \bbX \big\|_F
    +
    \tau \omega
    \|\bbX\|_F
    \sum_{k=1}^K b_k
    \big\|
        \tbA_{\rm rw}^{*k}
        -
        \tbA_{\rm rw}^{k}
    \big\|_F
&\nonumber\\&
    &~\leq~&
    \tau \omega (\bbb^\top \bbone)
    \big\| \bbX^* - \bbX \big\|_F
    +
    \tau \omega
    \|\bbX\|_F
    \sum_{k=1}^K a_k b_k.
\label{eq:aagcn_layer1}}
Then, we obtain a bound for the output of the $\ell$-th layer
\alna{
    \big\|
        \bbZ^{*(\ell-1)}
    \big\|_F
    &~\leq~&
    \tau \omega
    \Bigg\|
        \sum_{k=0}^K
        b_k
        \tbA_{\rm rw}^{*k}
        \bbZ^{*(\ell-2)}
    \Bigg\|_F
    ~\leq~
    \tau\omega (\bbb^\top \bbone)
    \big\| \bbZ^{*(\ell-2)} \big\|_F
    ~\leq~
    \big(\tau\omega (\bbb^\top \bbone)\big)^{\ell-1}
    \| \bbX \|_F.
\nonumber}
With this bound, we bound the difference in outputs for an arbitrary layer as
\alna{
    \big\| \bbZ^{*(\ell)} - \bbZ^{(\ell)} \big\|_F
    &~\leq~&
    \tau \omega
    \Bigg\|
        \sum_{k=0}^K
        b_k
        \Big(
            \tbA_{\rm rw}^{*k}
            \bbZ^{*(\ell-1)}
            -
            \tbA_{\rm rw}^{k}
            \bbZ^{(\ell-1)}
        \Big)
    \Bigg\|_F
&\nonumber\\&
    &~\leq~&
    \tau\omega
    \sum_{k=1}^K
    b_k
    \Big\|
        \tbA_{\rm rw}^{*k}
        -
        \tbA_{\rm rw}^{k}
    \Big\|_F
    \big\|
        \bbZ^{*(\ell-1)}
    \big\|_F
    +
    \tau\omega
    \sum_{k=0}^K
    b_k
    \big\| \bbZ^{*(\ell-1)} - \bbZ^{(\ell-1)} \big\|_F
&\nonumber\\&
    &~\leq~&
    \tau^{\ell} \omega^{\ell}
    (\bbb^\top \bbone)^{\ell-1}
    \|\bbX\|_F
    \sum_{k=1}^K a_k b_k
    +
    \tau \omega
    (\bbb^\top \bbone)
    \big\| \bbZ^{*(\ell-1)} - \bbZ^{(\ell-1)} \big\|_F
&\nonumber\\&
    &~\leq~&
    \tau^{\ell} \omega^{\ell}
    (\bbb^\top \bbone)^{\ell-1}
    (\ell-1)
    \|\bbX\|_F
    \sum_{k=1}^K a_k b_k
    +
    \tau^{\ell-1} \omega^{\ell-1}
    (\bbb^\top \bbone)^{\ell-1}
    \big\| \bbZ^{*(1)} - \bbZ^{(1)} \big\|_F
\nonumber}
to which we apply~\eqref{eq:aagcn_layer1} for the result in~\eqref{eq:aagcn_error}.

\subsection{Proof of Theorem~\ref{thm:hetgcn_error}}\label{Ss:hetgcn_bound_proof}
The steps to prove Theorem~\ref{thm:hetgcn_error} are almost identical to those of Theorem~\ref{eq:gcn_error}, with slight differences due to the use of the random-walk adjacency matrix $\bbA_{\rm rw}$ without self-loops and the fact that the ideal graph $\bbA^*$ connects nodes in different classes according to the bijection $\phi$.
More specifically, we bound the different in outputs of the first layer for the GCN in~\eqref{eq:hetgcn_single_layer} as
\alna{
    \big\|
        \bbZ^{*(1)}
        -
        \bbZ^{(1)}
    \big\|_F
    &~\leq~&
    \tau\omega
    \big\|
        \bbA_{\rm rw}^*
        \bbX^*
        -
        \bbA_{\rm rw}
        \bbX
    \big\|_F
&\nonumber\\&
    &~\leq~&
    \tau \omega
    \sum_{i=1}^N
    \Big\|
        \big[
            \bbA_{\rm rw}^*
            \bbX^*
            -
            \bbA_{\rm rw}
            \bbX
        \big]_{i,:}
    \Big\|_2
&\nonumber\\&
    &~=~&
    \tau \omega
    \sum_{i=1}^N
    \Bigg\|
        \sum_{k=1}^N
        \frac{A^*_{ik}}{d_i^*} 
        \bbX_{k,:}^*
        -
        \sum_{j=1}^N
        \frac{A_{ij}}{d_i}
        \bbX_{j,:}
    \Bigg\|_F
&\nonumber\\&
    &~=~&
    \tau \omega
    \sum_{i=1}^N
    \Bigg\|
        \sum_{k=1}^N
        \sum_{j=1}^N
        \frac{A^*_{ik}}{d_i^*} 
        \frac{[\bbY\bbY^\top]_{kj}}{p_{y_k}}
        \bbX_{j,:}
        -
        \sum_{j=1}^N
        \frac{A_{ij}}{d_i}
        \bbX_{j,:}
    \Bigg\|_F
&\nonumber\\&
    &~=~&
    \tau \omega
    \sum_{i=1}^N
    \Bigg\|
        \sum_{k=1}^N
        \frac{A^*_{ik}}{d_i^*} 
        \sum_{j=1}^N
        \frac{ \mbI( y_j = \phi(y_i) ) }{p_{\phi(y_i)}}
        \bbX_{j,:}
        -
        \sum_{j=1}^N
        \frac{A_{ij}}{d_i}
        \bbX_{j,:}
    \Bigg\|_F
&\nonumber\\&
    &~=~&
    \tau \omega
    \sum_{i=1}^N
    \Bigg\|
        \sum_{j=1}^N
        \frac{ \mbI( y_j = \phi(y_i) ) }{p_{\phi(y_i)}}
        \bbX_{j,:}
        -
        \sum_{j=1}^N
        \frac{A_{ij}}{d_i}
        \bbX_{j,:}
    \Bigg\|_F
&\nonumber\\&
    &~\leq~&
    \tau \omega
    \sum_{i=1}^N
    \sum_{j=1}^N
    \Bigg|
        \frac{ \mbI( y_j = \phi(y_i) ) }{p_{\phi(y_i)}}
        -
        \frac{A_{ij}}{d_i}
    \Bigg|
    \| \bbX_{j,:} \|_2
&\nonumber\\&
    &~=~&
    \tau \omega
    \sum_{c=1}^C
    \sum_{i=1}^N
    \sum_{j=1}^N
    \Bigg|
        \frac{ Y_{ic} Y_{j\phi(c)} ) }{p_{\phi(c)}}
        -
        \frac{A_{ij}}{d_i}
    \Bigg|
    \| \bbX_{j,:} \|_2.
\label{eq:hetgcn_layer1}}
The bound for the output of the GCN without self-loops in~\eqref{eq:hetgcn_single_layer} is identical to that of the GCN in~\eqref{eq:gcn_output_bound}.
Thus, we proceed with bounding the output difference for the $\ell$-th layer,
\alna{
    \big\|
        \bbZ^{*(\ell)}
        -
        \bbZ^{(\ell)}
    \big\|_F
    &~\leq~&
    \tau\omega
    \big\|
        \bbA_{\rm rw}^* \bbZ^{*(\ell-1)}
        -
        \bbA_{\rm rw} \bbZ^{(\ell-1)}
    \big\|_F
&\nonumber\\&
    &~\leq~&
    \tau \omega
    \Big\|
        \big(
            \bbA_{\rm rw}^*
            -
            \bbA_{\rm rw}
        \big)
        \bbZ^{*(\ell-1)}
    \Big\|_F
    +
    \tau \omega
    \big\| \bbZ^{*(\ell-1)} - \bbZ^{(\ell-1)} \big\|_F
&\nonumber\\&
    &~\leq~&
    \tau \omega
    (1 + \sqrt{N})
    \| \bbDelta \|_F
    \Big\|
        \bbZ^{*(\ell-1)}
    \Big\|_F
    +
    \tau \omega
    \big\| \bbZ^{*(\ell-1)} - \bbZ^{(\ell-1)} \big\|_F
&\nonumber\\&
    &~\leq~&
    \tau^{\ell} \omega^{\ell}
    (1 + \sqrt{N})
    \| \bbDelta \|_F
    \| \bbX \|_F
    +
    \tau^{\ell-1} \omega^{\ell-1}
    \big\| \bbZ^{*(1)} - \bbZ^{(1)} \big\|_F,
\nonumber}
which we combine with~\eqref{eq:hetgcn_layer1} for our desired result in~\eqref{eq:hetgcn_error}.

\subsection{Proof of Corollary~\ref{cor:perm_hetgcn_error}}\label{Ss:perm_hetgcn_bound_proof}

For the bound in~\eqref{eq:perm_hetgcn_error}, we need only apply the same steps as for the proof of Theorem~\ref{thm:perm_gcn_error}, except we define
\alna{
    \psi(\bbpi)
    :=
    X_{\pi(i),m}^2.
\nonumber}
The steps of the proof proceed up to~\eqref{eq:Qdiff_bound},
for which we have a similar bound
\alna{
    \big|
        \psi(\bbpi) - \psi( (k\ell) \bbpi)
    \big|
    =
    \big|
        \tilde{X}_{im}^2
        -
        (\tilde{X}_{im}^{(k\ell)})^2
    \big|
    =
    \big|
        \tilde{X}_{im}
        -
        \tilde{X}_{im}^{(k\ell)}
    \big|
    \cdot
    \big|
        \tilde{X}_{im}
        +
        \tilde{X}_{im}^{(k\ell)}
    \big|
    \leq
    2\alpha.
\nonumber}
The remainder of the proof follows, where we instead have the expectation $\mbE[\psi(\bbpi)] = \frac{1}{N} \| \bbX_{:,m} \|_2^2$, yielding the error bound in~\eqref{eq:perm_hetgcn_error} with probability at least $1 - e^{-t^2/2 + \log M}$, as desired.

\section{Experimental Details}\label{app:exp_details}

This section provides further details regarding the simulations in this work, including the datasets employed.

\subsection{Dataset details}\label{app:data_info}

We share the statistics of the datasets used in our experiments in Table~\ref{tab:dataset_stats}.
Information about dataset context, that is, the interpretation of nodes, features, edges, and labels is provided in Section~\ref{S:sim}.

\begin{table}[t]
\centering
\footnotesize
\setlength{\tabcolsep}{5pt}
\caption{
% \ibmblue{arxiv} 
Statistics of the benchmark datasets in experiments, including the number of nodes, edges, features per node, and class labels.}
\vspace{-.2cm}
\begin{tabular}{lrrrr}
\toprule
\textbf{Dataset} & \textbf{\#Nodes} & \textbf{\#Edges} & \textbf{\#Feats} & \textbf{\#Classes} \\
\midrule
Cora        & 2{,}708  & 10{,}556 & 1{,}433 & 7 \\
CiteSeer    & 3{,}327  & 9{,}104  & 3{,}703 & 6 \\
PubMed      & 19{,}717 & 88{,}648 & 500     & 3 \\
\hline
Photo    & 7{,}650  & 119{,}043 & 745   & 8 \\
Computers& 13{,}752 & 245{,}778 & 767   & 10 \\
ogbn-arxiv& 169{,}343 & 1{,}166{,}243 & 128   & 40 \\
\hline
Cornell     & 183    & 298    & 1{,}703 & 5 \\
Texas       & 183    & 325    & 1{,}703 & 5 \\
Wisconsin   & 251    & 515    & 1{,}703 & 5 \\
\bottomrule
\end{tabular}
\label{tab:dataset_stats}
\end{table}

\subsection{Table~\ref{t:real_glob_perm} simulation details}\label{app:table1_info}

We elaborate on training details for the results in Table~\ref{t:real_glob_perm}.
All results are averaged over five runs, except for the results in Figures 2 and 5-8, which are averaged over twenty runs. 
In each run, we randomly split the nodes into 70\% training, 10\% validation, and 20\% test, and we report the test accuracy corresponding to the epoch with the highest validation accuracy. 
The train/validation/test masks are re-sampled independently for each run. 
For Cora, CiteSeer, and PubMed, we use a 2-layer GCN; for Amazon Computers and Photo we use a 2-layer GIN; for Cornell, Texas, and Wisconsin we use a 2-layer TAGCN; and for the MLP baseline we use a 2-layer MLP, all with 512 hidden units. 
Models on Computers and Photo are trained for 800 epochs, while the remaining datasets are trained for 400 epochs. 
We use the Adam optimizer with a learning rate of 0.01 and weight decay of $5\times 10^{-4}$.

\subsection{Table~\ref{t:real_fs} baselines and simulation details}\label{app:table2_info}

We next describe our process for the results in Table~\ref{t:real_fs}.
Our evaluation follows a two-stage pipeline: in the first stage, we train a model following the setup in Table~\ref{t:real_glob_perm} and compute feature importance scores using the validation set. 
We then select the top $r\%$ of features according to each FS method ($r=2\%$ for all datasets except PubMed, Photo, and Computers, where $r=5\%$ due to their smaller feature dimension). 
In the second stage, we retrain the model using only these selected features, with the same architecture and training configuration as in the first stage, and report the test accuracy.

As for baselines, we evaluate several feature selection (FS) methods listed below. 

\begin{itemize}
    \item {\bf NPT:} Our {\bf node feature permutation tests (NPTs)} for feature importance ranks features based on the drop in validation accuracy upon permuting each feature. 
    \item {\bf NPT-gaussian:} We introduce a variant of {\bf NPT} where, rather than permuting a feature to remove its effect, we instead replace it with i.i.d. Gaussian random variables with the same mean and variance as the original feature.
    \item {\bf NPT-mask:} This is another variant of {\bf NPT} where, rather than permuting a feature to remove its effect, we instead mask its values, that is, set all of its values to zero.
    \item {\bf MI:} We measure the mutual information (MI) between each feature and the node labels.
    \item {\bf TFI:} The {\bf Topological Feature Informativeness (TFI)} metric was introduced in \citep{zheng2025letyourfeatures} to measure feature importance prior to training to be applicable for GCNs. 
    \item {\bf Feature homophily:} The homophily-based metrics, $h_{\rm attr}$~\citep{yang2021DiverseMessagePassing}, $h_{\rm Euc}$~\citep{chen2023LSGNN}, and $h_{\rm GE}$~\citep{jin2022rawgnn}, score features according to different measures of homophily, that is, measuring the smoothness of each node feature according to different distance metrics.
    \item {\bf Rnd.:} Our \textbf{random (Rnd.)} selection baseline, where we select features uniformly at random to be retained or removed. 
\end{itemize}

% Our GFI metric ranks features based on the drop in validation accuracy when each feature is shuffled, while GFI-mask measures the performance drop when each feature is masked (set to zero). 
% Mutual Information (MI) scores features according to their mutual information with the labels, whereas Topological Feature Informativeness (TFI) computes the mutual information between the labels and $AX$, where $A$ is the adjacency matrix and $X$ the feature matrix. 
% The homophily-based metrics ($h_{attr}$, $h_{sim-euc}$, and $h_{GE}$) score features according to different measures of homophily, while Rnd. corresponds to a random selection baseline. 

\subsection{Table~\ref{t:adaptive_fs} and Figure~\ref{f:cora_algo} simulation details}\label{app:table3_info}

% \ibmblue{arxiv}
For adaptive node feature selection, we set $r=0.5$ in Algorithm~\ref{alg:nfpt} for all datasets, dropping half of the features at each step of the feature importance calculation, except for ArXiv, where we use $r=0.4$ due to its relatively small feature dimension (128).
The burn-in period $T_{\rm burn}$ and interval period $T$ are fixed to 50 for all datasets, except for Computers, Photos, and ArXiv, where $T_{\rm burn},T=100$. 
The model is trained for 400 epochs on all datasets and 800 epochs on Computers, Photos, and arxiv. 
Test accuracy for each feature percentage is reported based on the epoch with the highest validation accuracy: within each $T$ interval, we identify the epoch that achieves the best validation accuracy and use its corresponding test accuracy. 
This procedure is applied consistently across all feature selection methods.
The architecture and optimizer settings follow the configuration described in Section~\ref{app:table1_info}.

\section{Choice of number of feature permutations}\label{app:featperm}

Let $\{ \tbx^{(k)} \}_{k=1}^K$ denote $K\in\naturals$ independent permutations of the vector $\bbx \in \reals^N$, where $\tilde{x}_{i} = x_{\pi^{(k)}(i)}$ for every $i\in[N]$ for i.i.d. $\pi^{(k)} \in \Pi$.
We seek to sample a large enough $K$ such that the empirical expected value $\frac{1}{K} \sum_{k=1}^K \tbx^{(k)}$ approximates the true expected value $\mbE[ \tbx^{(k)} ] = \mu \bbone$ for any $k \in [K]$, where $\mu := \frac{1}{N} \bbone^\top \bbx$.
This will indicate that the empirical distribution of feature permutations approximates the true distribution.
To this end, we consider the following result.

\begin{proposition}\label{prop:sample_bound}
    For the vector $\bbx \in \reals^N$, we define $\{ \tbx^{(k)} \}_{k=1}^K$ such that $\tilde{x}_i = x_{\pi^{(k)}(i)}$ for every $k \in [K]$ and $i\in [N]$, where $\pi^{(k)} \in \Pi$ denote i.i.d. permutations of $[N]$.
    Then, with $x_{\max} := \max_i |x_i|$, we have that
    \alna{
        \mbP\Bigg[
            \Bigg\|
                \frac{1}{K}
                \sum_{k=1}^K
                \tbx^{(k)}
                -
                \mbE[\tbx^{(k)}]
            \Bigg\|_2^2
            \leq
            \frac{ t \sqrt{K-1} }{2K}
        \Bigg]
        \geq
        1 - 2
        \exp\left\{
            -\frac{K t^2 }{ 4N^2 ( x_{\max}^2 - \mu^2 )^2 }
        \right\}.
    \label{eq:sample_bound}}
\end{proposition}

Thus, we may choose $K$ in Algorithm~\ref{alg:nfpt} such that our feature permutations are similar enough to the true distribution of random feature permutations, where we determine a satisfactory similarity via choice of $t$.
The proof of Proposition~\ref{prop:sample_bound} is as follows.

\textbf{Proof of Proposition~\ref{prop:sample_bound}.}
First, given that $\mbE[\tbx^{(k)}] = \mu \bbone$, we have that
\alna{
    \Bigg\|
        \frac{1}{K}
        \sum_{k=1}^K
        \tbx^{(k)}
        -
        \mbE[\tbx^{(k)}]
    \Bigg\|_2^2
    &~=~&
    \sum_{i=1}^N
    \left(
        \frac{1}{K} \sum_{k=1}^K
        ( \tilde{x}_i^{(k)} - \mu )
    \right)^2
    &\nonumber\\&
    &~=~&
    \frac{1}{K^2}
    \sum_{i=1}^N
    \sum_{k=1}^K
    \sum_{\ell=1}^K
        ( \tilde{x}_i^{(k)} - \mu )
        ( \tilde{x}_i^{(\ell)} - \mu )
&\nonumber\\&
    &~=~&
    \frac{1}{K^2}
    \sum_{k=1}^K
    \sum_{\ell=1}^K
        ( \tbx^{(k)} - \mu\bbone )^\top
        ( \tbx^{(\ell)} - \mu\bbone ).
\nonumber}
% Then, since $\mu = \frac{1}{N} \bbone^\top \bbx$, we obtain
% \alna{
%     \Bigg\|
%         \frac{1}{K}
%         \sum_{k=1}^K
%         \tbx^{(k)}
%         -
%         \mbE[\tbx^{(k)}]
%     \Bigg\|_2^2
%     &~=~&
%     \frac{1}{K^2}
%     \sum_{k=1}^K
%     \sum_{\ell=1}^K
%     \tbx^{(k)\top} \tbx^{(\ell)}
%     +
%     \frac{K-2}{K}
%     \cdot
%     N\mu^2
% &\nonumber\\&
%     &~=~&
%     \frac{2}{K^2}
%     \sum_{k < \ell}
%     \tbx^{(k)\top} \tbx^{(\ell)}
%     +
%     \frac{1}{K^2}
%     \| \bbx \|_2^2
%     +
%     \frac{K-2}{K}
%     \cdot
%     N\mu^2
% &\nonumber\\&
%     &~=~&
%     \frac{2}{K^2}
%     \sum_{k < \ell}
%     (\tbx^{(k)\top} \tbx^{(\ell)} - N\mu^2)
%     +
%     \frac{K-1}{K}\cdot N\mu^2
%     +
%     \frac{1}{K^2} \| \bbx \|_2^2
%     +
%     \frac{K-2}{K}
%     \cdot
%     N\mu^2
% &\nonumber\\&
%     &~=~&
%     \frac{2}{K^2}
%     \sum_{k < \ell}
%     (\tbx^{(k)\top} \tbx^{(\ell)} - N\mu^2)
%     +
%     \frac{1}{K^2}
%     ( \| \bbx \|_2^2 - KN \mu^2 ).
% \label{eq:diff_rewrite}}
% We next bound the first term of the equality above.
Since $\pi^{(k)}$ are independently sampled uniformly at random from $\Pi$, for each $k,\ell \in [K]$ such that $k\neq \ell$, there exists some permutation $\rho^{(j)} \in \Pi$ such that $(\tbx^{(k)}-\mu\bbone)^\top (\tbx^{(\ell)}-\mu\bbone) = (\bbx-\mu\bbone)^\top (\hbx^{(j)}-\mu\bbone)$, where $\hbx^{(j)}$ denotes the permutation of $\bbx$ by $\rho^{(j)}$ for every $j\in [J]$ with $J := K(K-1)/2$.
Thus, our next step is to apply Hoeffding's inequality.
To this end, first observe that the $j$-th inner product $(\bbx-\mu\bbone)^\top (\hbx^{(j)}-\mu\bbone)$ denotes an independent random variable bounded between $-N |x_{\max}^2 - \mu^2|$ and $N |x_{\max}^2 - \mu^2|$.
Then, for any $t_0 > 0$, Hoeffding's inequality states that
\alna{
    \mbP\Bigg[
        \Bigg|
            \sum_{j=1}^J
            (\bbx - \mu \bbone)^\top
            (\hbx^{(j)} - \mu \bbone)
        \Bigg|
        >
        t_0
    \Bigg]
    \leq
    2 \exp\Bigg\{
        -\frac{t_0^2}{ 2J N^2 (x_{\max}^2 - \mu^2)^2 }
    \Bigg\}.
\nonumber}
Recalling that $J = \frac{K(K-1)}{2}$, we then let $t = \frac{t_0 \sqrt{K-1}}{J}$ for
\alna{
    \mbP\Bigg[
        \Bigg\|  
            \frac{1}{K} \sum_{k=1}^K \tbx^{(k)} - \mu\bbone
        \Bigg\|_2^2
        >
        t
        \frac{\sqrt{K-1}}{2K}
    \Bigg]
    \leq
    2 \exp\Bigg\{
        -\frac{Kt^2}{ 4N^2 ( x_{\max}^2 - \mu^2 )^2 }
    \Bigg\},
\nonumber}
as desired.
$\hfill\blacksquare$

\section{Additional plots on adaptive node feature selection}\label{app:adapt_plots}

\begin{figure*}[h]
    \centering
    % --------------------------------
    \scalebox{1.}{\includegraphics[width=\textwidth]{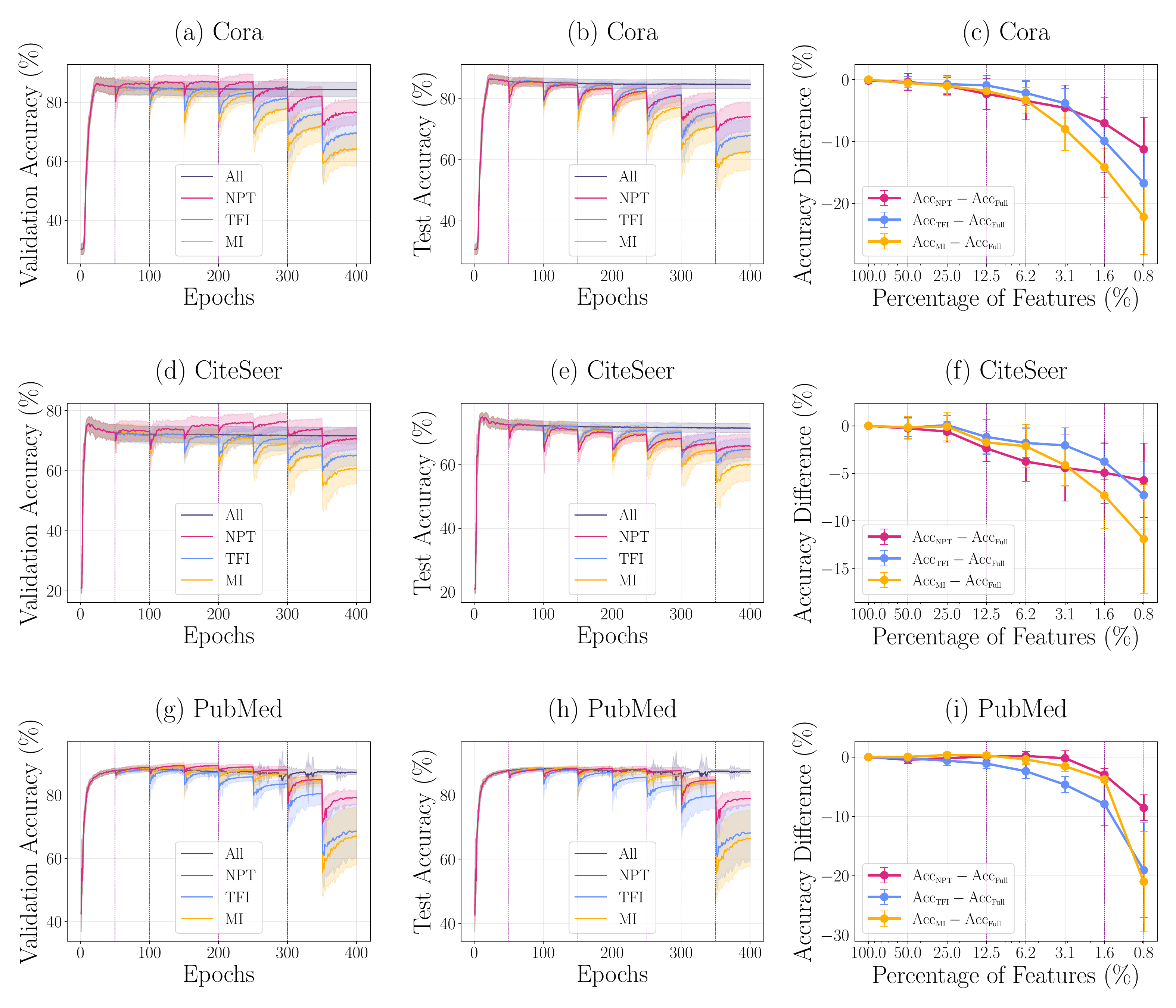}}
    % --------------------------------
    \caption{
        Node classification accuracy for homophilic datasets Cora, CiteSeer, and PubMed.
        (a,d,g) Validation accuracy for full, NPT, TFI, and MI.
        (b,e,h) Test accuracy for full, NPT, TFI, and MI.
        (c,f,i) Accuracy difference for full, NPT, TFI, and MI.
    }
    \label{f:acc_homophilic}
\end{figure*}

\begin{figure*}[h]
    \centering
    % --------------------------------
    \scalebox{1.}{\includegraphics[width=\textwidth]{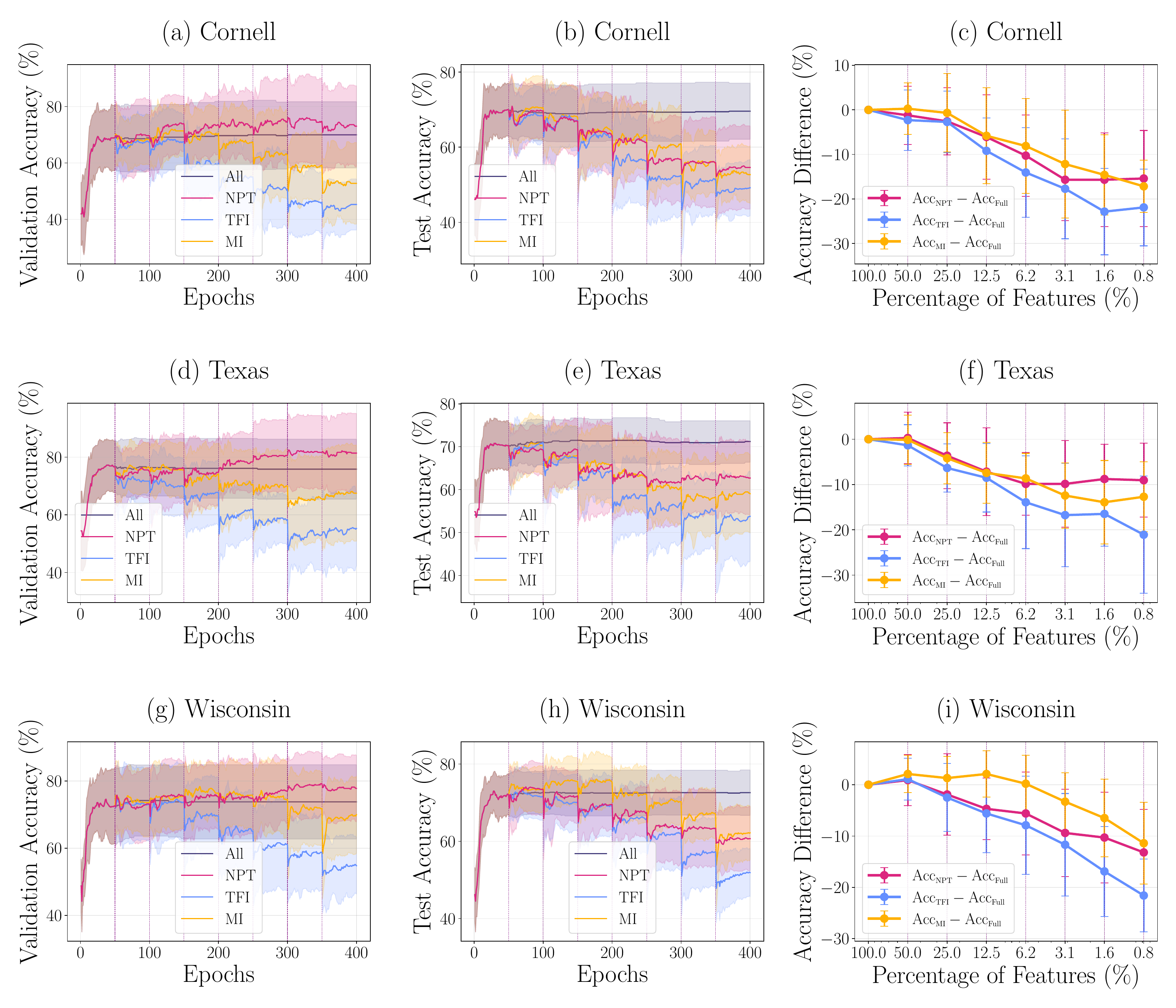}}
    % --------------------------------
    \caption{
        Node classification accuracy for heterophilic datasets Cornell, Texas, and Wisconsin.
        (a,d,g) Validation accuracy for full, NPT, TFI, and MI.
        (b,e,h) Test accuracy for full, NPT, TFI, and MI.
        (c,f,i) Accuracy difference for full, NPT, TFI, and MI.
    }
    \label{f:acc_heterophilic}
\end{figure*}

\begin{figure*}[h]
    \centering
    % --------------------------------
    \scalebox{1.}{\includegraphics[width=\textwidth]{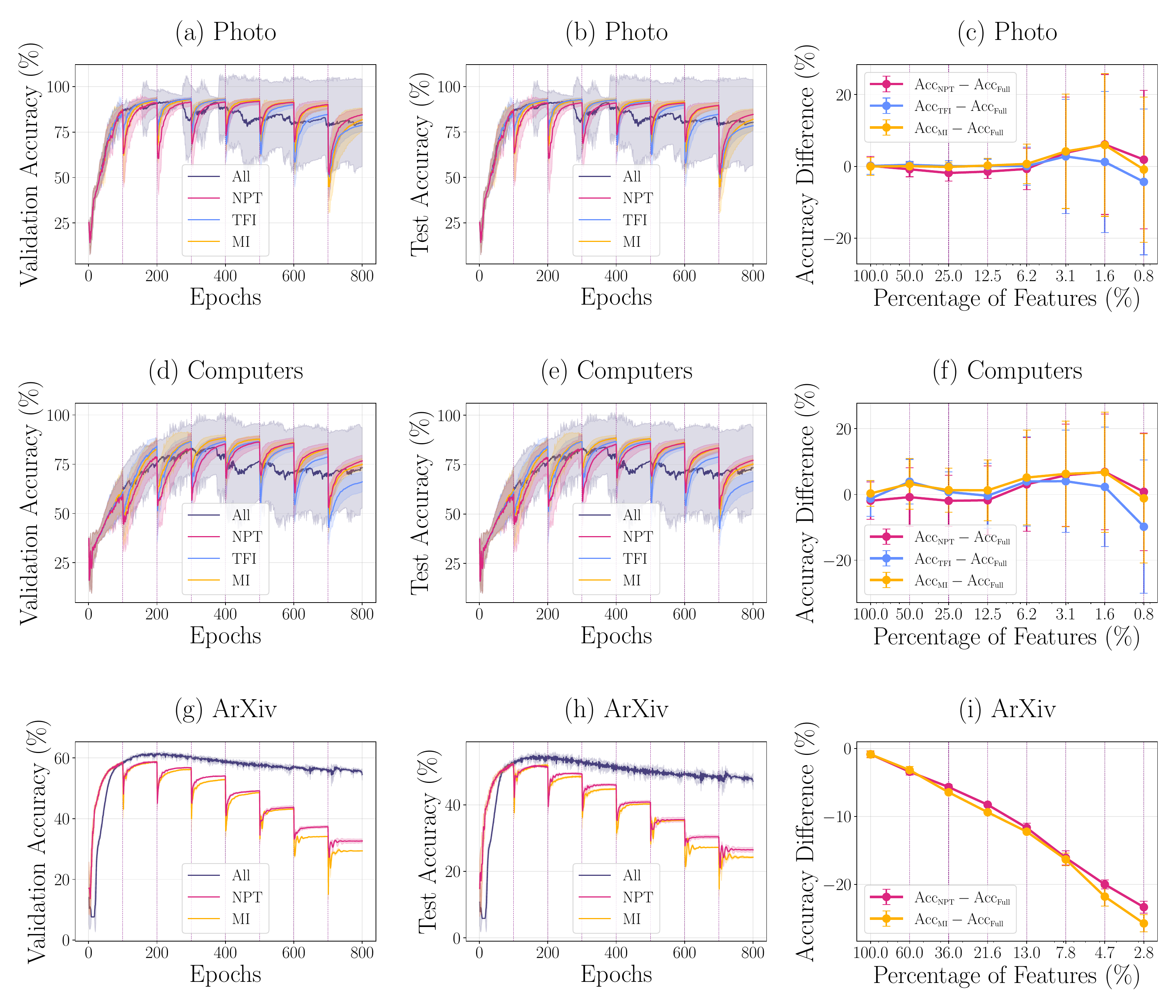}}
    % --------------------------------
    \caption{
        Node classification accuracy for larger-scale datasets Photo, Computers, and ArXiv.
        (a,d,g) Validation accuracy for full, NPT, TFI, and MI.
        (b,e,h) Test accuracy for full, NPT, TFI, and MI.
        (c,f,i) Accuracy difference for full, NPT, TFI, and MI.
    }
    \label{f:acc_larger}
\end{figure*}

\begin{figure*}[h]
    \centering
    % --------------------------------
    \scalebox{1.}{\includegraphics[width=\textwidth]{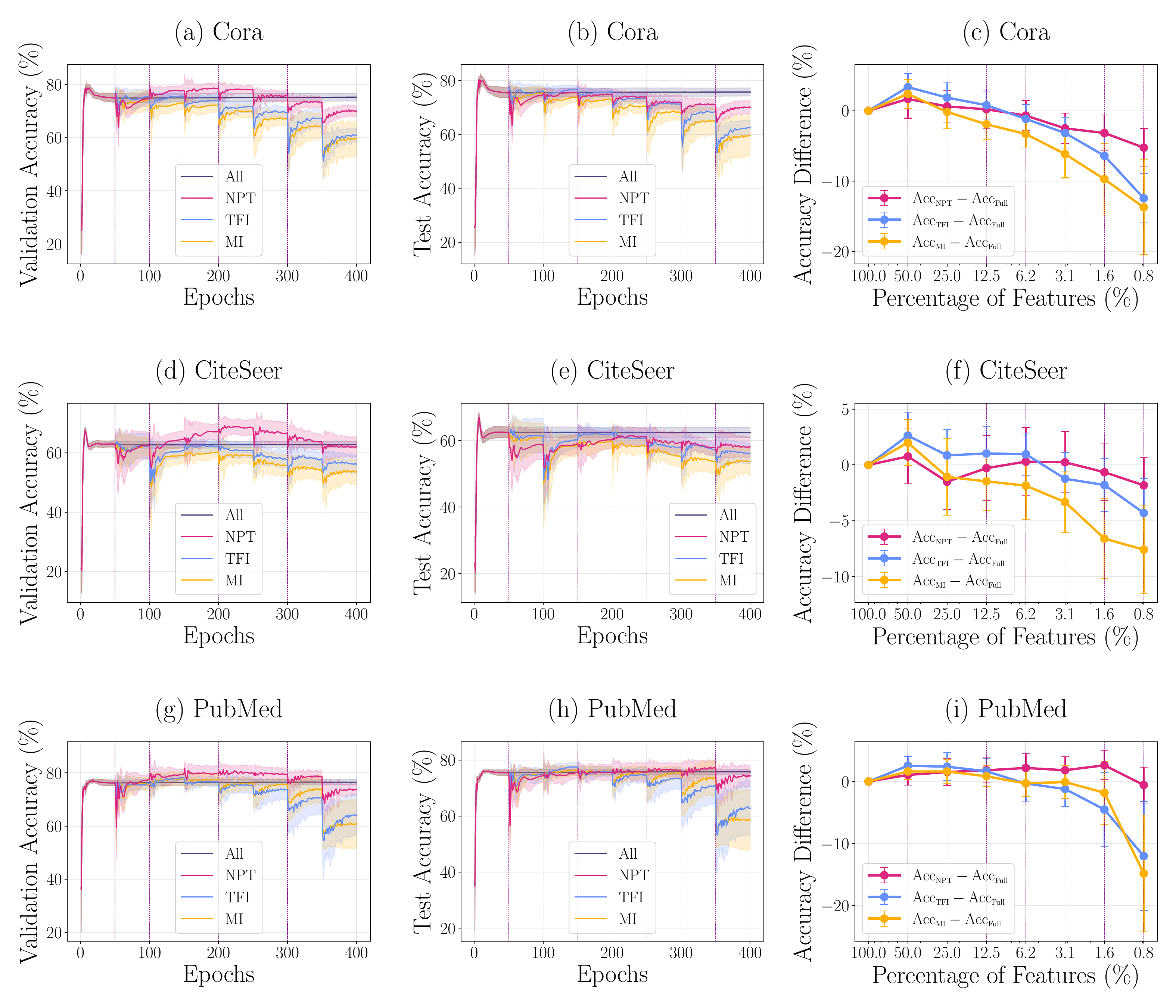}}
    % --------------------------------
    \caption{
        Node classification accuracy for homophilic datasets Cora, CiteSeer, and PubMed.
        Train, validation, and test node subsets are selected via canonical splits.
        (a,d,g) Validation accuracy for full, NPT, TFI, and MI.
        (b,e,h) Test accuracy for full, NPT, TFI, and MI.
        (c,f,i) Accuracy difference for full, NPT, TFI, and MI.
    } 
    \label{f:acc_homophilic_can}
\end{figure*}

We present additional plots analogous to those in Figure~\ref{f:cora_algo} measuring the accuracy of GNNs trained using either all features available in a dataset versus using our adaptive Algorithm~\ref{alg:nfpt}.
We compare our approach using our proposed permutation-based node feature importance scores, and we also evaluate using {\bf TFI} and {\bf MI} as importance metrics to rank feature relevance in Algorithm~\ref{alg:nfpt}.
Figure~\ref{f:acc_homophilic} presents results for the datasets with homophilic labels Cora, CiteSeer, and PubMed; Figure~\ref{f:acc_heterophilic} the datasets with heterophilic labels Cornell, Texas, and Wisconsin; and Figure~\ref{f:acc_larger} the larger-scale datasets Photo, Computers, and ArXiv.
To further verify our results, we repeat our experiments on homophilic datasets Cora, CiteSeer, and PubMed using their official dataset splits into training, validation, and testing in Figure~\ref{f:acc_homophilic_can}.

\section{Additional plots on feature importance analysis}\label{app:feat_import}

\begin{figure*}[h]
    \centering
    % --------------------------------
    \scalebox{1.}{\includegraphics[width=\textwidth]{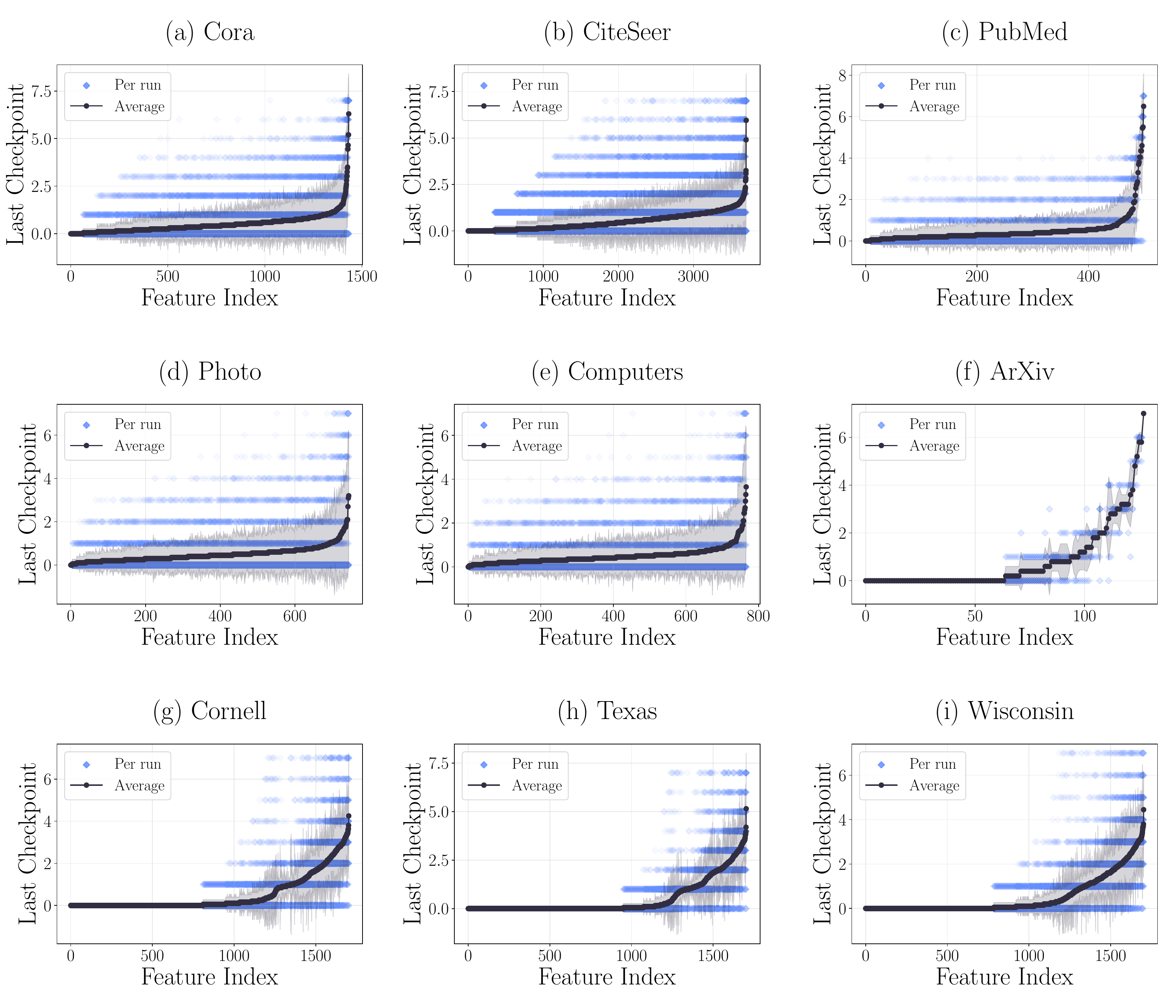}}
    % --------------------------------
    \caption{
        Last checkpoint kept per feature for various datasets.
        Plots (a) through (i) are presented in the following order: 
        (a) Cora, (b) CiteSeer, (c) PubMed, (d) Photo, (e) Computers, (f) ArXiv, (g) Cornell, (h) Texas, (i) Wisconsin.
        % (a) Checkpoints per feature for Cora.
        % (b) Checkpoints per feature for CiteSeer.
        % (c) Checkpoints per feature for PubMed.
        % (d) Checkpoints per feature for Photo.
        % (e) Checkpoints per feature for Computers.
        % (f) Checkpoints per feature for ArXiv.
        % (g) Checkpoints per feature for Cornell.
        % (h) Checkpoints per feature for Texas.
        % (i) Checkpoints per feature for Wisconsin.
        % Node classification accuracy for larger-scale datasets Photo, Computers, and ArXiv.
        % (a,d,g) Validation accuracy for full, NPT, TFI, and MI.
        % (b,e,h) Test accuracy for full, NPT, TFI, and MI.
        % (c,f,i) Accuracy difference for full, NPT, TFI, and MI.
    } 
    \label{f:acc_ranking_by_feat}
\end{figure*}

\begin{figure*}[h]
    \centering
    % --------------------------------
    \scalebox{.8}{\includegraphics[width=\textwidth]{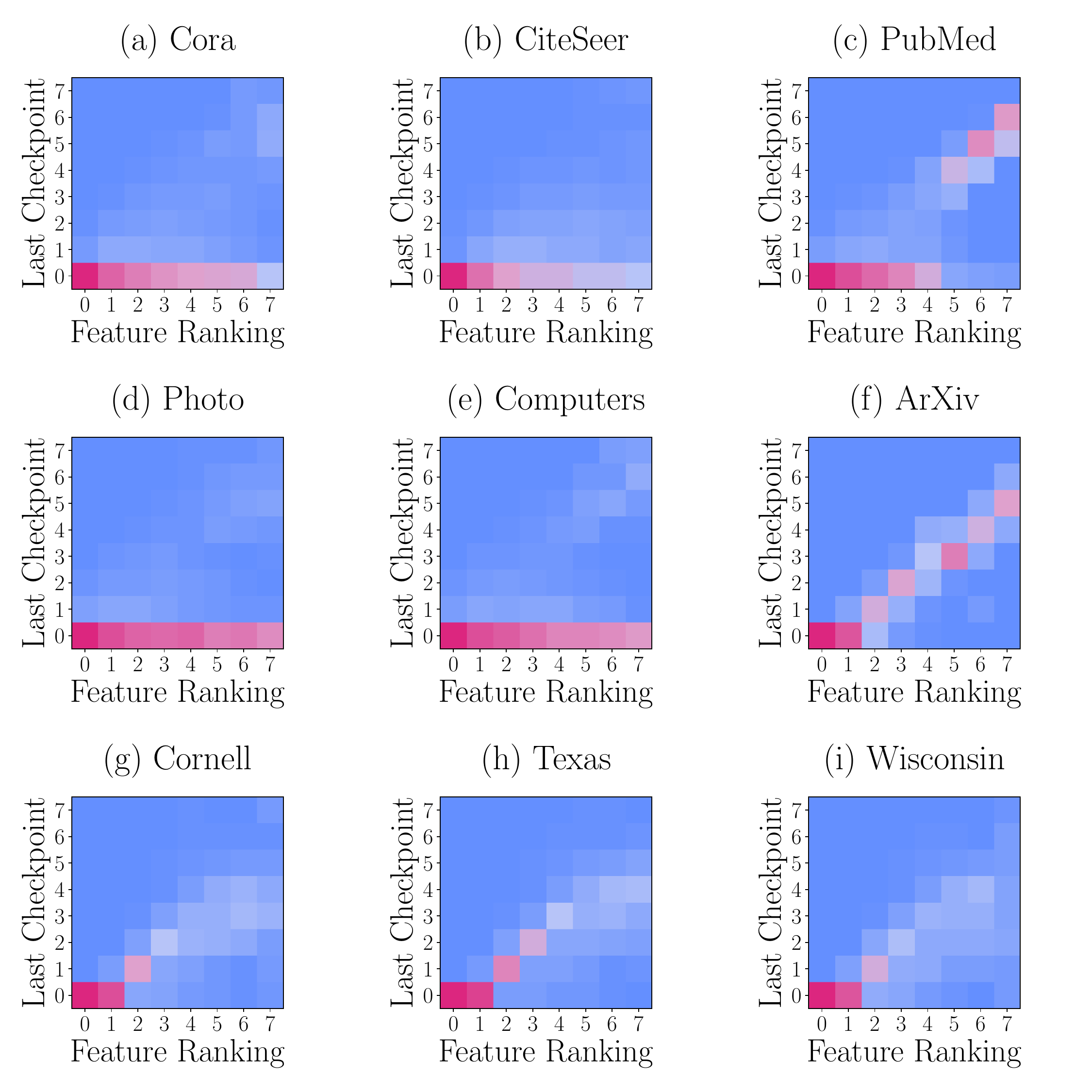}}
    % --------------------------------
    \caption{
        Last checkpoint kept per feature ranking for various datasets.
        Feature ranking for each feature corresponds to the most common last checkpoint the feature is kept before being dropped across independent trials.
        Each entry of a heatmap denotes the average last checkpoint across all features in the same ranking.
        Plots (a) through (i) are presented in the following order: 
        (a) Cora, (b) CiteSeer, (c) PubMed, (d) Photo, (e) Computers, (f) ArXiv, (g) Cornell, (h) Texas, (i) Wisconsin.
        % (a) Checkpoints per feature ranking for Cora.
        % (b) Checkpoints per feature ranking for CiteSeer.
        % (c) Checkpoints per feature ranking for PubMed.
        % (d) Checkpoints per feature ranking for Photo.
        % (e) Checkpoints per feature ranking for Computers.
        % (f) Checkpoints per feature ranking for ArXiv.
        % (g) Checkpoints per feature ranking for Cornell.
        % (h) Checkpoints per feature ranking for Texas.
        % (i) Checkpoints per feature ranking for Wisconsin.
        % Node classification accuracy for larger-scale datasets Photo, Computers, and ArXiv.
        % (a,d,g) Validation accuracy for full, NPT, TFI, and MI.
        % (b,e,h) Test accuracy for full, NPT, TFI, and MI.
        % (c,f,i) Accuracy difference for full, NPT, TFI, and MI.
    } 
    \label{f:acc_ranking_by_rank}
\end{figure*}

\begin{figure*}[h]
    \centering
    % --------------------------------
    \scalebox{1.}{\includegraphics[width=\textwidth]{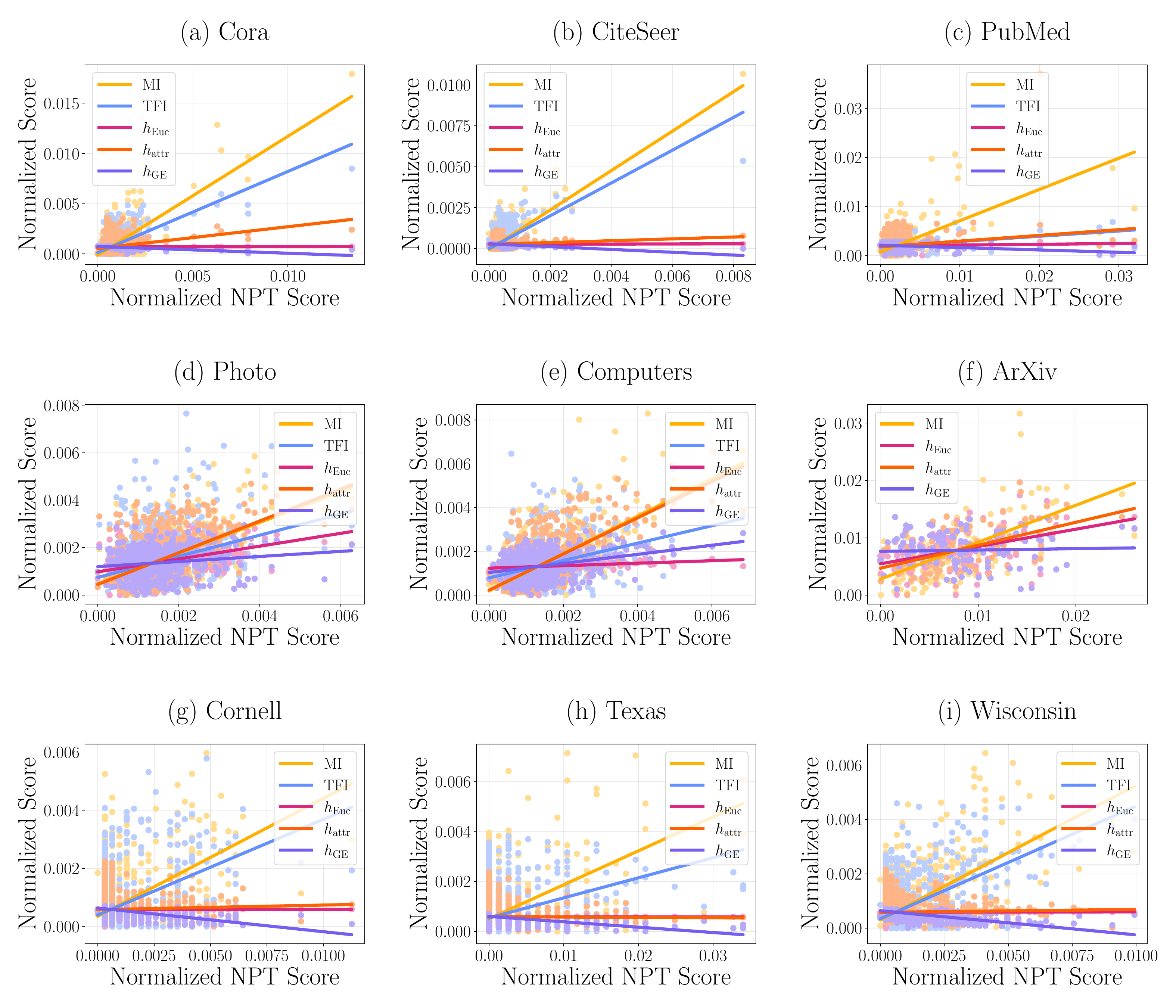}}
    % --------------------------------
    \caption{ 
        Plot of normalized importance scores per baseline versus normalized NPT scores.
        Plots (a) through (i) are presented in the following order: 
        (a) Cora, (b) CiteSeer, (c) PubMed, (d) Photo, (e) Computers, (f) ArXiv, (g) Cornell, (h) Texas, (i) Wisconsin.
        % (a) Baseline scores versus NPT scores for Cora.
        % (b) Baseline scores versus NPT scores for CiteSeer.
        % (c) Baseline scores versus NPT scores for PubMed.
        % (d) Baseline scores versus NPT scores for Photo.
        % (e) Baseline scores versus NPT scores for Computers.
        % (f) Baseline scores versus NPT scores for ArXiv.
        % (g) Baseline scores versus NPT scores for Cornell.
        % (h) Baseline scores versus NPT scores for Texas.
        % (i) Baseline scores versus NPT scores for Wisconsin.
    } 
    \label{f:correlations}
\end{figure*}

This section includes Figure~\ref{f:acc_ranking_by_feat}, which contains additional plots analogous to Figure~\ref{f:heatmaps}c.
For each feature, we measure the average last checkpoint of Algorithm~\ref{alg:nfpt} in which a feature is kept before being dropped for all datasets.
In addition, Figure~\ref{f:acc_ranking_by_rank} plots the average last checkpoint for each feature rank, that is, the most frequent last checkpoint assigned to each feature, analogous to Figure~\ref{f:heatmaps}d.
We also include additional plots analogous to Figure~\ref{f:correlations} for all datasets.

\begin{figure*}[h]
    \centering
    % --------------------------------
    \scalebox{1.}{\includegraphics[width=\textwidth]{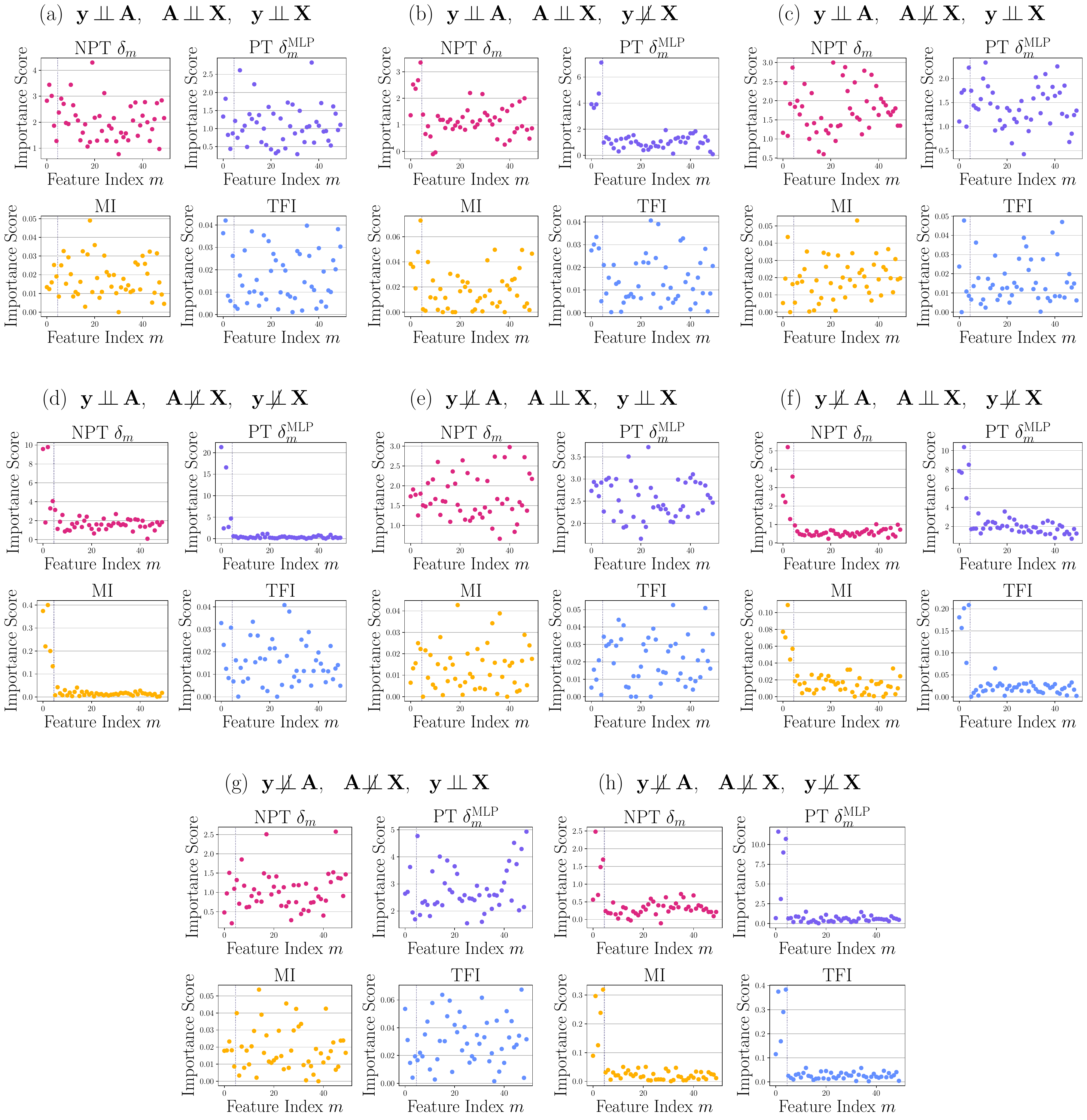}}
    % --------------------------------
    \caption{ 
        Feature importance scores for NPT, PT (permutation testing via MLP), TFI, and MI for synthetic datasets.
        Analogous to $\delta_m$ for NPT scores, $\delta_m^{\rm MLP}$ denotes PT scores.
        (a) Graph, labels, and features are all independent.
        (b) Labels and features are correlated, but both are independent of graph.
        (c) Graph and features are correlated, but both are independent of labels.
        (d) Features are correlated with graph and labels, but labels and graph are independent.
        (e) Graph and labels are correlated, but both are independent of features.
        (f) Labels are correlated with graph and features, but graph and features are independent.
        (g) Graph is correlated with labels and features, but labels and features are independent.
        (h) Graph, labels, and features are all pairwise correlated.
    }
    \label{f:fi_synth}
\end{figure*}

Finally, we plot in Figure~\ref{f:fi_synth} the feature importance for synthetic graph data measured by \textbf{NPT}, \textbf{TFI}, \textbf{MI}, and \textbf{PT}, which is analogous to \textbf{NPT} but uses an MLP instead of a GNN to compute feature importance.
More specifically, we generate five independent trials of graphs of $N = 500$ nodes and $M = 50$ features.
We assign nodes to one of $C = 2$ classes.
We vary the relationships between features, labels, and graph structure as follows.

\begin{itemize}%[left= 2pt .. 12pt, noitemsep]
    \item {\bf Graph structure $\bbA$ $\leftrightarrow$ labels $\bby$}:
          When the graph and labels are independent ($\bbA \indep \bby$), we generate an Erdos-Renyi graph with edge probability $0.1$.
          Otherwise, when ($\bbA \not\! \indep \bby$), we sample the graph from a stochastic block-model whose communities correspond to classes, where within-class edges are sampled with probability $0.1$ and across-class edges with $0.05$.
    \item {\bf Graph structure $\bbA$ $\leftrightarrow$ node features $\bbX$}:
          When the graph and features are independent ($\bbA \indep \bbX$), we sample node features as Gaussian white noise $\bbX_0 \sim \ccalN( \bbzero, \sigma \bbI )$ for $\sigma = 3$.
          Otherwise, when $\bbA \not\! \indep \bbX$, we obtain the eigendecomposition of $\bbA = \bbV\bbLambda \bbV^\top$ and generate bandlimited graph signals as $\bbX_0 = \bbV_{:,\ccalB} \bbW$ for $\bbW \sim \ccalN(\bbzero, \sigma \bbI)$, where $\ccalB$ denotes the indices of graph frequencies in $\diag(\bbLambda)$ that are below $\lambda_{\rm max} = 0.5 |\max_i \Lambda_{ii}|$.
    \item {\bf Labels $\bby$ $\leftrightarrow$ node features $\bbX$}:
          When the graph and labels are independent ($\bby \indep \bbX$), we further process node features by sampling $\bbB_0 \sim \ccalN(\bbzero,\bbI)$ for $\bbB_0 \in \reals^{C \times 5}$.
          We normalize the columns of $\bbB_0$ to sum to zero and rescale for $\bbB = 5 \diag^{-1}( |\bbB_0| \bbone ) \bbB_0$, where $|\bbB_0|$ denotes the element-wise absolute value of entries of $\bbB_0$.
          Finally, we update relevant entries of $\bbX$ as $\bbX_{:,m} = [\bbX_0]_{:,m} + [\bbY\bbB]_{:,m}$ for $m \in [5].$
          Otherwise, when $\bby \not\! \indep \bbX$, we simply let $\bbX = \bbX_0$.
\end{itemize}

\section{Additional plots on model performance analysis}\label{app:model_perf}

\begin{figure*}[h]
    \centering
    % --------------------------------
    \scalebox{1.}{\includegraphics[width=\textwidth]{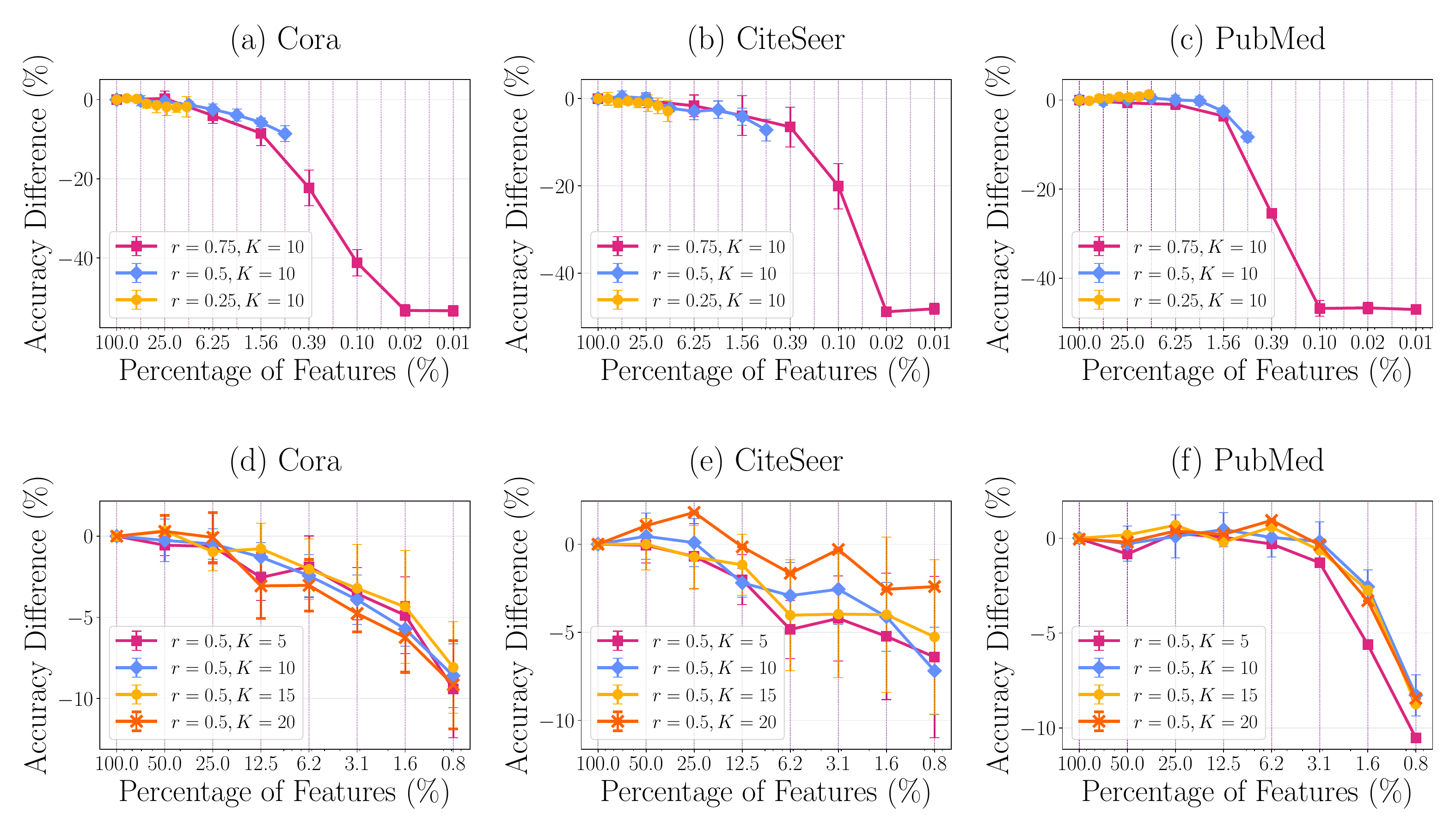}}
    % --------------------------------
    \caption{ 
        Hyperparameter tuning by comparing GCN performance across various dropping rates $r$ and shuffling instances $K$ for Cora, CiteSeer, and PubMed.
        The top row corresponds to fixing $K = 10$ while varying $r \in \{ 0.25,0.5,0.75 \}$, and the bottom row denotes fixing $r = 0.5$ while varying $K \in \{ 5, 10, 15, 20 \}$.
        % (a) Cora with varying $r$.
        % (b) CiteSeer with varying $r$.
        % (c) PubMed with varying $r$.
        % (d) Cora with varying $K$.
        % (e) CiteSeer with varying $K$.
        % (f) PubMed with varying $K$.
    } 
    \label{f:hyperparam}
\end{figure*}

We include additional plots in Figure~\ref{f:hyperparam} on hyperparameter tuning for Cora, CiteSeer, and PubMed, which correspond to Figure~\ref{f:ablation}a,b.
In particular, we fix $r = 0.5$ and vary $K \in \{ 5,10,15,20 \}$ in the top row, whereas for the bottom row, we fix $K = 10$ and vary $r \in \{ 0.25, 0.5, 0.75 \}$.

%%%%%%%%%%%%%%%%%%%%%%%%%%%%%%%%%%%%%%%%%%%%%%%%%%%%%%%%%%%%%%%%%%%%%%%%%%%%%%%

\end{document}